%% file: main.tex
\documentclass[acmtog]{acmart}

\acmJournal{TOG}

\copyrightyear{2025}
\acmYear{2025}
\setcopyright{cc}
\setcctype{by}
\acmConference[SA Conference Papers '25]{SIGGRAPH Asia 2025 Conference Papers}{December 15--18, 2025}{Hong Kong, Hong Kong}
\acmBooktitle{SIGGRAPH Asia 2025 Conference Papers (SA Conference Papers '25), December 15--18, 2025, Hong Kong, Hong Kong}
\acmDOI{10.1145/3757377.3763973}
\acmISBN{979-8-4007-2137-3/2025/12}

\input{macros.tex}

\title{Lang3D-XL: Language Embedded 3D Gaussians for Large-scale Scenes}

\author{Shai Krakovsky}
\affiliation{%
  \institution{Tel Aviv University}
  \city{Tel Aviv}
  \country{Israel}
}
\email{krakovsky1@mail.tau.ac.il}

\author{Gal Fiebelman}
\affiliation{%
  \institution{The Hebrew University of Jerusalem}
  \city{Jerusalem}
  \country{Israel}
}
\email{gal.fiebelman@mail.huji.ac.il}

\author{Sagie Benaim}
\affiliation{%
  \institution{The Hebrew University of Jerusalem}
  \city{Jerusalem}
  \country{Israel}
}
\email{sagiebenaim@gmail.com}

\author{Hadar Averbuch-Elor}
\affiliation{%
  \institution{Cornell University}
  \city{Ithaca, NY}
  \country{USA}
}
\email{hadarelor@cornell.edu}

\begin{document}

\input{abstract_arxiv}

\begin{teaserfigure}
    \centering    \includegraphics[width=1.0\textwidth]{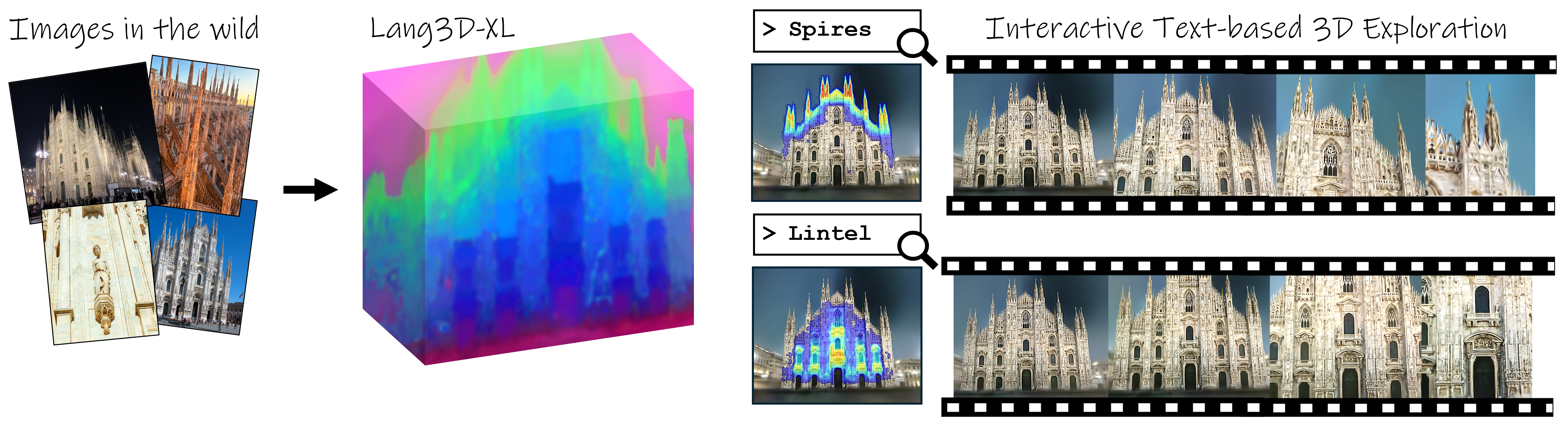}
    \caption{\textbf{Language Embedded 3D Gaussians for Large-scale Scenes}. Given Internet collections of images depicting large-scale scenes (left), we augment a 3D Gaussian representation with a learnable semantic bottleneck (center). Our approach enables interactive text-based virtual exploration (right), enabling users to zoom-in and view semantic regions of interests, such as the spires and lintels$^*$ in the Milano Catedral depicted above. \\
    {\scriptsize  $^*$A spire refers to a tall, slender, pointed structure on top of a roof of a building or tower, and a lintel is a horizontal structural element that spans openings such as doors and windows.} 
    }\label{fig:teaser}
\end{teaserfigure}


\maketitle

\input{01-intro}

\input{02-rw}

\input{03-method}

\input{04-results}

\input{05-conclusion}
\begin{acks}
This work was partially supported by ISF (grant number 2510/23).
\end{acks}
\bibliographystyle{ACM-Reference-Format}
\bibliography{main}

\input{figures/our_results}

\input{figures/final_results/final_results_visualization}

\input{figures/final_results/final_results_2D_visualization}

\input{figures/ablation_results/attention_seg_ablation}

\clearpage
\appendix

\noindent {\LARGE\textbf{\methodname{} Supplementary Material}}

\input{supp/supp_details}

\input{supp/supp_additional_results}

\end{document}

%% file: macros.tex
\usepackage{multirow}
\usepackage{xcolor}
\usepackage{xspace}

\newcommand{\rev}[1]{{#1}}

\newcommand{\revB}[1]{{#1}}

\newcommand{\methodname}{\emph{Lang3D-XL}\xspace}

\newcommand{\whitetxt}[1]{{\color{white}#1}\normalfont}
\usepackage{color}
\definecolor{gray}{rgb}{0.6,0.6,0.6}
\definecolor{red}{rgb}{1,0,0}
\definecolor{green}{rgb}{0,1,0}
\definecolor{blue}{rgb}{0,0,1}
\definecolor{dark-green}{rgb}{0,0.4,0}
\definecolor{orange}{rgb}{1,0.55,0}
\definecolor{white}{rgb}{1,1,1}
\definecolor{black}{rgb}{0,0,0}
\definecolor{dark-brown}{rgb}{0.2,0.1,0}
\definecolor{light-blue}{rgb}{0.4,0.6,0.99}
\definecolor{dark-red}{rgb}{0.6,0,0}
\definecolor{light-red}{rgb}{1,0.2,0.6}
\definecolor{pink}{rgb}{1,0.2,0.6}
\definecolor{dark-pink}{rgb}{0.6,0,0.3}

\newbox\jsavebox
\newcommand{\jsubfig}[2]{%
	\sbox\jsavebox{#1}%
	\parbox[t]{\wd\jsavebox}{\centering\usebox\jsavebox\\#2}%
	}

%% file: abstract_arxiv.tex
\begin{abstract}
    
Embedding a language field in a 3D representation %
enables richer semantic understanding of spatial environments by linking geometry with descriptive meaning. This allows for a more intuitive human-computer interaction, enabling querying or editing scenes using natural language, and could potentially improve tasks like scene retrieval, navigation, and multimodal reasoning. 
While such capabilities could be transformative, in particular for large-scale scenes, we find that recent feature distillation approaches cannot effectively learn over massive Internet data due to challenges in semantic feature misalignment and inefficiency in memory and runtime.  %
To this end, we propose a novel approach to address these challenges. First, we introduce extremely low-dimensional semantic bottleneck features as part of the underlying 3D Gaussian representation. These are processed by rendering and passing them through a multi-resolution, feature-based, hash encoder. This significantly 
improves efficiency both in runtime and GPU memory. Second, we introduce an Attenuated Downsampler module and propose several regularizations addressing the semantic misalignment of ground truth 2D features. 
We evaluate our method on the in-the-wild HolyScenes dataset and demonstrate that it surpasses existing approaches in both performance and efficiency. Project page: \url{https://tau-vailab.github.io/Lang3D-XL/}.
\end{abstract}

%% file: 01-intro.tex
\section{Introduction}
World landmarks such as the Milano Cathedral are more than architectural marvels---they are living testaments to centuries of cultural, artistic, and historical evolution. 
The emergence of scene-scale neural representations~\cite{martin2021nerf,chen2022hallucinated,zhang2024gsw} has enabled the creation of realistic virtual worlds, presenting new possibilities for exploring the world's monuments via digital platforms.
Such digital exploration, however,  necessitates a tight coupling between the scene's low-level fine-grained geometry and a higher-level semantic understanding.

Recently, HaLo-NeRF~\cite{dudai2024halo}
proposed to augment a neural representation of large-scale scenes with a semantic localization head that optimizes volumetric probabilities for each queried text prompt---however, this yields a typical runtime of roughly two \emph{hours} for each prompt.
This time-consuming \emph{per-prompt} optimization procedure significantly hinders its potential for digital exploration or other human-assisted applications, such as real-time text-based navigation or visual question answering. Inspired by the impressive open-vocabulary capabilities of feature distillation techniques---from DFF~\cite{kobayashi2022decomposing} to Feature3DGS~\cite{zhou2023feature}, LangSplat~\cite{qin2023langsplat} and others~\cite{shi2023language,liao2024clipgs,kim2025drsplat}, we ask: \emph{Can these feature distillation techniques successfully represent large-scale Internet scenes?} 

Surprisingly, we find that existing distillation methods are challenged by multiple factors that prevent them from effectively learning over large-scale Internet scenes. These include practical limitations, such as GPU memory constraints, but not only---our analysis reveals that direct supervision using pixel-aligned 2D feature pyramids yields non-negligible semantic misalignments, where spatially adjacent regions are "contaminated" with false semantics.

Accordingly, we introduce \methodname{}, an approach that augments a 3D Gaussian Splatting representation with a learnable semantic bottleneck, allowing for efficiently distilling language embeddings for large-scale scenes.
This \emph{extremely} low-dimensional semantic bottleneck is rendered and then processed by a multi-scale semantic hash encoder, which outputs high-dimensional semantic features. Specifically, it predicts concatenated CLIP~\cite{radford2021learning} and DINOv2~\cite{oquab2023dinov2} features. Directly supervising  these features using pixel-aligned 2D feature pyramids yields non-negligible semantic misalignments. To this end, 
we propose an Attenuated Downsampler module and several regularizations aimed at bridging this semantic misalignment.

\methodname{} unlocks interactive text-guided exploration over large-scale scenes, allowing users to zoom-into regions containing the queried text and view them from various viewpoints; see Fig.~\ref{fig:teaser} for an illustration. We evaluate our framework on the HolyScenes benchmark~\cite{dudai2024halo}, comparing segmentation performance to prior segmentation techniques---both in 2D and 3D---and feature distillation techniques. Our experiments show that \methodname{} significantly outperforms existing feature distillation techniques, yielding results comparable with HaLo-NeRF, while being orders of magnitude faster. Explicitly stated, our contributions are:
\begin{itemize}
    \item An approach for efficiently augmenting large-scale scenes represented using 3D Gaussians with language features.
    \item Novel mechanisms for bridging semantic misalignments in image-level feature distillation pipelines, allowing for significantly boosting downstream performance.
    \item Strong performance over challenging Internet scenes, achieved with interactive inference times, orders of magnitude faster than prior work targeting such large-scale scenes.
\end{itemize}

%% file: 02-rw.tex
\section{Related Work}

\subsection{\revB{Language Embedded Neural Representations}}
Connecting 3D representations with semantics is key for a diverse set of applications, such as object localization, recognition, and semantic segmentation. Earlier works augmented Neural Radiance Fields~\cite{mildenhall2021nerf} (NeRFs) with semantic information via a classification branch~\cite{zhi2021in-place, kundu2022panoptic, siddiqui2022panoptic}. In particular, HaLo-NeRF~\cite{dudai2024halonerf} optimizes a semantic neural field over large-scale landmarks. However, this procedure requires timely optimization \emph{per} text prompt (yielding inference times of roughly two hours).

DFF~\cite{kobayashi2022decomposing} and N3F~\cite{tschernezki2022neural} first introduced the concept of 2D distillation, using a NeRF as the underlying 3D model to which features are distilled. This paradigm enables open-vocabulary segmentation, as the distilled features can be probed directly, for example, using free text.
LERF~\cite{kerr2023lerf} has extended these approaches to learn 3D open-vocabulary language embeddings by using (i). A pyramid of 2D CLIP features for training, and (ii). A DINO~\cite{caron2021emerging} regularization loss on rendered features.

The emergence of 3D Gaussian Splatting (3DGS) has spurred numerous efforts to integrate language features into its efficient, explicit representation. 
Several approaches focus on how to associate semantic features with Gaussians. For instance, OpenGaussian~\cite{wu2024opengaussian} links full-fidelity CLIP features to individual Gaussians for point-level understanding, emphasizing 3D consistency. Semantic Gaussians~\cite{shi2024semanticgaussians} distill 2D semantics via either direct projection or a learned 3D network. Semantic Consistent Language Gaussian Splatting~\cite{yin2025semanticconsistent} enhances distillation fidelity using SAM2 ``masklets'' for more accurate ground-truth supervision. SuperGSeg~\cite{chen2024supergseg} first learns segmentation features and then builds ``Super-Gaussians'' as structured units to ground language features for open-vocabulary segmentation. 
Differently, our work targets semantic misalignments arising from 2D feature pyramid distillation in diverse, large-scale static internet scenes by introducing an \textbf{Attenuated Downsampler Module} and complementary regularization. We also consider the challenge of accurately and efficiently distilling language features in \textit{large-scale static internet scenes}. 

\medskip
\subsection{\revB{Mitigating Memory and Runtime Overhead}}
A key challenge in lifting features from powerful vision-language models is the significant memory and computational overhead associated with their typically high-dimensional embeddings. Efficiently managing these is essential for scalability in large-scale scenes.
Several strategies aim to learn or distill \textit{compact per-Gaussian representations}. LangSplat~\cite{qin2023langsplat} and Online Language Splatting~\cite{katragadda2025online} employ autoencoders to compress CLIP features. Feature3DGS~\cite{zhou2023feature} distills a reduced-dimension feature field, reconstructing full features via a learned CNN. CLIP-GS~\cite{liao2024clipgs} learns compact semantic attributes, while LEGaussians~\cite{shi2023language} and Dr. Splat~\cite{kim2025drsplat} utilize dedicated quantization techniques for per-Gaussian semantics. 

\input{figures/overview/architecture}

Other approaches focus on \textit{efficient data structures and processing pipelines}. FastLGS~\cite{ji2024fastlgs} and FMLGS~\cite{tan2025fmlgs} utilize semantic feature grids for speed. Semantic Gaussians~\cite{shi2024semanticgaussians} enables direct and efficient queries. Algorithmic speed-ups are achieved by Occam's LGS~\cite{cheng2024occam}, GAGS~\cite{peng2024gags}, and CAGS~\cite{sun2025cags} through filtering or sampling. GOI~\cite{qu2024goi} uses hyperplanes and compression for region selection. For sparse views, SLGaussian~\cite{chen2024slgaussian} and SparseLGS~\cite{hu2024sparselgs} provide fast inference with low-dimensional features. Application-specific efficiencies are seen in LEGS~\cite{yu2024language} and GaussianGrasper~\cite{zheng2024gaussiangrasper}, while GSemSplat~\cite{wang2024gsemsplat} and PanoGS~\cite{zhai2025panogs} target reduced costs for specialized large-scale tasks.
A distinct strategy involves \textit{implicit or hash-based representations}. FAST-Splat \cite{shorinwa2024fast} uses direct semantic codes with hash-based similarity. FMGS~\cite{zuo2024fmgsfoundationmodelembedded}, inspired by Instant-NGP~\cite{M_ller_2022}, learns a hash-based embedding from 3D Gaussian centers to avoid storing explicit high-dimensional features per Gaussian.
Our approach, also leveraging a hash-based encoder in our \textbf{Semantic Bottleneck Decoder}, offers two innovations which significantly improve efficiency and overall quality: (i). Unlike FMGS and Fast-Splat, our hash-based encoding is performed in 2D feature space, following rendering, and not in 3D Gaussian space. 
This significantly reduces 3D memory and rendering costs compared as rendering is performed with low dimensional features. (ii). Our hashing is performed with respect to \emph{similar features} and not \emph{similar locations}, allowing similar features in different locations of the rendered image to be hashed similarly.

\medskip
\subsection{In-the-wild Novel View Synthesis}
A parallel stream of research improves 3GS robustness for novel view synthesis from "in-the-wild" static image collections. SWAG~\cite{dahmani2024swag} models appearance variations with per-image embeddings and transient object opacity. GS-W~\cite{zhang2024gsw} uses separated intrinsic/dynamic appearance features and adaptive sampling. Wild-GS~\cite{xu2024wildgs} employs hierarchical appearance models and triplane feature sampling for high-fidelity rendering. While these works tackle the visual complexities of in-the-wild static data for view synthesis, our primary focus is different: we aim to efficiently and accurately \textit{embed and localize natural language features} within these challenging large-scale static 3D scenes given "in-the-wild" static image collections.

%% file: figures/overview/architecture.tex
\begin{figure*} %
\centering
\jsubfig{\includegraphics[width=0.99\textwidth]{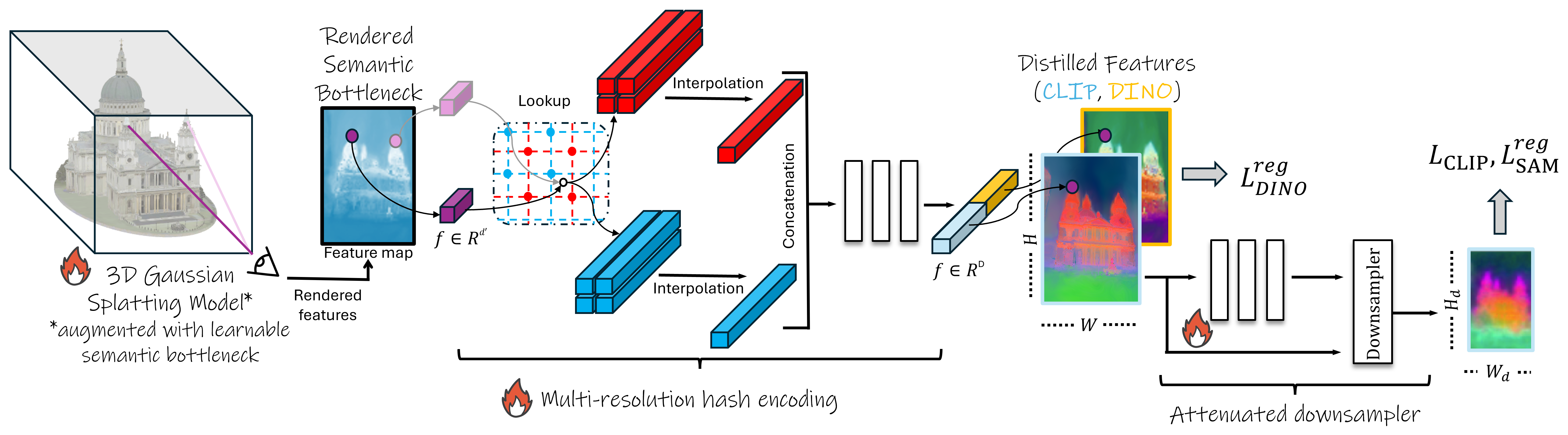}}{}
\vspace{-0.3cm}
\caption{\rev{\textbf{Method Overview}. 
We first augment a 3D Gaussian Splatting model with a learnable low-dimensional semantic bottleneck features. We then render these semantic bottleneck features---each low-dimensional feature $\mathbf{m}_{u,v} \in \mathbb{R}^{d'}$ in this map (visualized above) serves as an input coordinate to a multiresolution hash grid (we show two resolutions here in \textcolor{red}{red} or \textcolor{blue}{blue}), meaning that two similar features (as shown) are mapped to the same location. 
To leverage feature similarity, the output feature vector is obtained by \textit{linearly interpolating} between the learnable feature vectors stored at grid points neighboring $\mathbf{m}_{u,v}$ in the $d'$-dimensional feature space (following their lookup).
The resulting features are then concatenated and passed through a small, shallow MLP, $\mathcal{G}$, which outputs the predicted high-dimensional CLIP and DINOv2 features (depicted in light blue and yellow). To mitigate semantic misalignments, %
we incorporate an attenuated downsampler module. Additionally, we propose two regularization objectives ($L_\text{DINO}^{reg}$,$L_\text{SAM}^{reg}$); see Sec.~\ref{sec:misalignments} for additional details.} %
}
\label{fig:architecture}
\end{figure*}

%% file: 03-method.tex
\section{Method}

We now describe our approach for efficiently and compactly lifting language embeddings to 3D for large-scale in-the-wild scenes. We first describe our underlying compact 3D representation in Sec.~\ref{sec:representation}, resulting in rendered semantic bottleneck features. This is followed by our novel 
multi-resolution semantic hash encoding (Sec.~\ref{sec:bottlenecks}), compactly mapping low-dimensional rendered features to high-dimensional CLIP features. Next, in Sec.~\ref{sec:misalignments}, we describe our approach for bridging the semantic misalignment of ground truth 2D features. 
Lastly, in Sec.~\ref{sec:details}, we provide necessary mechanisms for handling in-the-wild Internet collections in the context of feature distillation. An illustration of our approach is shown in Fig.~\ref{fig:architecture}.

\subsection{3D Representation and Semantic Bottleneck}
\label{sec:representation}

We follow earlier approaches of representing our 3D scene using an underlying 3D Gaussian Splatting~\cite{kerbl20233d} (3DGs) together with low-dimentional features~\cite{luiten2023dynamic,li2023spacetime,lu20243d,yang2023deformable,wu20234d, sun20243dgstream, lin2024gaussian}. 
Specifically, in 3DGs, a static scene is modeled as a collection of $N$ 3D anisotropic Gaussians. Each Gaussian $G_i$ is defined by its mean (position) $\mathbf{\mu}_i \in \mathbb{R}^3$, a covariance matrix $\mathbf{\Sigma}_i \in \mathbb{R}^{3 \times 3}$ (parameterized by scaling factors $\mathbf{s}_i \in \mathbb{R}^3$ and a rotation quaternion $\mathbf{q}_i \in \mathbb{R}^4$), an opacity $\alpha_i \in [0,1]$, and coefficients for view-dependent color, typically Spherical Harmonics (SH) $\mathbf{sh}_i \in \mathbb{R}^{K \times 3}$.

To embed semantic information within this efficient framework, we augment each 3D Gaussian $G_i$ with an additional learnable \textbf{low-dimensional feature vector} $\mathbf{f}_i \in \mathbb{R}^{d'}$. Here, $d'$ is the dimensionality of our compact semantic features, where $d' \ll D$, and $D$ is the dimensionality of the original high-level semantic features from a pre-trained vision-language model. Thus, each Gaussian in our model is represented by the tuple $(\mathbf{\mu}_i, \mathbf{s}_i, \mathbf{q}_i, \alpha_i, \mathbf{sh}_i, \mathbf{f}_i)$. This low-dimensional feature vector $\mathbf{f}_i$ is optimized alongside other Gaussian attributes and is rendered via the differentiable rasterization process to produce a 2D feature map, serving as our rendered semantic bottleneck. We find that a value as small as  $d'=3$ is sufficient to produce high-quality output, enabling a highly compact corresponding 3D representation (each Gaussian's feature $\mathbf{f}_i$ is only 3-dimensional).

\subsection{Feature-Based Multi-Resolution Hash Encoding}
\label{sec:bottlenecks}
Following rasterization, we obtain a 2D feature map $\mathbf{M} \in \mathbb{R}^{H \times W \times d'}$ from a given camera viewpoint, acting as our semantic bottleneck. Each pixel $(u,v)$ in $\mathbf{M}$ thus contains a low-dimensional feature vector $\mathbf{m}_{u,v} \in \mathbb{R}^{d'}$.
This $\mathbf{m}_{u,v}$ serves as the input coordinate to our multi-resolution hash-embedder, $\mathcal{H}$. Inspired by InstantNGP \cite{M_ller_2022}, $\mathcal{H}$ comprises $L$ levels of learnable hash tables. For each level $l \in \{1, \dots, L\}$, the input $\mathbf{m}_{u,v}$ is used to query a hash table $\mathbf{T}_l$. This involves computing a spatial hash of $\mathbf{m}_{u,v}$ to obtain an initial index into $\mathbf{T}_l$. Critically, to leverage feature similarity, the output feature vector $\mathbf{e}_l(\mathbf{m}_{u,v})$ is obtained by \textbf{linearly interpolating} between the learnable feature vectors stored at grid points neighboring $\mathbf{m}_{u,v}$ in the $d'$-dimensional feature space of $\mathbf{T}_l$. This interpolation ensures that similar low-dimensional feature inputs result in similar high-dimensional embeddings. Intuitively, as large-scale scenes often contain many repetitive structures (\emph{e.g.}, such as the columns in the St. Paul Cathedral), this provides additional flexibility, which is not constrained by the position in 3D space (as in prior work), without compromising efficiency. %
The resulting features from all levels, $\mathbf{e}_l(\mathbf{m}_{u,v})$, are then concatenated to form a high-dimensional feature vector $\mathbf{E}_{u,v} = \text{Concatenate}(\mathbf{e}_1(\mathbf{m}_{u,v}), \dots, \mathbf{e}_L(\mathbf{m}_{u,v}))$. Finally, $\mathbf{E}_{u,v}$ is passed through a small, shallow MLP, $\mathcal{G}$, which outputs the predicted high-dimensional semantic features $\hat{\mathbf{S}}_{u,v}$, subsequently used for 2D supervision. $\hat{\mathbf{S}}_{u,v}$ is a concatenation of two features, one corresponding to a CLIP feature and another to a DINOv2 feature. 
\rev{Note that unlike prior work which interpolates based on spatially similar features, our framework enables low-dimensional \emph{rendering}, which is critical for GPU memory and speed, especially in large scenes; this is further demonstrated in our experiments. }

\subsection{Bridging Semantic Misalignments}
\label{sec:misalignments}
Given our 2D (higher-dimensional) feature maps $\hat{\mathbf{S}}$, we wish to compare them with ground truth pixel-aligned 2D feature maps, so as to supervise the 3DGs representation and the hash encoder.
To this end, prior work typically follows the multi-scale approach of LERF~\cite{kerr2023lerf} and extracts an image pyramid, containing image crops of different scales $s\in S$ which are then passed through the CLIP image encoder, resulting in a ground truth feature pyramid. These are then compared to rendered features through a reconstruction loss.  As observed by prior work~\cite{qin2023langsplat, ji2024fastlgsspeedinglanguageembedded, yu2024sparsegrasproboticgrasping3d} (and also illustrated in the supplementary material), this results in noticeable semantic misalignments, especially in our challenging large-scale in-the-wild setting. %
We therefore propose several novel components aimed at mitigating such misalignments.

\subsubsection{Attenuated Downsampler Module}
Rather than directly comparing rendered features with the misaligned 2D features, we propose to first render semantic features with a higher resolution than the target ground truth features (the hash encoder does not change the resolution) and subsequently aggregate the features over spatial regions using a learnable attenuated downsampler.
As shown in Sec.~\ref{sec:ablations}, Fig.~\ref{fig:ablations_attention_seg}, such adaptive weighting boosts overall performance. 

Formally, given an input feature map $\mathbf{F} \in \mathbb{R}^{H \times W \times C}$ (the component of $\hat{\mathbf{S}}$ corresponding to CLIP features), our \emph{Attenuated Downsampler Module} outputs a downsampled feature map $\mathbf{F}_d \in \mathbb{R}^{H_d \times W_d \times C}$. This $\mathbf{F}_d$ is directly compared to the 2D CLIP embedding pyramid's ground truth of the same resolution and channel dimension. To achieve this, we employ a small network consisting of three 3x3 convolutional layers, which learns to predict a scalar weight $w_i$ for each input feature vector $\mathbf{f}_i$ within a local spatial window. These weights are then used in a weighted downsampler:
\[ \mathbf{F}_d = \frac{\sum_{i \in S} \exp(w_i) \cdot \mathbf{f}_i}{\sum_{i \in S} \exp(w_i)} \]
where $w_i$ are the learned attention weights associated with each input feature vector $\mathbf{f}_i$, and $S$ is the set of all feature vectors within a local $R \times R$ spatial window in $\mathbf{F}$ that corresponds to a single output pixel in $\mathbf{F}_d$. The dimensions of this window are determined by the downsampling ratio ($R = H/H_d = W/W_d$).

\subsubsection{Regularization Objectives}
To further mitigate semantic misalignments and encourage consistent feature representations, we introduce two regularization as part of our full objective (detailed in Sec.~\ref{sec:full_objective}).

\begin{itemize}
    \item \revB{\textbf{DINO-based Regularization.}} Prior work has shown that predicting DINOv2 features (together with CLIP), enables depicting more fine-grained details. As a further step, in our setting, we (1). Ensure that a shared MLP ($\mathcal{G}$, detailed in Sec.~\ref{sec:bottlenecks}) predicts both CLIP and DINOv2 features and (2). Incorporate weight decay during training. This forces both feature maps to be coupled more tightly.
    \item \input{figures/langsplat_gt_visualization/langsplat_gt_problem}
\revB{\textbf{SAM-based Regularization.}} Several prior works~\cite{qin2023langsplat, ji2024fastlgsspeedinglanguageembedded, zhou2022extractfreedenselabels, wu2024opengaussianpointlevel3dgaussianbased} leverage Segment Anything (SAM)~\cite{kirillov2023segany} generated segmentation masks to constrain 2D feature maps to specified regions. While effective when objects are well separated, it is unsuitable for complex in-the-wild scenes where object boundaries are highly entangled, as is the case in our setting (see Fig.~\ref{fig:langsplat_gt_problem}).

Instead, we propose employing SAM masks as a regularization objective to enforce intra-object feature consistency. For each rendered feature map $\mathbf{F}_d$, we compute the variance of CLIP features within each SAM generated mask and penalize high intra-object variance. This encourages consistent feature representations for semantic objects. The SAM regularization loss for a single image is:
\begin{equation}
L_{\text{SAM}} = \sum_{M_i \in \text{Masks}} \text{Var} \left[ \mathbf{F}_d(\mathbf{M}_i > 0) \right] \label{eq:reg_sam}
\end{equation}
where $\mathbf{M}_i$ is the binary SAM mask for object $i$. This effectively promotes feature uniformity within individual elements in structured scenes, avoiding excessive averaging across distinct structures.
\end{itemize}

\subsubsection{Full Objective}
\label{sec:full_objective}

Our overall training objective aims to reconstruct ground truth RGB views and minimize intra-object variations in the feature space. The full objective is:
$$L = L_\text{rec} + \lambda_\text{CLIP}L_\text{CLIP} + \lambda_\text{DINO} L_\text{DINO} + \lambda_\text{SAM} L_\text{SAM}$$
where $L_\text{rec}$ is the standard 3DGs reconstruction loss on RGB rendered views (see Sec.~\ref{sec:details}), $L_\text{CLIP}$ is the CLIP-based supervision loss is the L1 loss between downsampled features $\mathbf{F}_d$ and corresponding ground truth CLIP features (averaged over the different scales, see Sec.~\ref{sec:details}). 
$L_\text{DINO}$ and $L_\text{SAM}$ are our novel regularization objectives described above (ablated in Sec.~\ref{sec:results}).  $\lambda_\text{CLIP}$, $\lambda_\text{DINO}$ and $\lambda_\text{SAM}$ are weighting coefficients (hyperparameters). \rev{Note that our method jointly optimizes the Gaussian model and language field, without using pre-existing 3D Gaussians. }%
Additional training and implementation details are provided in the supplementary.

\subsection{Handling In-The-Wild Views}
\label{sec:details}

We employ several additional strategies to ensure we effectively handle large-scale in-the-wild scenes. Further details are provided in the supplementary materials.

\subsubsection{\revB{Gaussian Splatting In-The-Wild Adaptation}}
Our dataset's ``in-the-wild" characteristics, which include diverse cameras, varying conditions (e.g., varying illumination, weather, time of day), and transient objects (e.g., people, vehicles, trees), pose significant challenges for high-quality 3DGs. To robustly address these, we integrate the SWAG method~\cite{dahmani2025swag} into our pipeline. This component affects our photometric loss $L_{rec}$ only and is detailed further in the supplementary material. 

\subsubsection{\revB{Enhancing the query prompts}}
We use an LLM to enhance the quality of prompts, asking it to output the prompt synonyms that are more common linguistically. %

\subsubsection{\revB{Adjustable Scale CLIP Pyramids}}
Our in-the-wild scenes exhibit significant scale variation across camera parameters, rendering standard CLIP pyramid scales (typically image resolution-dependent~\cite{kerr2023lerf}) impractical. Instead, we define \emph{physical scales} consistent across all images. %
We achieve this by estimating the average physical pixel size from COLMAP metadata (details in the supplementary material) and computing a custom CLIP pyramid scale for each image based on its individual pixel size.
Additionally, we employ the finetuned CLIP model proposed in HaLo-NeRF~\cite{dudai2024halo}, which is better adapted for the architectural domain.

%% file: figures/langsplat_gt_visualization/langsplat_gt_problem.tex
\begin{figure}
    \centering
    \hbox{
        \hbox{            \includegraphics[height=0.182\textwidth]{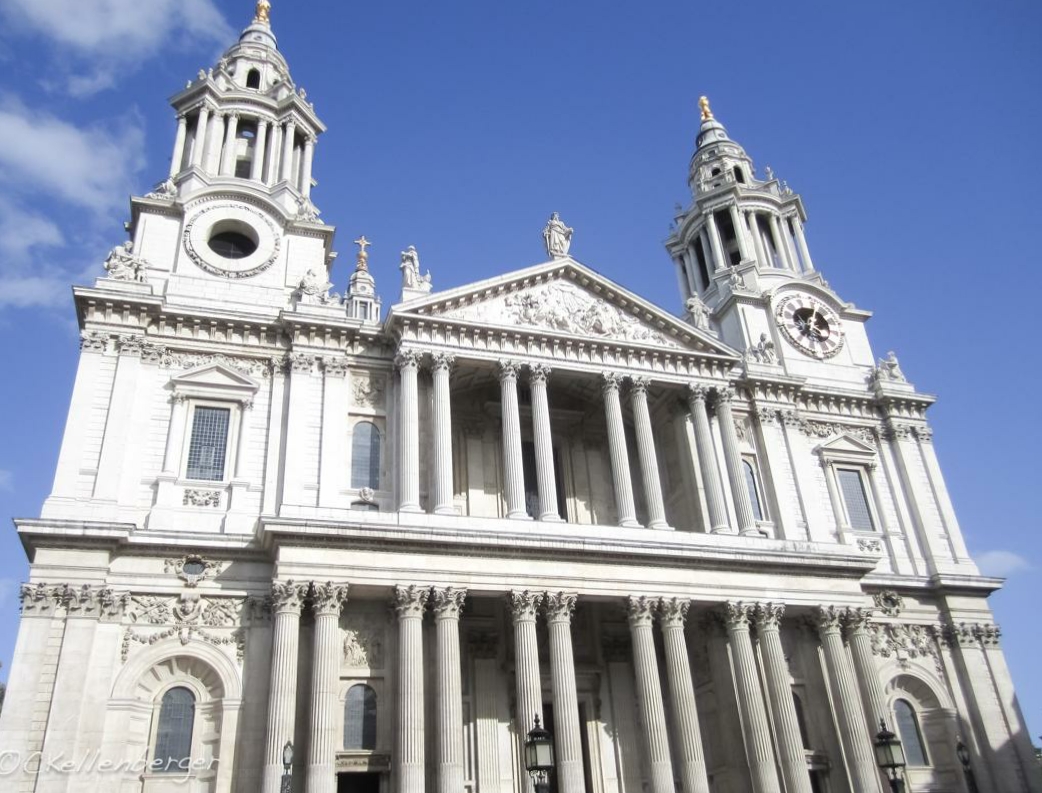}
            
        }

        \vbox{
            \hbox{
                \fcolorbox{red}{red}{\includegraphics[width=0.050\textwidth]{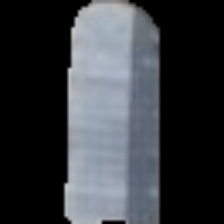}}

                \fcolorbox{red}{red}{\includegraphics[width=0.050\textwidth]{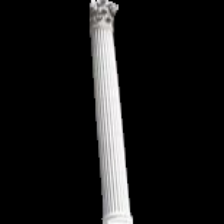}}
                
                \fcolorbox{red}{red}{\includegraphics[width=0.050\textwidth]{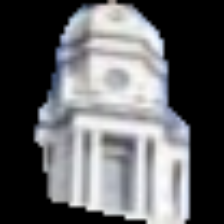}}

            }
            \hbox{
                \fcolorbox{red}{red}{\includegraphics[width=0.050\textwidth]{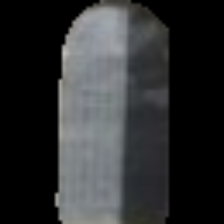}}

                \fcolorbox{white}{white}{\includegraphics[width=0.050\textwidth]{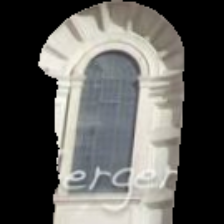}}

                \fcolorbox{white}{white}{\includegraphics[width=0.050\textwidth]{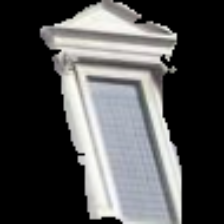}}
            }
            \hbox{
                \fcolorbox{green}{green}{\includegraphics[width=0.050\textwidth]{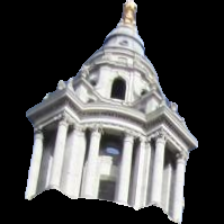}}

                \fcolorbox{red}{red}{\includegraphics[width=0.050\textwidth]{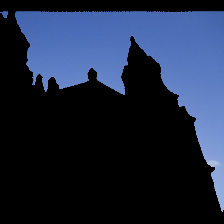}}

                \fcolorbox{green}{green}{\includegraphics[width=0.050\textwidth]{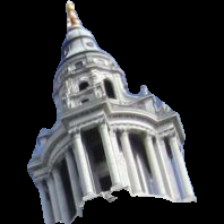}}
        
            }
        }
    }

    \caption{
    \textbf{Motivation for our SAM-based Regularization.} 
    For the input view on the left, previous works, such as LangSplat \cite{qin2023langsplat}, constrain 2D CLIP feature maps to masked regions produced by SAM, such as those shown on the right. However, CLIP embeddings produced on these masked regions are often misclassified; as visualized above using red boxes. In the example, seven bounded images are detected as \emph{Towers}, but five are incorrectly classified as such. Our novel SAM-based regularization  (Eq.~\ref{eq:reg_sam}) avoids this by only softly constraining the variance of CLIP features within a SAM-generated masked region. 
    }
    \label{fig:langsplat_gt_problem}
\end{figure}

%% file: 04-results.tex
\section{Experiments}
\label{sec:results}

We evaluate our method, comparing performance against segmentation-based and feature-based baselines.
We first describe our experimental setup and baselines in Sec.~\ref{sec:baselines}. We then present quantitative and qualitative comparisons in Sec.~\ref{sec:comparisons}. We analyze the contribution of individual components of our method through ablation studies in Sec.~\ref{sec:ablations}. Lastly, we discuss the limitations of our approach in Sec.~\ref{sec:limitations}.

\subsection{Experimental Setup}
\label{sec:baselines}

\subsubsection{\revB{Data}}
We consider the HolyScenes~\cite{dudai2024halo} dataset, which contains large-scale in-the-wild scenes of religious landmarks, captured from various viewpoints, lighting, and weather conditions. %
\rev{This dataset contains scenes with over 2000 images, varying resolutions, and large physical dimensions (\emph{e.g.}, Badshahi Mosque spans 25,600 square meters).} %
\rev{We additionally show results on indoor scenes from Lerf-OVS~\cite{kerr2023lerf} and DL3DV-1K~\cite{ling2023dl3dv10klargescalescenedataset} in the supplementary material.}

\subsubsection{\revB{Metrics}}
We follow prior work~\cite{dudai2024halonerf} and report the Average Precision (AP) across different semantic categories, enabling us to evaluate  
both the precision of segmentation boundaries and the model's ability to correctly identify different semantic regions across architectural landmarks.

\subsubsection{\revB{Baselines}}
We compare our method against: (1) 3D feature-based approaches that embed language features for open-vocabulary querying, and (2) Segmentation-based methods that extract per-prompt segmentation maps.

For segmentation-based methods, we compare against one 3D-based segmentation approach, HaLo-NeRF~\cite{dudai2024halo}, and the 2D methods they compare against in their work: CLIPSeg~\cite{luddecke2022image}, LSeg~\cite{li2022language}, and ToB~\cite{wu2021towers} (a categorical baseline, and hence some results are missing). Two results are reported for CLIPSeg, one for the pretrained model, and the other (CLIPSeg$_\text{FT}$) is the fine-tuned segmentation component from HaLo-NeRF that adapts CLIPSeg to architectural domains using multi-view correspondences. We additionally compare against LangSAM, which combines SAM~\cite{kirillov2023segment} with GroundingDINO~\cite{liu2024grounding} detection to generate masks for text-prompted objects through instance segmentation.

For 3D feature-based methods, we compare against  DFF~\cite{kobayashi2022decomposing} and LERF~\cite{kerr2023lerf}, reporting the performance obtained using the Ha-NeRF backbone, as evaluated in HaLo-NeRF. %
\rev{We also compare against three additional methods: LangSplat~\cite{qin2023langsplat}, Feature3DGS~\cite{zhou2023feature} and
FMGS~\cite{zuo2024fmgsfoundationmodelembedded}. We adapt these methods to enable optimization on large-scale scenes. For example, Feature3DGS uses our semantic features and in-the-wild modifications; additional details are provided in the supplementary material. }
\input{tables/results}

\input{figures/final_results/results_feature_based_visualization}

\subsection{Comparison to Baselines}
\label{sec:comparisons}

\subsubsection{\revB{Quantitative Evaluation}}

A quantitative evaluation is provided in 
Tab.~\ref{fig:results_table}. \rev{These results are obtained over a subset of input views, following the evaluation protocol reported in HaLo-NeRF. As further detailed in the supplementary material, we also extract a small validation set, to additionally validate performance on novel views, finding that our method maintains consistent performance over novel views (achieving mAP 0.60 compared to 0.59).}

Compared to the 3D segmentation-based approach of HaLo-NeRF, we obtain comparable results while being orders of magnitude faster (2 hours vs. >0.1 seconds per text prompt for ours). While our overall mAP is slightly lower, we achieve improved performance over some of the semantic categories, specifically over \emph{Minarets} and \emph{Spires}, which can be considered more specialized architectural terminology, demonstrating our method's strong performance over unique and more esoteric vocabulary. Among the 2D segmentation methods, CLIPSeg$_{\text{FT}}$ performs best, demonstrating the value of domain adaptation for architectural scenes. LangSAM and ToB achieve low scores, while LSeg performs worst.

Among feature-based methods, our approach substantially outperforms all baselines, illustrating that feature distillation in our problem setting cannot be achieved with existing methods. \rev{In particular, we observe that even with our \emph{Semantic Hash Encoder}, Feature3DGS yields extremely poor performance in these scenes (yielding a mAP score of 0.05 vs. our 0.59). This highlights the importance of our \emph{Attenuated Downsampler Module} and proposed regularizations, but most importantly, it indicates that their 2D features lack sufficient generalization. LangSplat also struggles to localize features due to its strong reliance on SAM masks, as also demonstrated in Fig.~\ref{fig:langsplat_gt_problem}. FMGS, which also uses multi-resolution hash encoding, yields a smaller performance gap (mAP of 0.51 vs. our 0.59). As can be observed, our architectural modifications (e.g., our \emph{Attenuated Downsampler Module}) and proposed regularizations allow for achieving improved performance. Furthermore, we note that its embedding module, unlike our \emph{Semantic Hash Encoder}, operates prior to rendering, requiring the full feature dimensionality to be rendered---yielding significantly longer and more memory-intensive training times (as can be further seen in Table~\ref{tab:ablations_encoding_tabel}).}

\subsubsection{\revB{Qualitative Evaluation}}
Fig.~\ref{fig:comp_halo} provides a comparison to the \textit{3D segmentation} based approach HaLo-NeRF which is orders of magnitude slower at inference (for a given text). HaLo-NeRF often segments more regions in comparison to our method, but does so in a less accurate manner. For instance, for \emph{Domes}, Halo-NeRF correctly predicts three domes, but unlike our method, also segments other regions, \rev{or alternatively, it fails to detect all the \emph{Spires} regions.}

Fig.~\ref{fig:feature_based_result_table} provides a comparison \rev{with \textit{feature} based methods. Feature3DGS produces segmentations over the whole building, failing to accurately localize architectural elements. This stems from LSeg features not being discriminative enough for distinguishing between different semantic regions in large architectural buildings (as also illustrated by the poor results of LSeg itself; see Tab.~\ref{fig:results_table}). LangSplat and FMGS yield more fine-grained segmentations due to the usage of CLIP features, but are still are not competitive with our method.}

Fig.~\ref{fig:result_table} provides a qualitative comparison to the \textit{2D segmentation}-based approach, LangSAM, across different views of the same scene for dome segmentation. As a 2D method, LangSAM produces inconsistent results across different viewpoints. While it succeeds in assigning higher probabilities to dome regions in some cases, its performance varies significantly between views of the same architectural elements. Our 3D feature distillation approach yields  consistent results across all viewpoints, \rev{as also illustrated over novel views in the supplementary material.}

\input{figures/ablation_results/encoding_ablation/encoding_ablation_table}

\input{figures/ablation_results/simultaneous_optimisation/ablation_table_2}

\input{figures/final_results/encoders_ablation_visualization}

\subsection{Ablation Studies}
\label{sec:ablations}

Our ablations are grouped into two: (1). semantic bottleneck components, and (2). semantic distillation mechanisms. As illustrated next, these studies validate the importance of each proposed component in achieving robust language embedding for large-scale scenes.

\subsubsection{\revB{Semantic Bottleneck Ablations}}
We evaluate our proposed semantic bottleneck features (which are rendered and processed through our semantic hash encoder) %
against alternative feature embedding methods. Specifically, \rev{while keeping all other aspects of our method the same, we replace only our feature embedding method} %
with (i) the Speedup CNN encoder from Feature3DGS~\cite{zhou2023feature}  and (ii) the hash encoder from Foundation Model Embedded 3D Gaussian
(FMGS)~\cite{zuo2024fmgsfoundationmodelembedded}, \rev{which interpolates spatially}. 
To facilitate a fair comparison that enables running all these alternatives under a similar experimental setup, we reduce the resolution of the images (using the alternatives on the original resolution cannot fit in memory). We conduct this ablation over the Milano Cathedral, reporting both the reconstruction loss over held-out views not used in training (\emph{i.e.}, comparing the distilled features to the ground truth 2D signal) and efficiency. As illustrated in Tab.~\ref{tab:ablations_encoding_tabel}, \rev{our semantic bottleneck consistently outperforms both alternatives while providing significant computational advantages, demonstrating the effectiveness of feature space interpolation}. Fig.~\ref{fig:encoding_ablation_table} provides a qualitative comparison, demonstrating that our low-dimensional semantic bottleneck enable achieving better segmentation quality compared to alternative feature embedding methods.

\subsubsection{\revB{Semantic Distillation Ablations}}
We ablate components designed to address the localization challenges inherent in 2D CLIP pyramid features. We consider three scenes (St. Paul, Milano, and Blue Mosque), focusing on the windows category, which often appears in multiple distinct regions.
and therefore highlights the effect of each component in isolation. %
We consider: 
(i) replacing our Attenuated Downsampler Module with bilinear interpolation (w/o AD Module (LR)) or direct CLIP feature interpolation as in LERF (w/o AD Module (HR)) , (ii) removal of DINO regularization loss (w/o DINO Reg.), (iii) removal of SAM regularization loss  (w/o SAM Reg.), and (iv) using standard LERF pyramids instead of our physical scale pyramids  (w/o Phys. Scale).

Tab.~\ref{fig:ablations_simultaneous_tabel_2} and Fig.~\ref{fig:ablations_attention_seg} demonstrate the importance of each component. The Attenuated Downsampler Module provides substantial improvements, 
enabling adaptive weighting of semantically relevant regions during downsampling. The visual results in Fig.~\ref{fig:ablations_attention_seg} clearly show that without the AD Module, segmentations become noisy and imprecise, particularly around architectural boundaries. DINO regularization proves essential across all scenes.
The SAM regularization provides consistent but smaller improvements, helping maintain feature consistency within architectural elements. Our physical scale pyramids show improvements over standard LERF pyramids, with the Blue Mosque scene showing the largest gains due to its complex multi-scale architectural features. %

\subsection{Limitations}
\label{sec:limitations}
Our method enables learning efficient language embeddings for large-scale scenes with interactive inference times. However, there are several limitations to consider. First, our reliance on %
multi-scale CLIP features introduces spatial ambiguities that are challenging to resolve completely. 
Future advances in 2D localization techniques could potentially improve our results by providing more spatially precise features. 
Second, the significant variance in architectural element sizes necessitates multiple pyramid scales, where large-scale features can dominate and obscure smaller architectural details during distillation. 
Different scales capture distinct levels of architectural detail, and our simple averaging may lose fine-grained semantic information. 
This is illustrated in Fig. \ref{fig:limitations} (right), where the smaller windows are partially missed by our approach.
Future research could explore more sophisticated scale fusion strategies, such as attention-based mechanisms or scale-specific decoders.
Third, our relevancy score computation, which follows the mechanism introduced in LERF, often struggles to differentiate between semantically related regions, such as the decorative openings and the windows illustrated in Figure \ref{fig:limitations} (left). 
Future work could resolve such fine-grained differences using scores that integrate context-dependent negative text-prompt selection (instead of simply using \emph{Objects}, \emph{Things}, etc.) or additional learned mechanisms.

\input{figures/final_results/limitation_vizualization}

%% file: tables/results.tex
\begin{table} %

\centering

\caption{\textbf{Quantitative Evaluation over the HolyScenes dataset.} We report per-category average precision, comparing our results to 2D segmentation-based methods (top), 2D segmentation-based method of HaLo-NeRF (middle) and feature-based methods (bottom). Note that all the segmentation methods operate on 2D images, \emph{except} HaLo-NeRF, which performs optimization per text prompt (yielding a runtime of $\sim$2 hours).}

\setlength{\tabcolsep}{2.4pt}

\begin{tabular}{@{ } ll cccccc c @{}}
    \hline
    &\hspace{-7pt}Category & \footnotesize{Win.} & \footnotesize{Minaret} & \footnotesize{Dome} &  \footnotesize{Tower} &  \footnotesize{Spire} & \footnotesize{Portal} & \footnotesize{mAP}  \\
    \hline
    \multicolumn{9}{l}{\hspace{-0pt}\textbf{2D Seg. Based (3D Inconsistent)}} \\ 
  &LSeg & 0.13 & 0.06 & 0.05 & 0.19 & 0.06 & 0.05 & 0.09 \\
    &ToB & 0.04 & x & x & 0.49 & x & 0.15 & 0.23 \\
    &CLIPSeg & 0.44 & 0.63 & 0.69 & \textbf{0.87} & 0.46 & 0.29 & 0.56 \\
    &CLIPSeg$_{\text{FT}}$ & 0.51  & 0.81 & \textbf{0.77} & \textbf{0.87} & 0.50 & \textbf{0.49} & 0.66 \\
    &LangSAM & 0.53  & 0.14 & 0.44 & 0.19 & 0.11 & 0.05 & 0.25 \\
    \hline
        \multicolumn{9}{l}{\hspace{-0pt}\textbf{3D Seg. Based (Slow)}} \\ 
    &HaLo-NeRF (\underline{\emph{2h}}/text) & \textbf{0.64}  & 0.80 & 0.74 & \textbf{0.87} & 0.61 & 0.45 & \textbf{0.68} \\
    \hline
           \multicolumn{9}{l}{\hspace{-0pt}\textbf{Feature Based (Fast and 3D Consistent)}} \\ 
   &DFF\textdagger & 0.04  & 0.17 & 0.11 & 0.17 & 0.09 & 0.06 & 0.11  \\
    & LERF\textdagger & 0.15 & 0.09 & 0.10 & 0.13 & 0.18 & 0.16 & 0.14 \\
    & \rev{LangSplat} & 0.03 & 0.22 & 0.25 & 0.38 & 0.20 & 0.08 & 0.19 \\
    & Feature3DGS$^*$ & 0.03  & 0.04 & 0.03 & 0.14 & 0.05 & 0.01 & 0.05 \\
    & \rev{FmGS$^*$} & 0.43 & 0.82 & 0.57 & 0.49 & 0.63 & 0.10 & 0.51 \\
    & \rev{Ours} & 0.51  & \textbf{0.84} & 0.70 & 0.46 & \textbf{0.67} & 0.38 & 0.59 \\
\hline
\end{tabular}

\begin{flushleft}
{\footnotesize
\textdagger{Adapted with a Ha-NeRF~\cite{chen2022hallucinated} backbone, as reported in HaLo-NeRF}. \\
$^*$Adapted architectures, as further detailed in the supplementary material.}
\end{flushleft}

\label{fig:results_table}
\end{table}

%% file: figures/final_results/results_feature_based_visualization.tex
\begin{figure}

    \rotatebox{90}{\whitetxt{xxp}\footnotesize{Feature3DGS}}
    \jsubfig{\includegraphics[height=2.0cm]{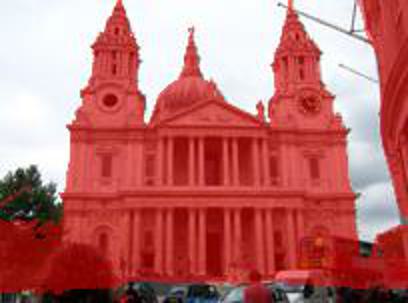}
    \includegraphics[height=2.0cm]{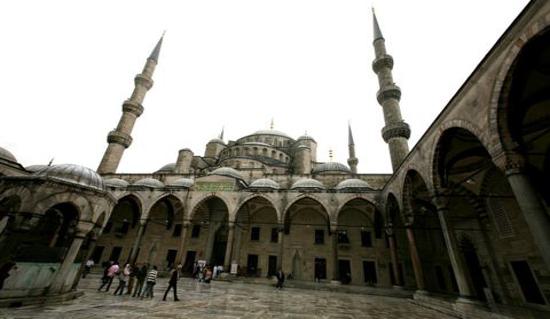}
    \includegraphics[height=2.0cm]{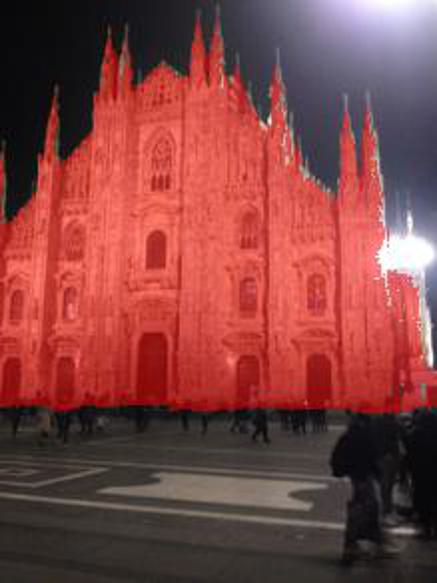}}{}
    \\
    \rotatebox{90}{\whitetxt{xxp}\footnotesize{\rev{LangSplat}}}
    \jsubfig{\includegraphics[height=2.0cm]{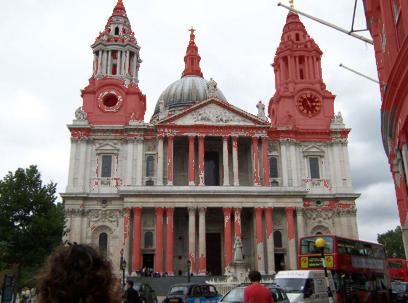}
    \includegraphics[height=2.0cm]{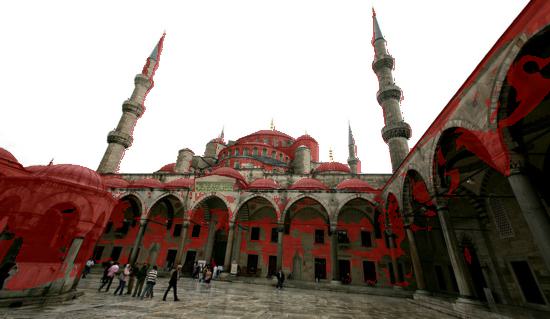}
    \includegraphics[height=2.0cm]{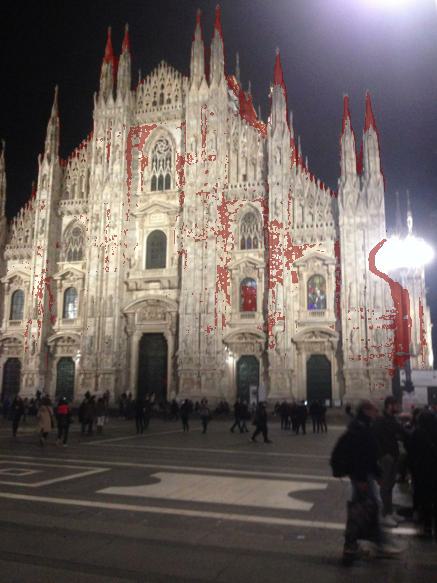}}{}
    \\
    \rotatebox{90}{\whitetxt{xxxp}\footnotesize{\rev{FMGS}}}
    \jsubfig{\includegraphics[height=2.0cm]{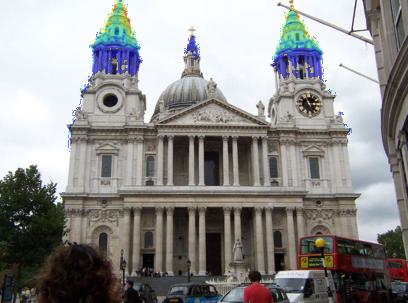}
    \includegraphics[height=2.0cm]{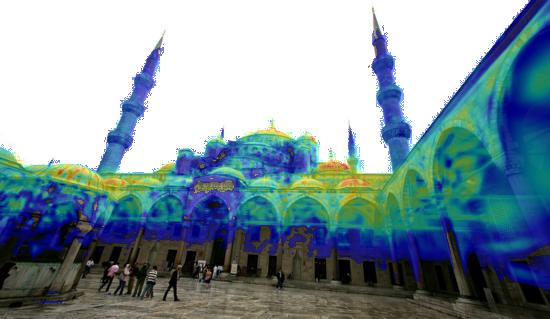}
    \includegraphics[height=2.0cm]{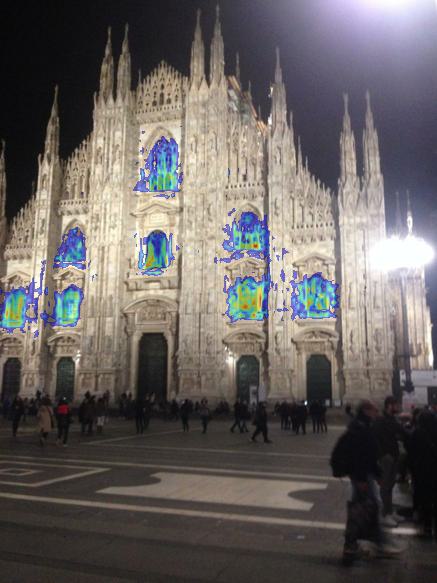}}{}
    \\
    \rotatebox{90}{\whitetxt{xp}\footnotesize{\methodname{}}}
    \jsubfig{\includegraphics[height=2.0cm]{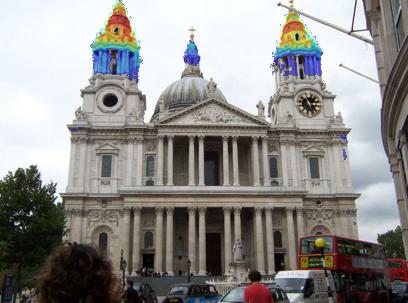}}{\textit{Towers}}
    \jsubfig{\includegraphics[height=2.0cm]{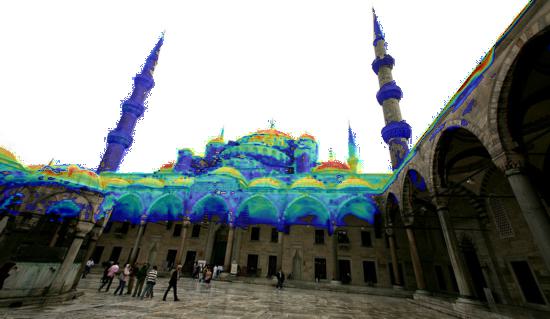}}{\textit{Domes}}
     \jsubfig{\includegraphics[height=2.0cm]{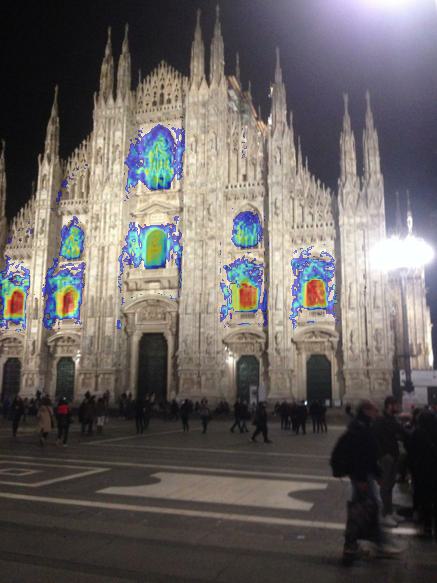}}{\textit{Windows}}
    \caption{
      \textbf{Qualitative Comparison to Feature-based Methods}.  We compare our method (bottom row) against Feature3DGS (top row), LangSplat (second row) and FMGS (third row) across three different semantic concepts. %
      }
      \label{fig:feature_based_result_table}

\end{figure}

%% file: figures/ablation_results/encoding_ablation/encoding_ablation_table.tex
\begin{table}[t] %

\centering

\caption{\textbf{Semantic Bottleneck Ablations}, comparing our semantic bottleneck features against alternative feature embedding methods: %
Feature3DGS's Speedup CNN encoder and FMGS's hash encoder. The first two metrics measure reconstruction quality, while the remaining three metrics evaluate computational efficiency. \rev{Note that the time measurements refer to a single image.}} 

\resizebox{0.95\linewidth}{!}{\begin{tabular}{lccc}
    \toprule   
    & $\text{Feature3DGS}$ & $\text{FMGS}$ & $\text{Ours}$ \\
    \midrule
      \textbf{Quality} \\
    \hspace{7pt}$\text{L1}\downarrow$ & 0.149 & 0.180 & \textbf{0.146} \\
    \hspace{7pt}$\text{PSNR}\uparrow$ & 13.3 & 11.6 & \textbf{13.9} \\
    \midrule
    \textbf{Efficiency} \\
     \hspace{7pt}$\text{Training time (sec)}\downarrow$ & 0.21 & 0.49 & \textbf{0.10} \\
     \hspace{7pt}$\text{Inference time (sec)}\downarrow$ & 0.04 & 0.07 & \textbf{0.026} \\
     \hspace{7pt}$\text{Train memory}\downarrow$ & \textasciitilde 15GB & \textasciitilde 15GB & \textasciitilde \textbf{10GB} \\
    \bottomrule
\end{tabular}}

\label{tab:ablations_encoding_tabel}
\end{table}

%% file: figures/ablation_results/simultaneous_optimisation/ablation_table_2.tex
\begin{table}[t] %
\centering

\caption{\textbf{Semantic Distillation Ablations,} ablating the use of each component in our semantic distillation pipeline:  Attenuated Downsampler Module with low resolution (w/o AD (LR)) and high resolution (w/o AD (HR)), DINO regularization (w/o Dino), SAM regularization (w/o SAM), and physical scale pyramids (w/o Physical Scale). Results over AP (average precision) are shown across three scenes (Milano, Blue Mosque, St. Paul) for the windows category, with mean Average Precision (mAP) reported across the scenes. %
}

\resizebox{0.99\linewidth}{!}{\begin{tabular}{lccccc}
        \toprule
        Category & \shortstack{Milano} & \shortstack{Blue Mosque} & \shortstack{St. Paul}  & \shortstack{mAP}  \\
        \midrule
        $\text{w/o AD (LR)}$ & 0.79 & 0.36 & 0.43 & 0.53 \\
        $\text{w/o AD (HR)}$ & 0.74 & 0.27 & 0.38 & 0.46 \\
        $\text{w/o DINO}$ & 0.47 & 0.18 & 0.26 & 0.30 \\
        $\text{w/o SAM}$ & 0.86 & 0.39 & 0.54 & 0.60 \\
        $\text{w/o Physical Scale}$ & \textbf{0.91} & 0.33 & 0.67 & 0.64 \\
        $\text{Ours}$ & 0.89 & \textbf{0.44} & \textbf{0.75} & \textbf{0.69} \\
        \bottomrule
    \end{tabular}}

\label{fig:ablations_simultaneous_tabel_2}
\end{table}

%% file: figures/final_results/encoders_ablation_visualization.tex
\begin{figure}[t]
    \centering
    \setlength{\tabcolsep}{2pt}
    \renewcommand{\arraystretch}{1.0}

    \begin{tabular}{c c c c}
        \rotatebox{90}{\whitetxt{Fgp}\footnotesize{Feature3DGS}} &
        \includegraphics[height=2.2cm]{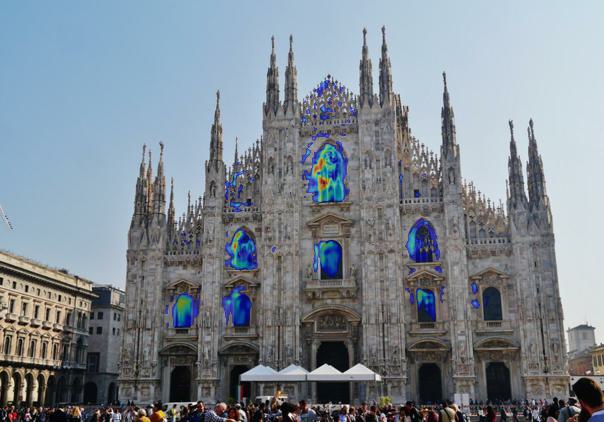} &
        \includegraphics[height=2.2cm]{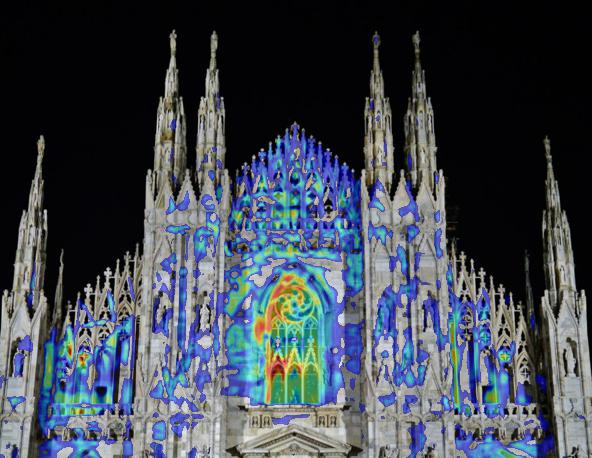} &
        \includegraphics[height=2.2cm]{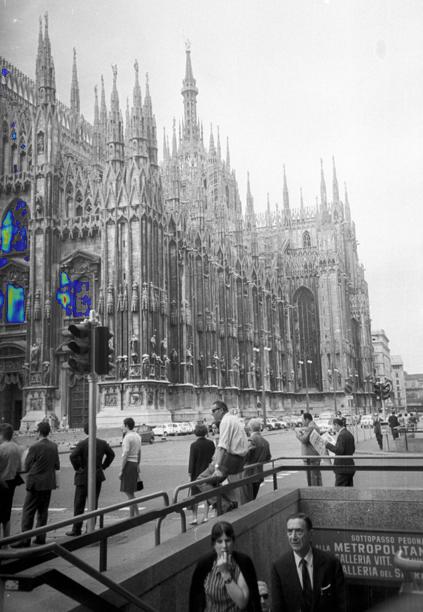} \\
        
        \rotatebox{90}{\whitetxt{xxFgp}\footnotesize{FMGS}} &
        \includegraphics[height=2.2cm]{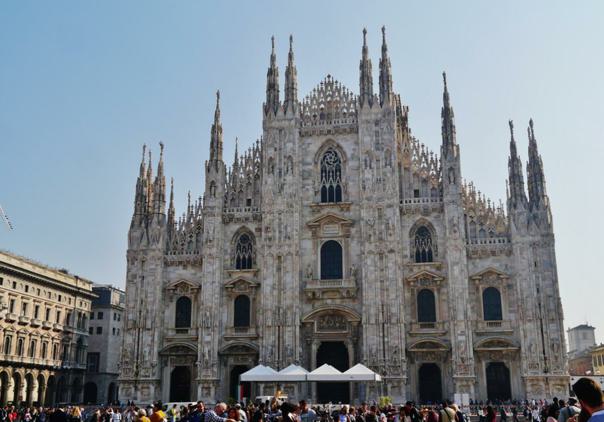} &
        \includegraphics[height=2.2cm]{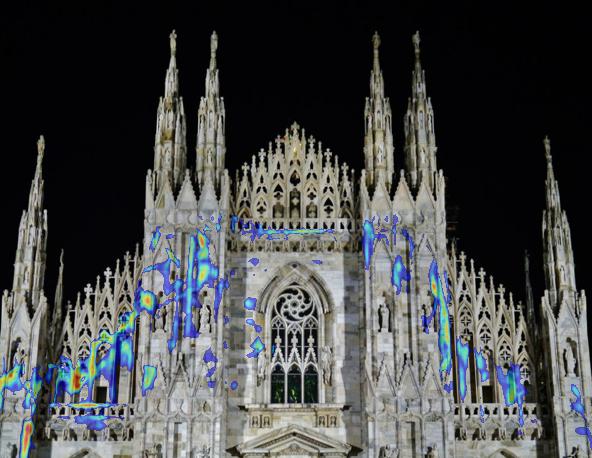} &
        \includegraphics[height=2.2cm]{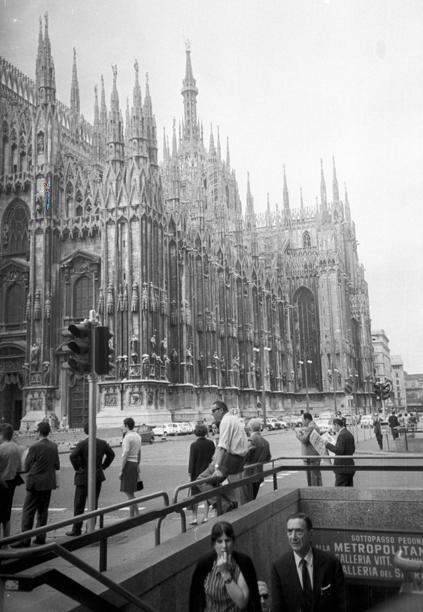} \\
        
        \rotatebox{90}{\whitetxt{Fgp}\footnotesize{Lang3D-XL}} &
        \includegraphics[height=2.2cm]{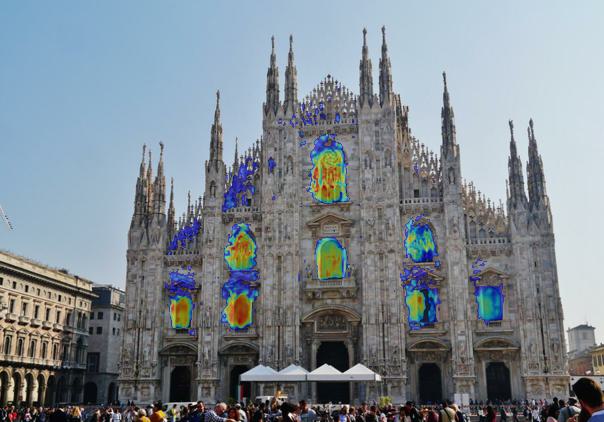} &
        \includegraphics[height=2.2cm]{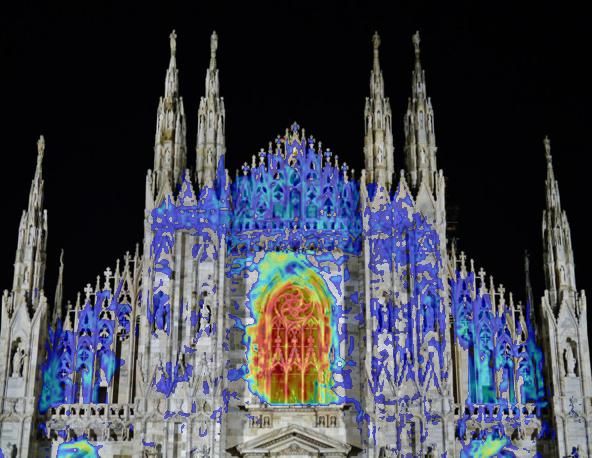} &
        \includegraphics[height=2.2cm]{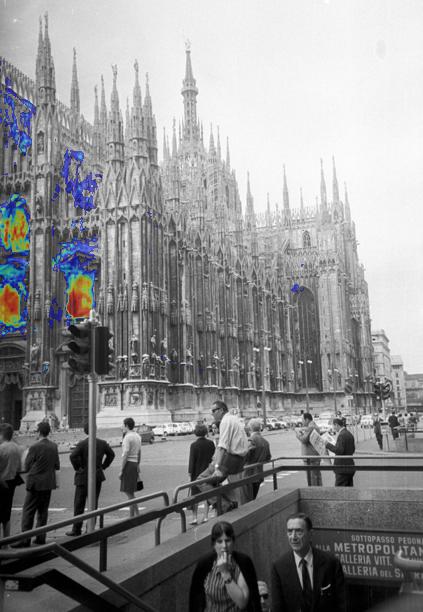} \\
    \end{tabular}
\caption{%
We visualize the effect of the distilled features, comparing our semantic bottleneck approach (bottom) against Feature3DGS's %
encoder (top) and FMGS's hash encoder (middle) on the \emph{Windows} prompt, demonstrating that our low-dimensional semantic bottleneck produces significantly cleaner and more accurate segmentations compared to alternatives. }
\label{fig:encoding_ablation_table}
\end{figure}

%% file: figures/final_results/limitation_vizualization.tex
\begin{figure}[t]
    \centering
    \jsubfig{
     \includegraphics[height=0.24\textwidth]{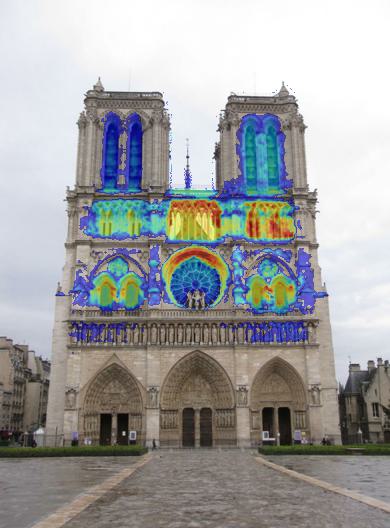}   \hfill
     \includegraphics[height=0.24\textwidth]{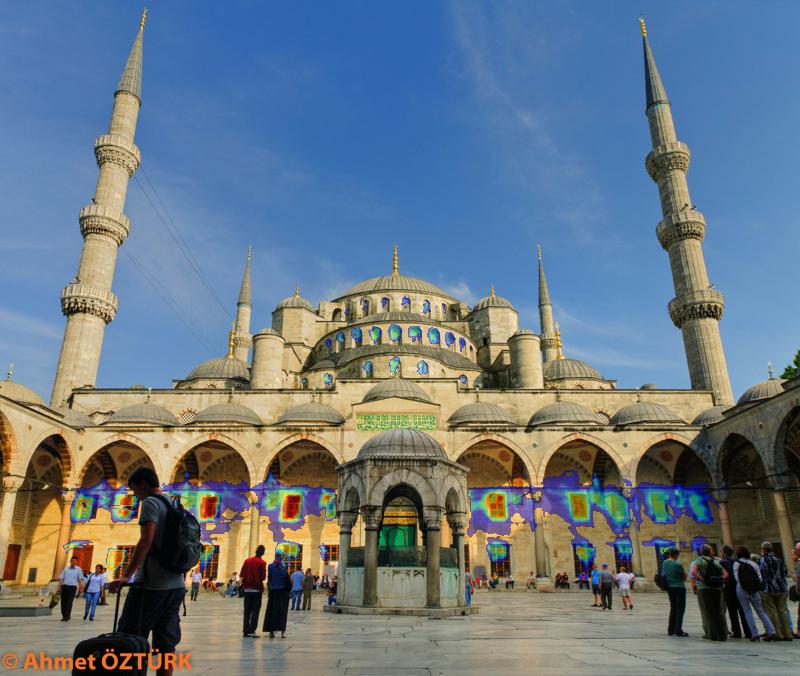}
    }{}

    \caption{Limitation examples, illustrated over queried \emph{Windows}. As shown above, our method may detect semantically similar regions, such as the decorative openings in the middle of the Notre Dame Cathedral (left), which yield higher probabilities than the windows located below and above it. Additionally, very small regions, such as the windows on top of the Blue Mosque (right) may be partially missed by our distillation technique that uses  CLIP embeddings averaged over multiple scales. 
    }
    \label{fig:limitations}
\end{figure}

%% file: 05-conclusion.tex
\section{Conclusion}

We have presented an approach that augments a 3D Gaussian Splatting representation with a learnable semantic bottleneck, enabling a distillation of language features over Internet collections capturing large-scale scenes. Additionally, we introduced several mechanisms that allow for significantly enhancing localization performance on these challenging image collections. As a whole, our approach outperforms prior work in both performance and efficiency.

Our work advances the broader goal of reconstructing historically and culturally significant sites as detailed, explorable 3D models using images captured in unconstrained settings. 
This broader goal could potential unlock transformative educational experiences and a fostering of a more inclusive connection to the world's monuments and landmarks---regardless of physical location or circumstance.

%% file: figures/our_results.tex
\begin{figure*}
  \centering
  \includegraphics[width=0.98\textwidth, trim={2.6cm 7.3cm 3.8cm 3.2cm},clip]{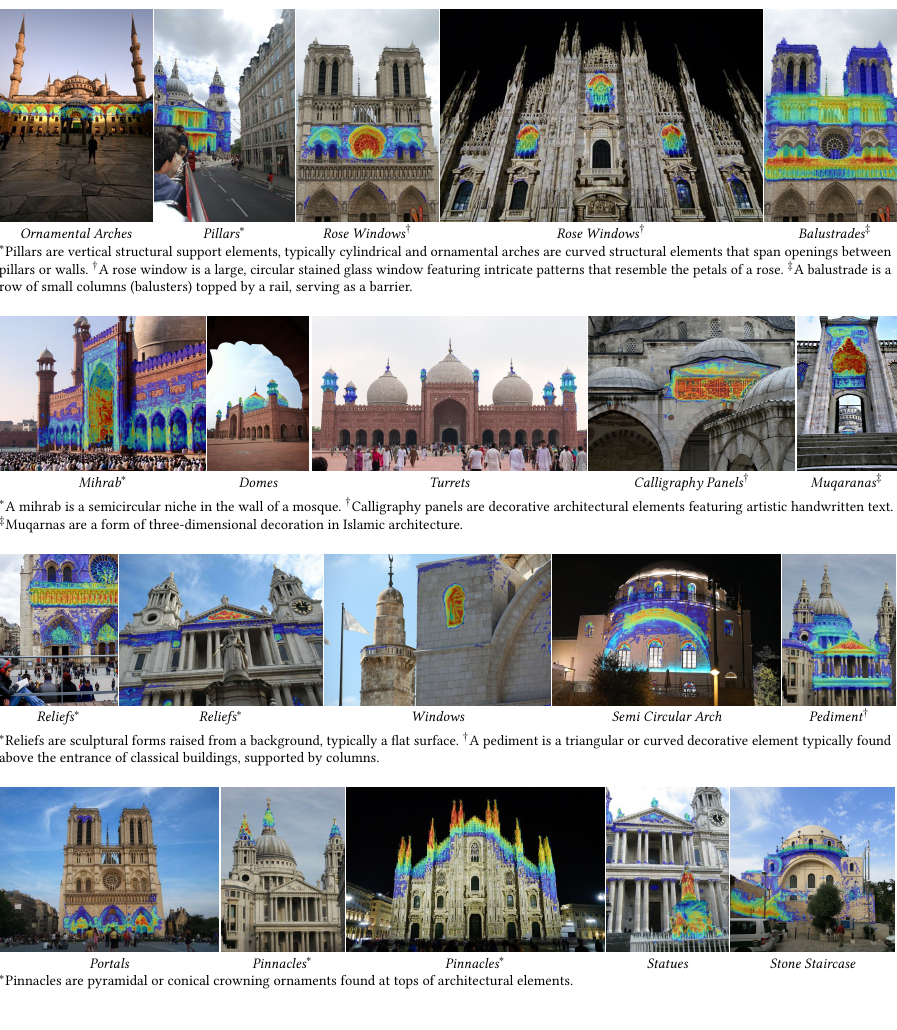}
  \caption{\textbf{3D Localization Results.} We illustrate localization results of our distillation technique over diverse architectural elements across multiple landmarks from the HolyScenes dataset. 
  In particular, our approach effectively localizes esoteric architectural terminology while maintaining precise spatial localization across varied lighting conditions, viewpoints, and architectural styles. The segmentation masks (shown in color overlays) accurately capture the boundaries and extent of each queried architectural feature, demonstrating the effectiveness of our approach for localizing semantic concepts over large-scale scenes. }
\label{fig:our_results}
\end{figure*}

%% file: figures/final_results/final_results_visualization.tex
\begin{figure*}
    \rotatebox{90}{\whitetxt{xxp}\footnotesize{HaLo-NeRF}}
    \jsubfig{\includegraphics[height=1.9cm]{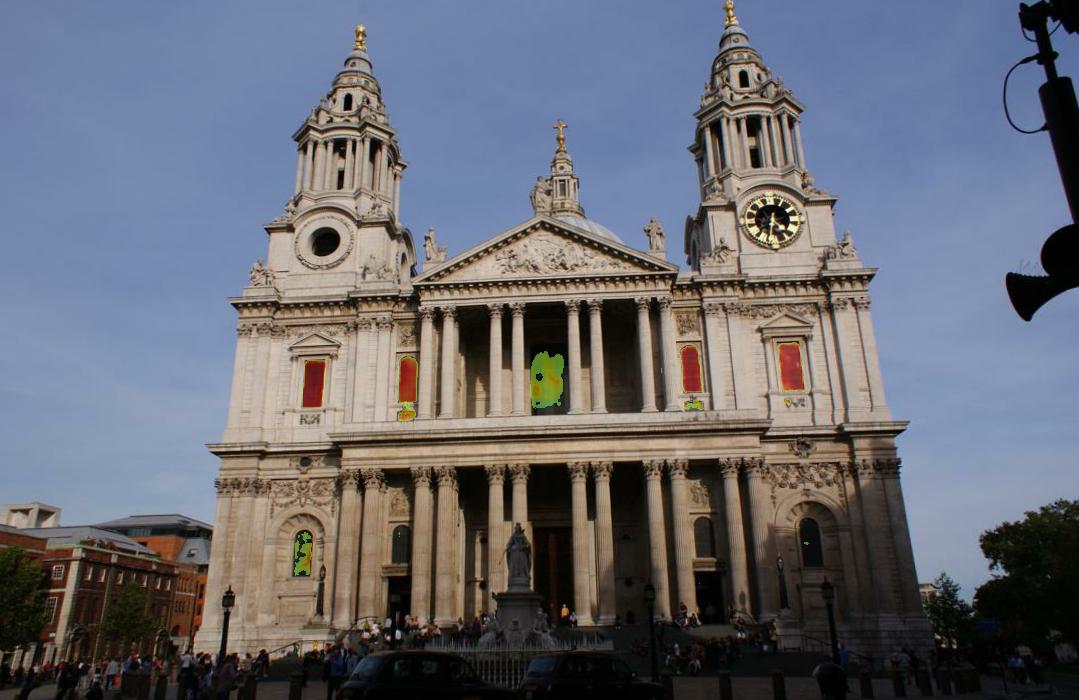}}{  }
    \hfill
    \jsubfig{\includegraphics[height=1.9cm]{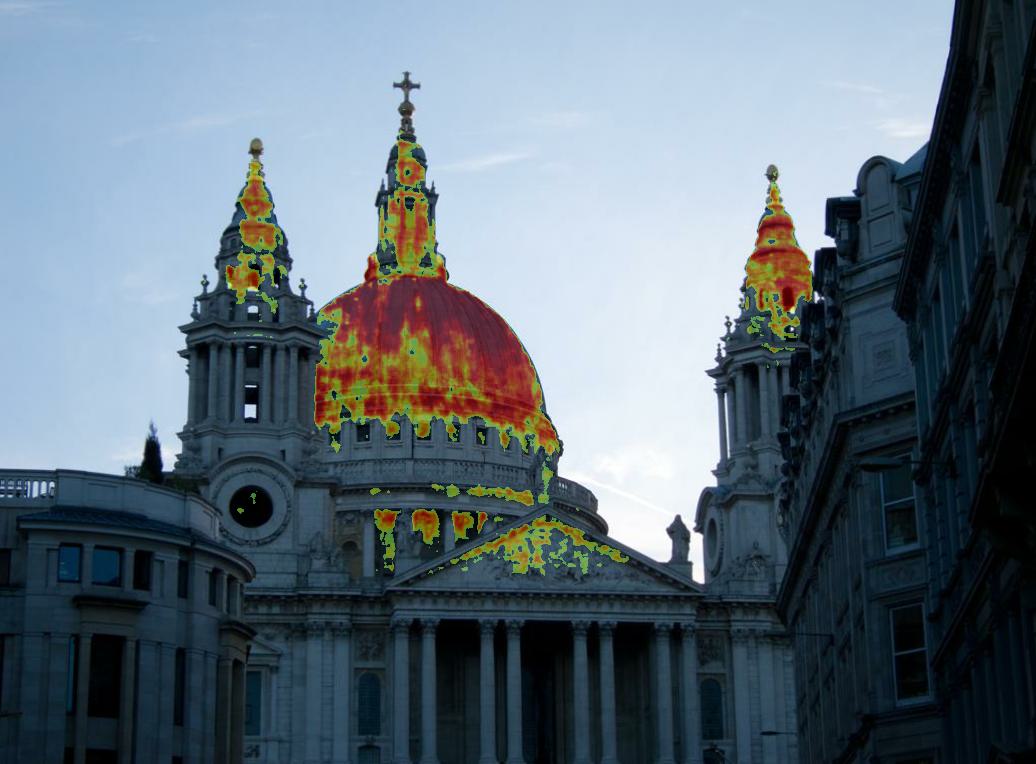}}{  }
    \hfill
    \jsubfig{\includegraphics[height=1.9cm]{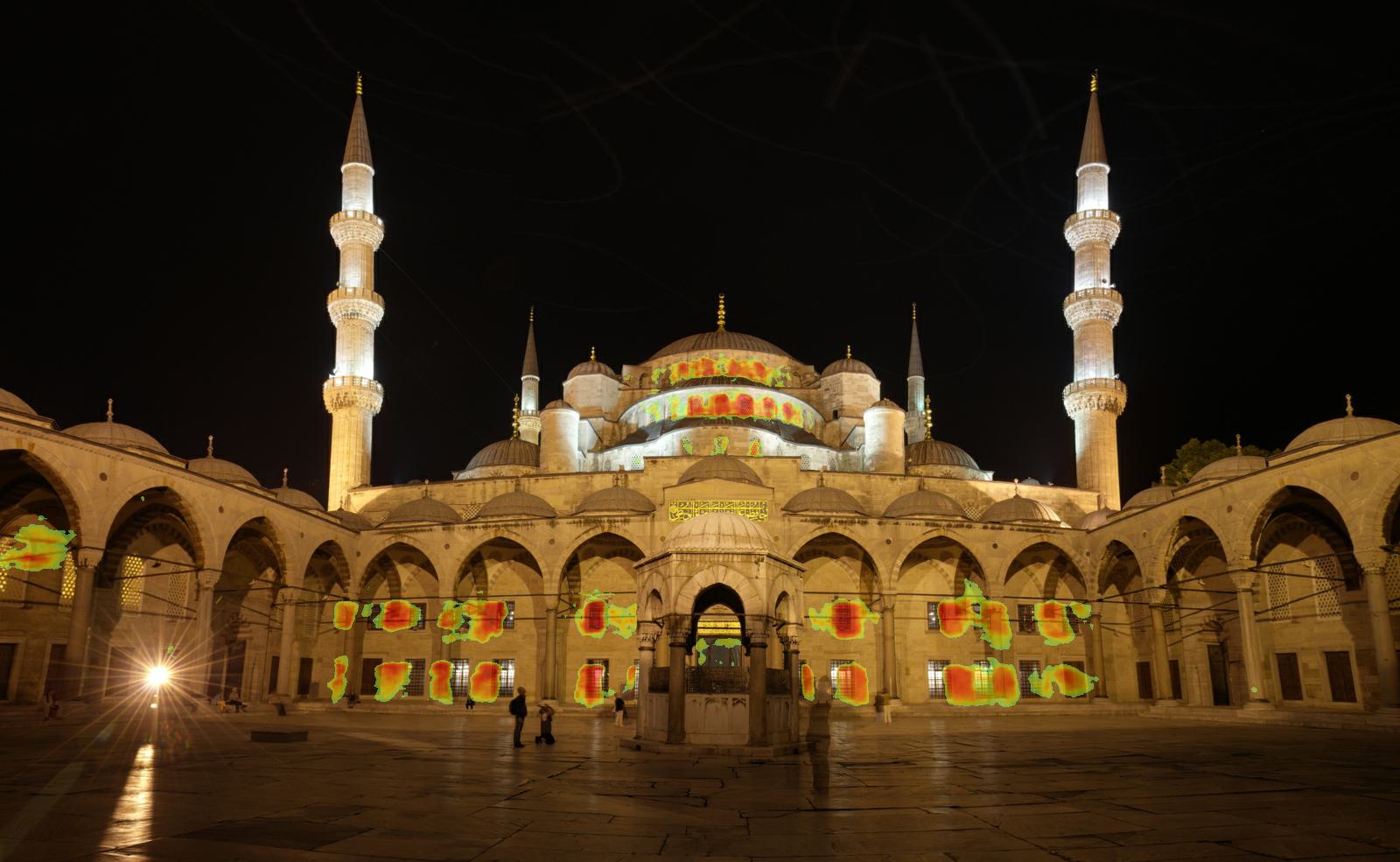}}{  }
    \hfill
    \jsubfig{\includegraphics[height=1.9cm]{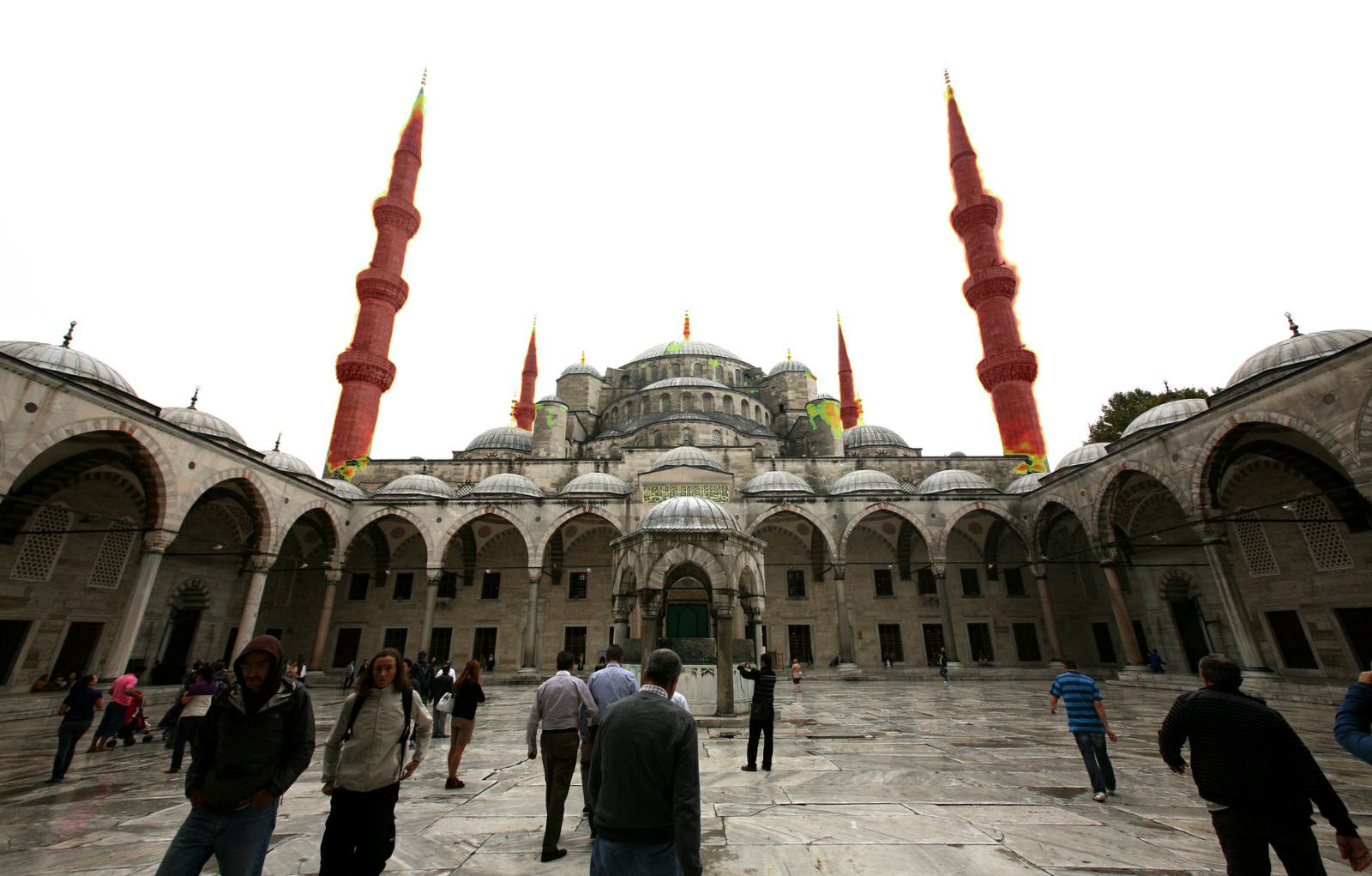}}{  }
    \hfill
    \jsubfig{\includegraphics[height=1.9cm]{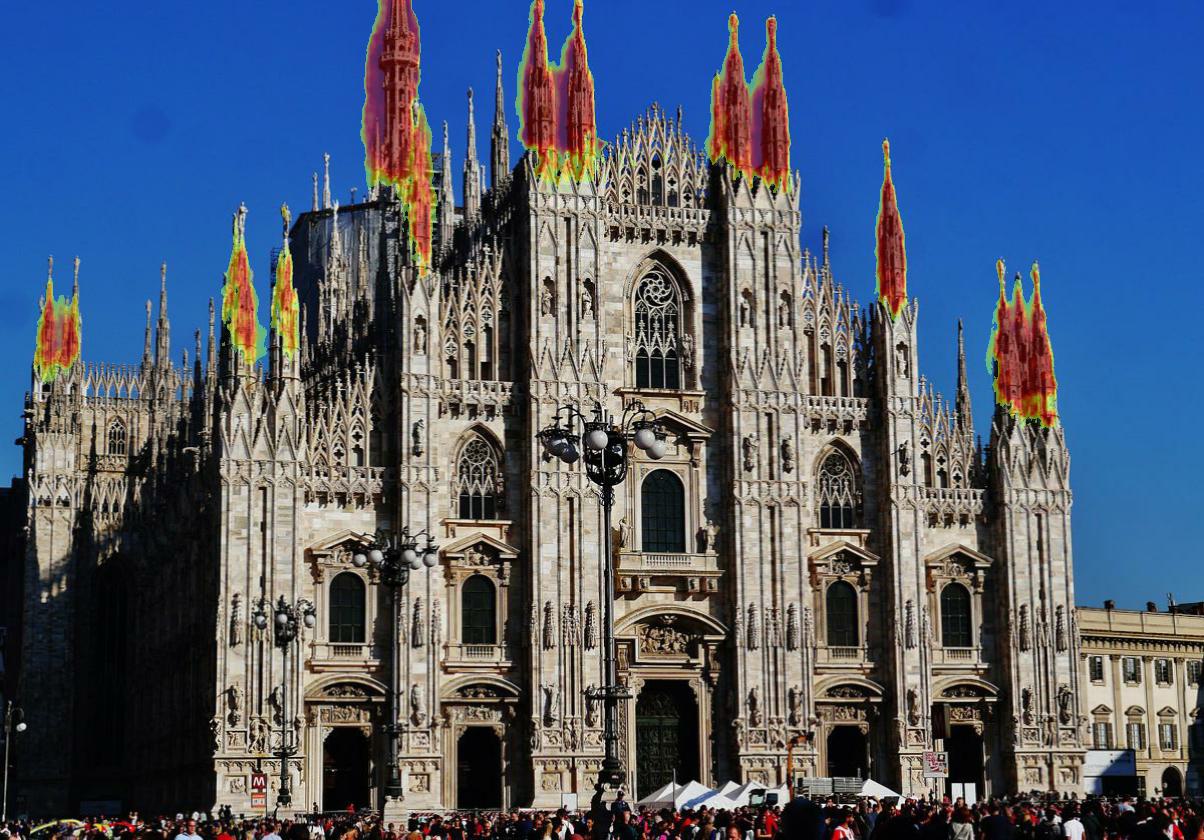}}{  }
    \hfill
    \jsubfig{\includegraphics[height=1.9cm]{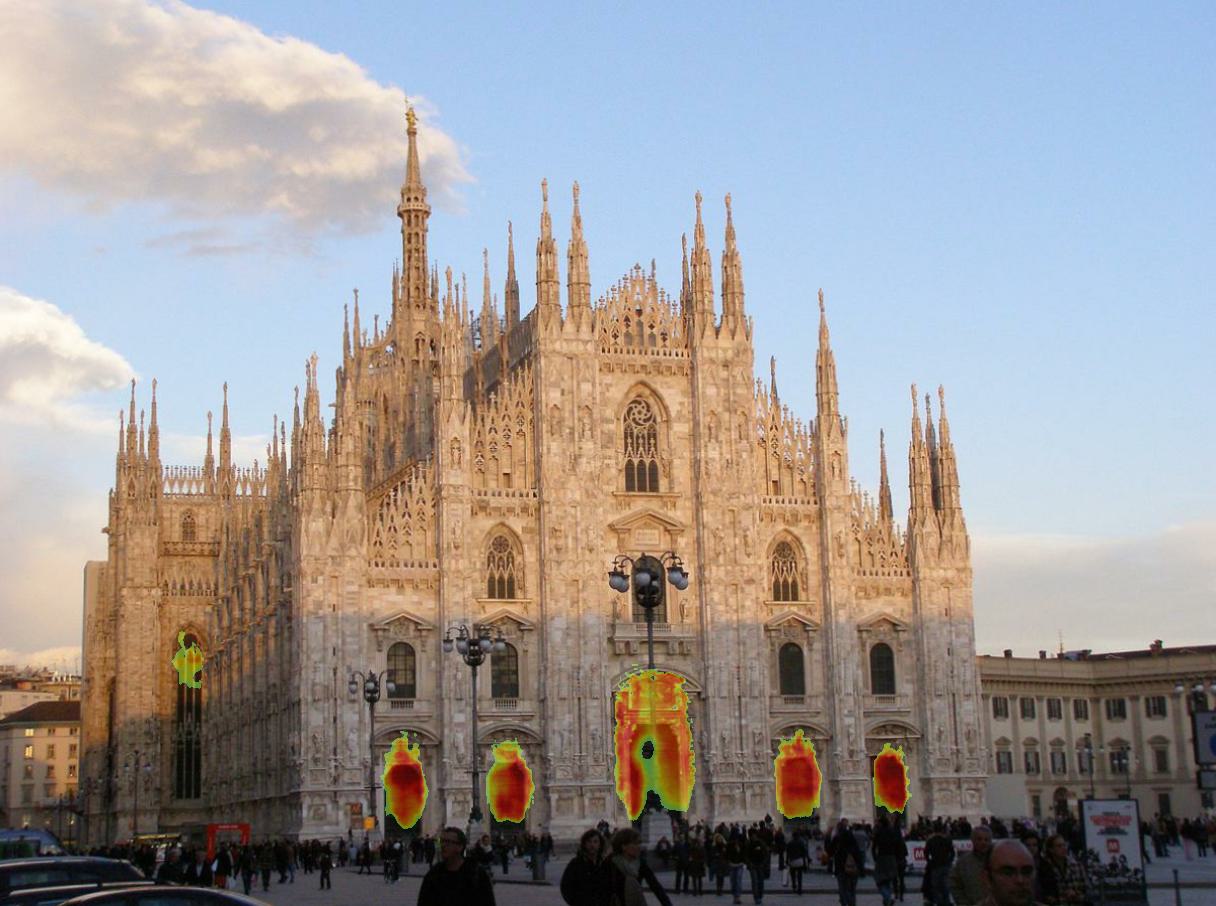}}{  }
    \\
    \rotatebox{90}{\whitetxt{x}\footnotesize{\methodname{}}}
    \jsubfig{\includegraphics[height=1.9cm]{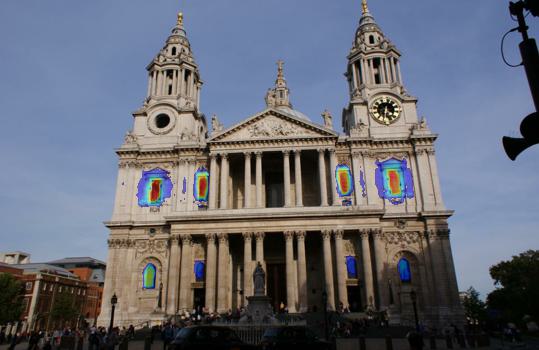}}{ \textit{Windows} }
    \hfill
    \jsubfig{\includegraphics[height=1.9cm]{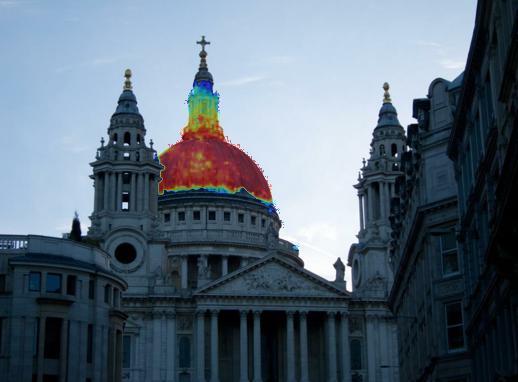}}{ \textit{Domes} }
    \hfill
    \jsubfig{\includegraphics[height=1.9cm]{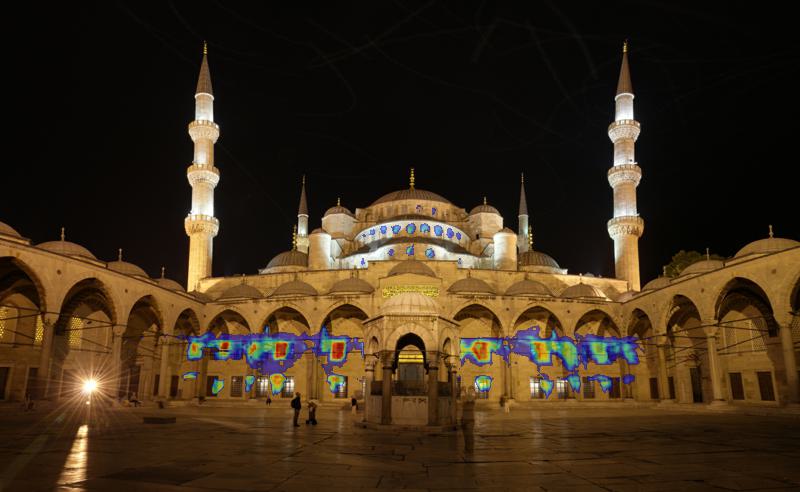}}{ \textit{Windows} }
    \hfill
    \jsubfig{\includegraphics[height=1.9cm]{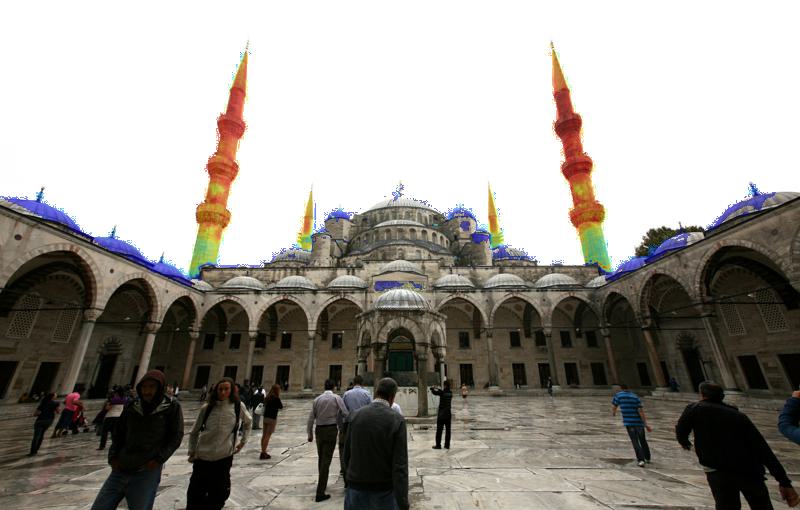}}{ \textit{Minarets} }
    \hfill
    \jsubfig{\includegraphics[height=1.9cm]{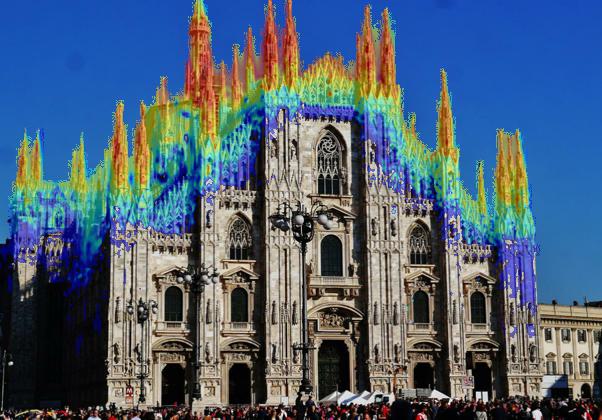}}{ \textit{Spires} }
    \hfill
    \jsubfig{\includegraphics[height=1.9cm]{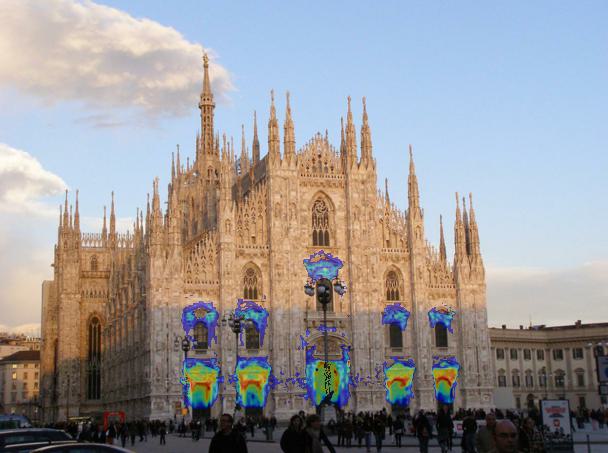}}{ \textit{Portals} }
    \caption{
      \textbf{Qualitative Comparison to HaLo-NeRF}. As illustrated above, our 3D localization is comparable with HaLo-NeRF, while being orders of magnitude faster at inference (HaLo-NeRF's runtime is roughly two hours for each queried text, vs. less than 0.1 seconds obtained with our approach).  %
      }
      \label{fig:comp_halo}

\end{figure*}

%% file: figures/final_results/final_results_2D_visualization.tex
\begin{figure*}
    \rotatebox{90}{\whitetxt{xxp}\footnotesize{LangSAM}}
    \jsubfig{\includegraphics[height=2.2cm]{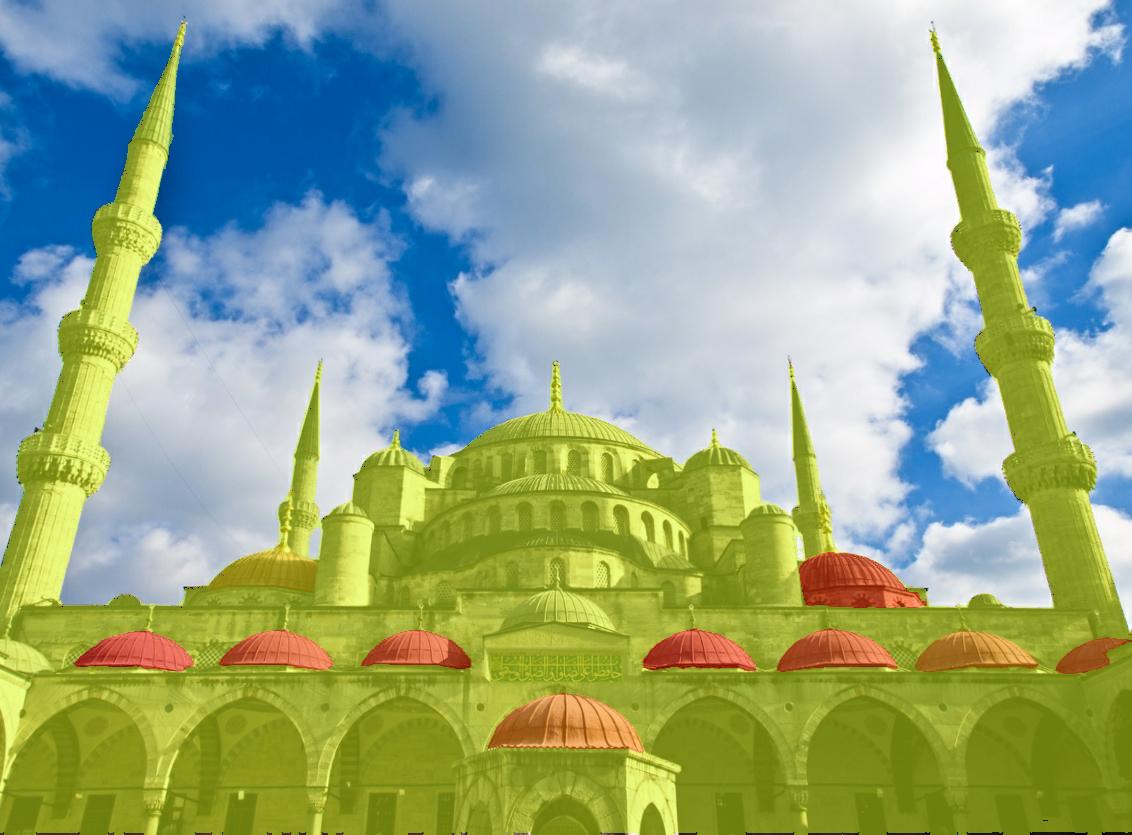}
    \includegraphics[height=2.2cm]{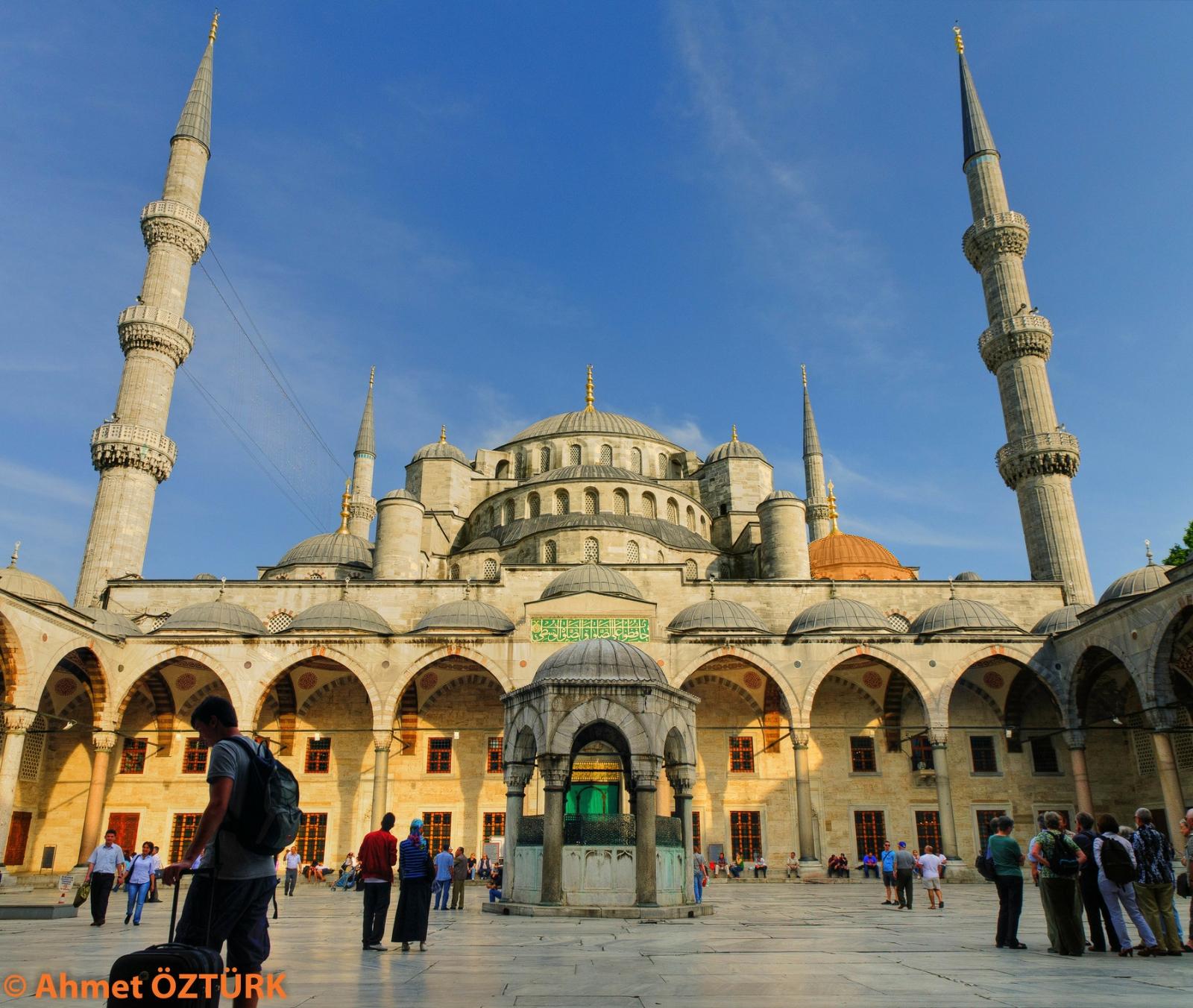}
    \includegraphics[height=2.2cm]{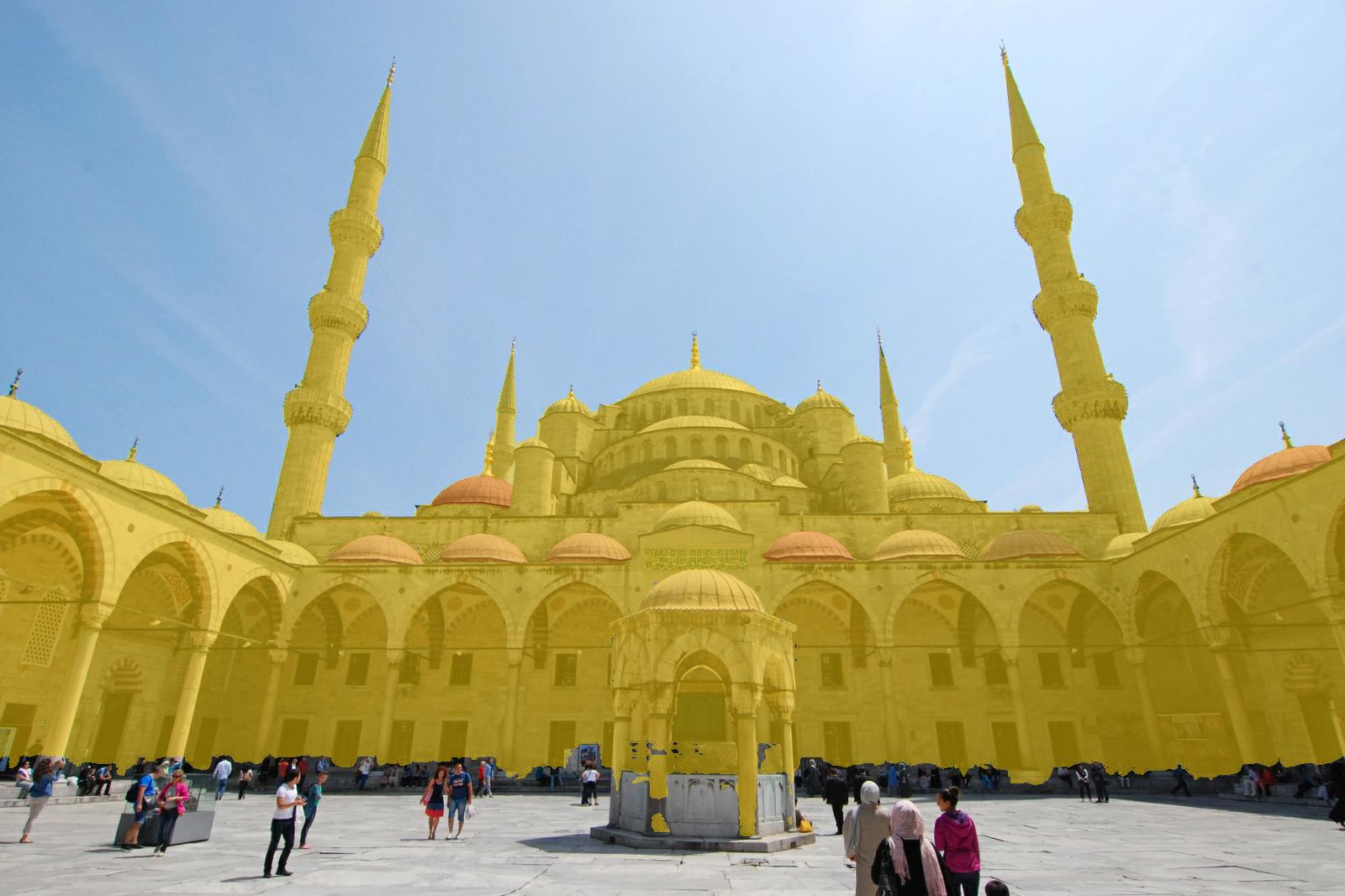}
    \includegraphics[height=2.2cm]{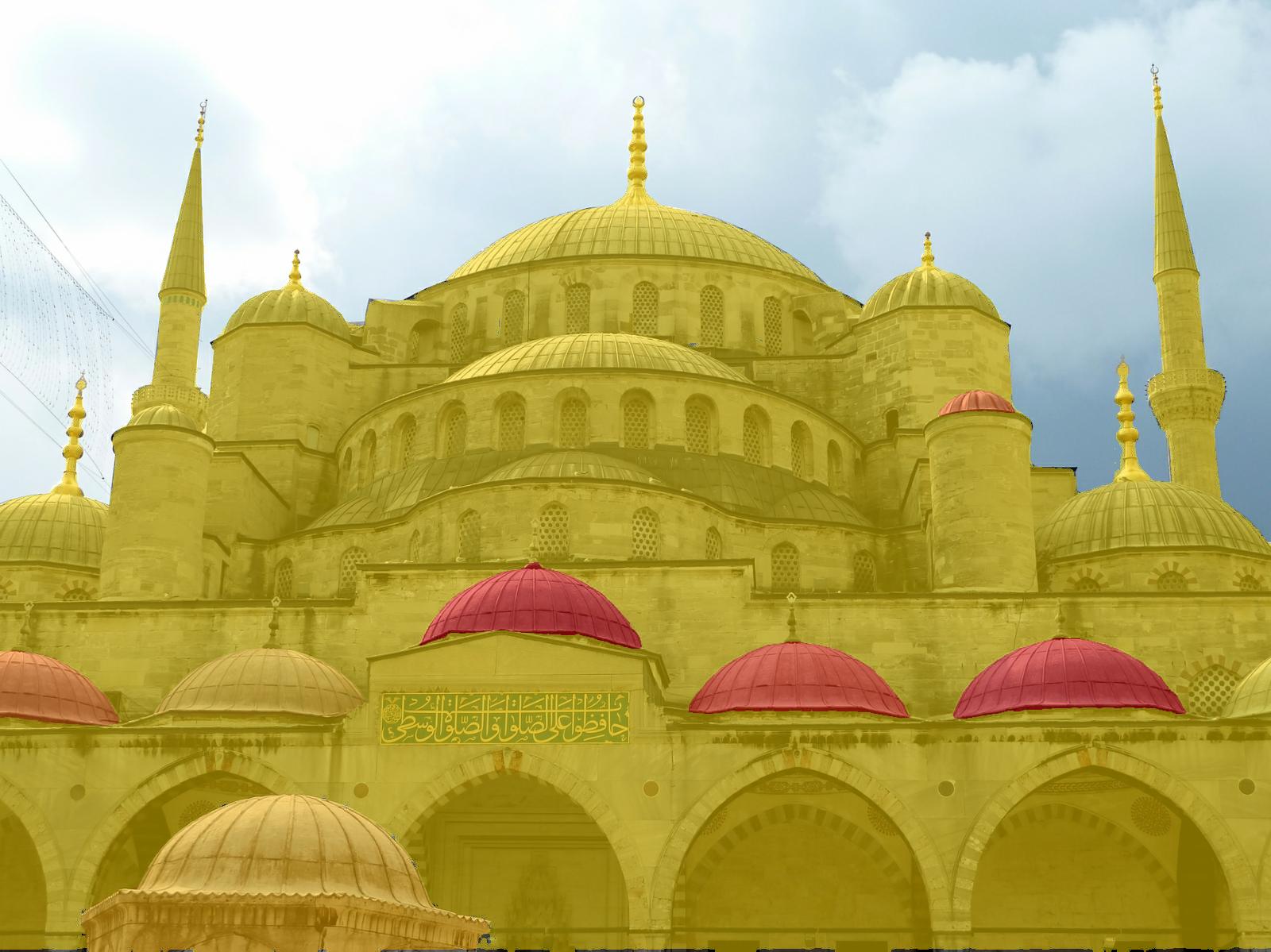}
    \includegraphics[height=2.2cm]{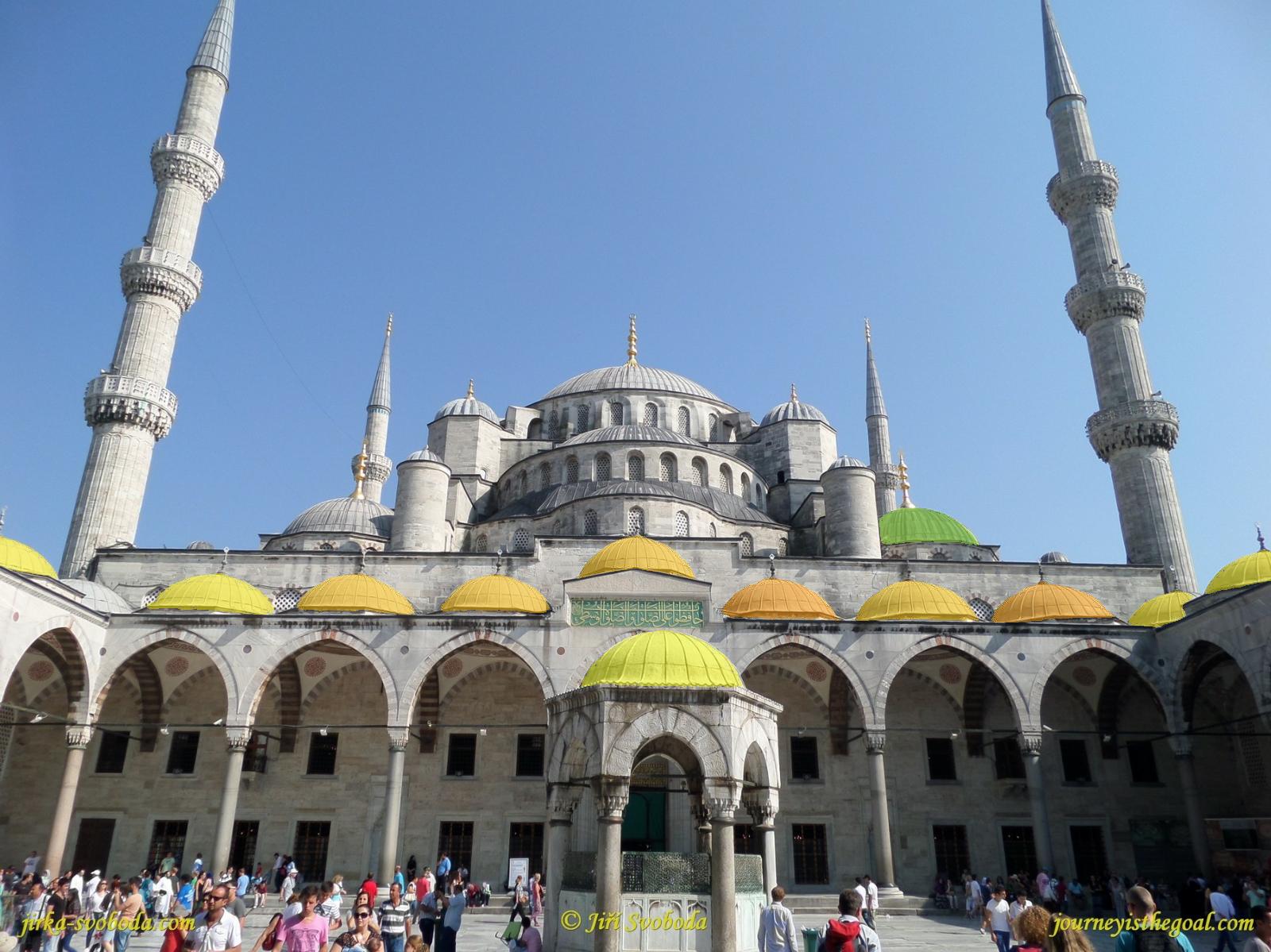}
    \includegraphics[height=2.2cm]{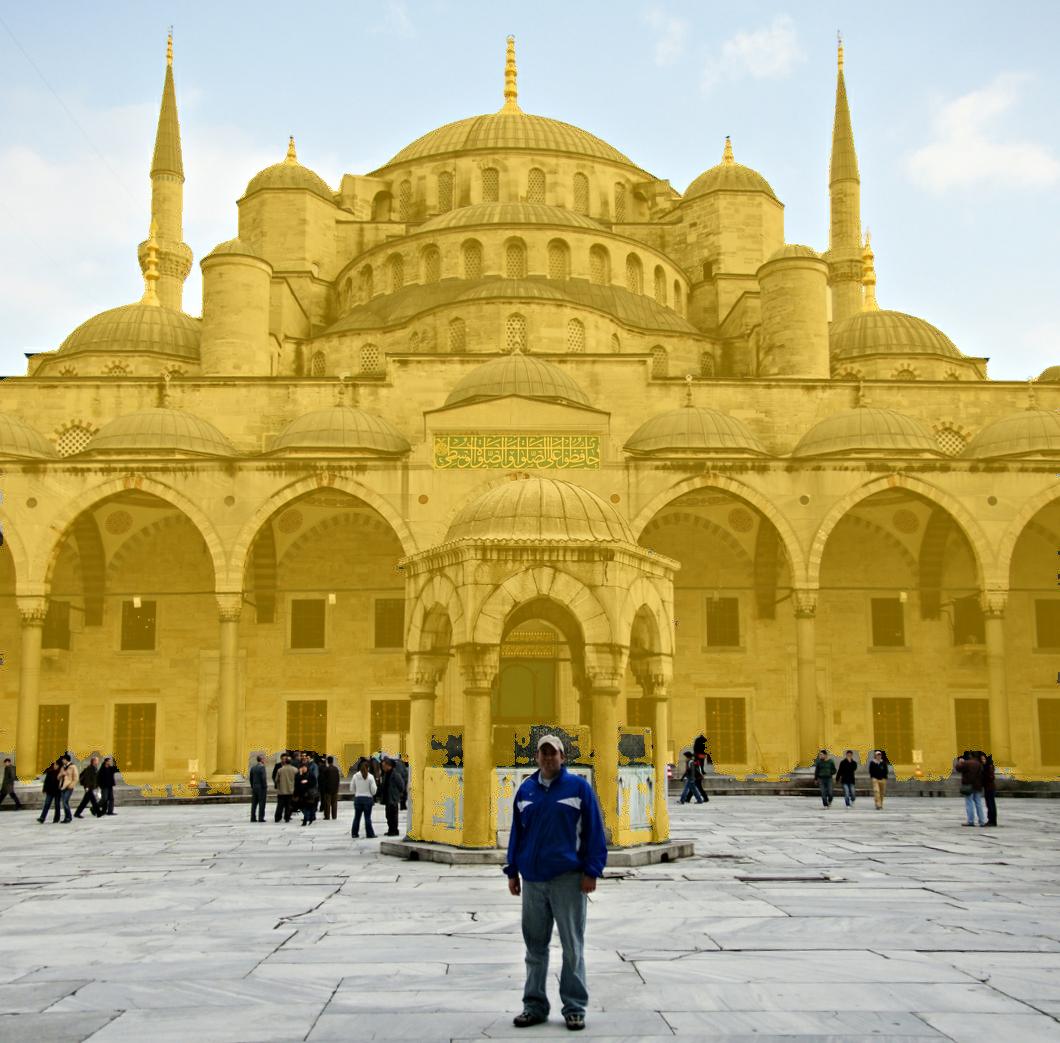}}{  }
    \\
    \rotatebox{90}{\whitetxt{x}\footnotesize{\methodname{}}}
    \jsubfig{\includegraphics[height=2.2cm]{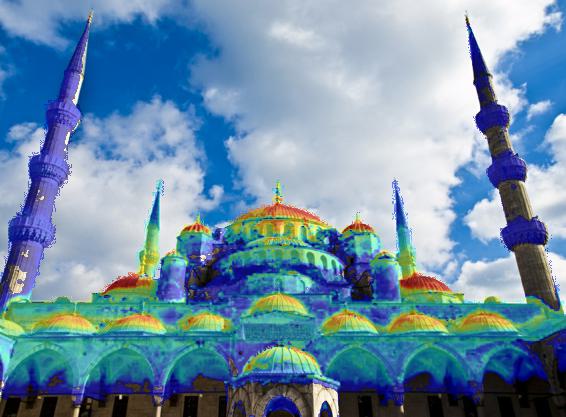}
    \includegraphics[height=2.2cm]{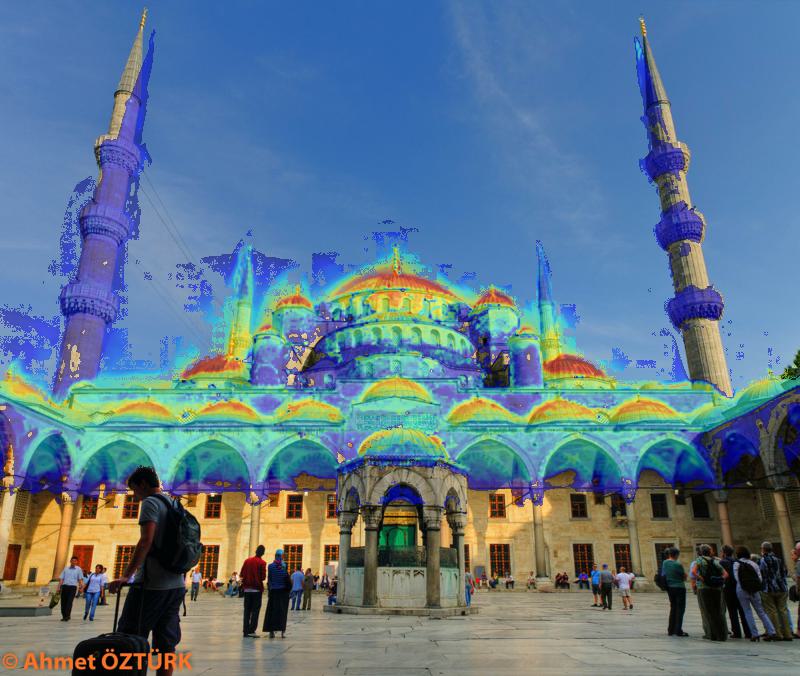}
    \includegraphics[height=2.2cm]{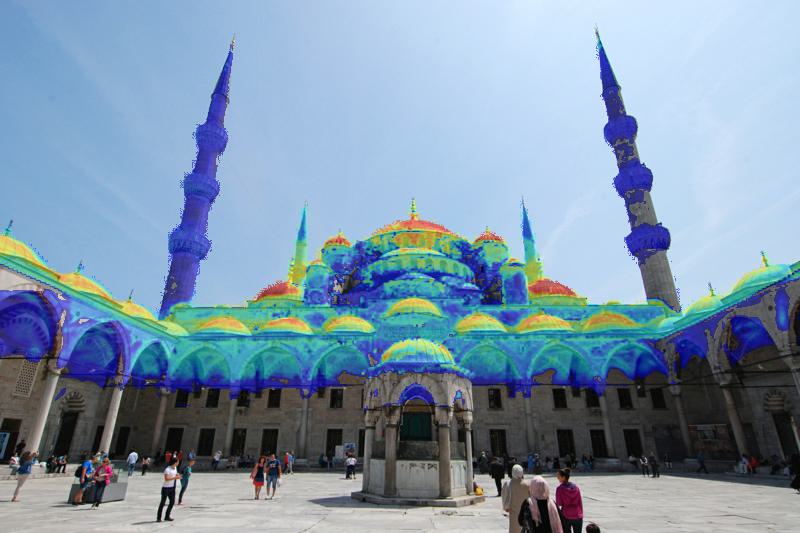}
    \includegraphics[height=2.2cm]{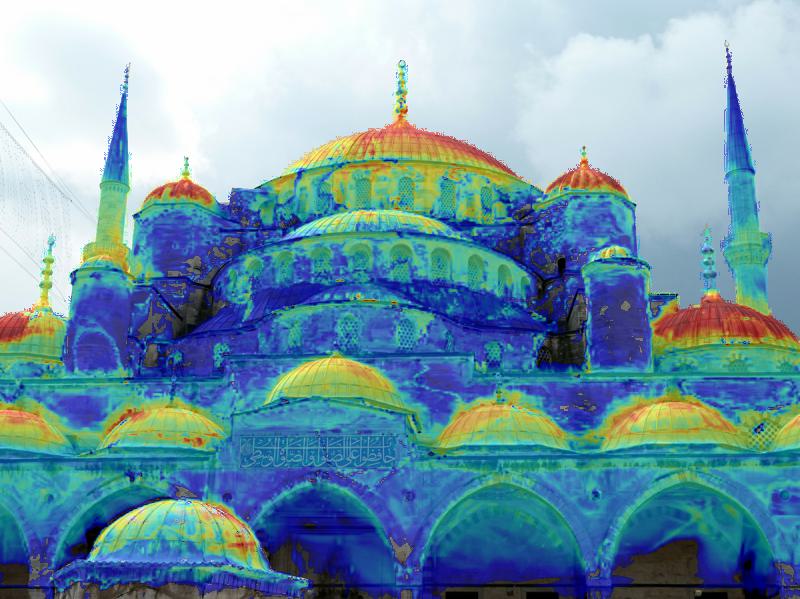}
    \includegraphics[height=2.2cm]{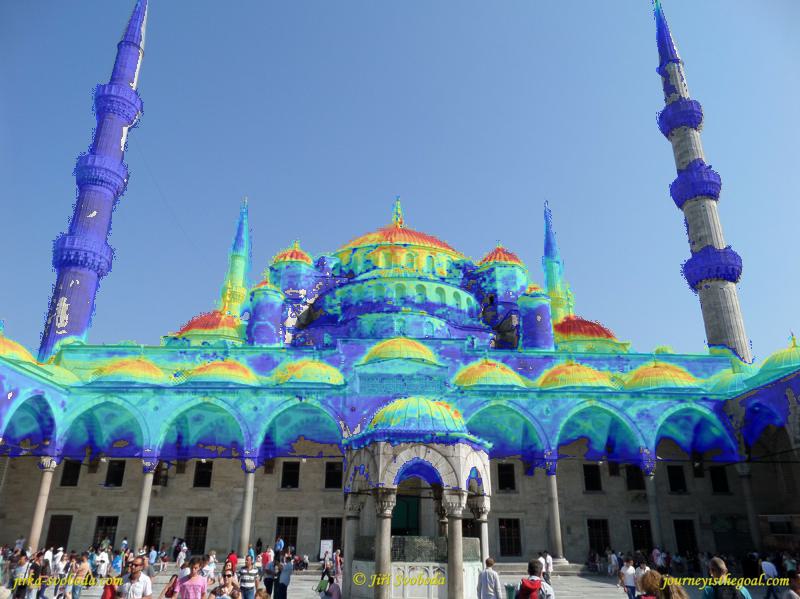}
    \includegraphics[height=2.2cm]{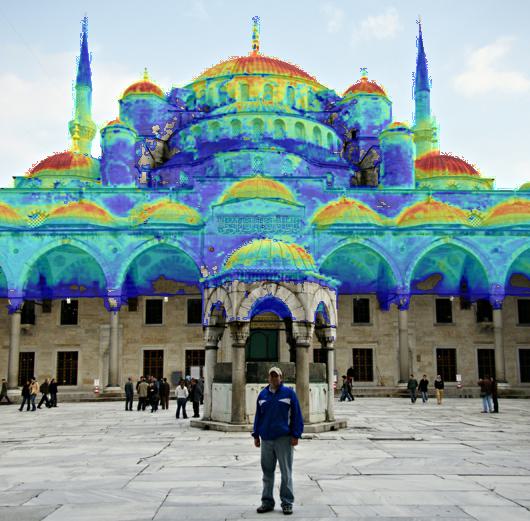}}{ \emph{Dome}  (Blue Mosque)}
    \caption{\textbf{Qualitative Comparison to LangSAM}. Above we illustrate results over multiple images belonging to the same scene. As illustrated above, while LangSAM succeeds in assigning higher probabilities to \emph{Dome} regions in some cases, its performance is not consistent across different views. Our  feature distillation approach, by contrast, yields 3D consistent results---as illustrated by the consistently-higher probabilities assigned to \emph{Dome} regions.}
    \label{fig:result_table}

\end{figure*}

%% file: figures/ablation_results/attention_seg_ablation.tex
\begin{figure*}
    \rotatebox{90}{\whitetxt{xxxq}Milano}
    \jsubfig{\includegraphics[width=2.45cm]{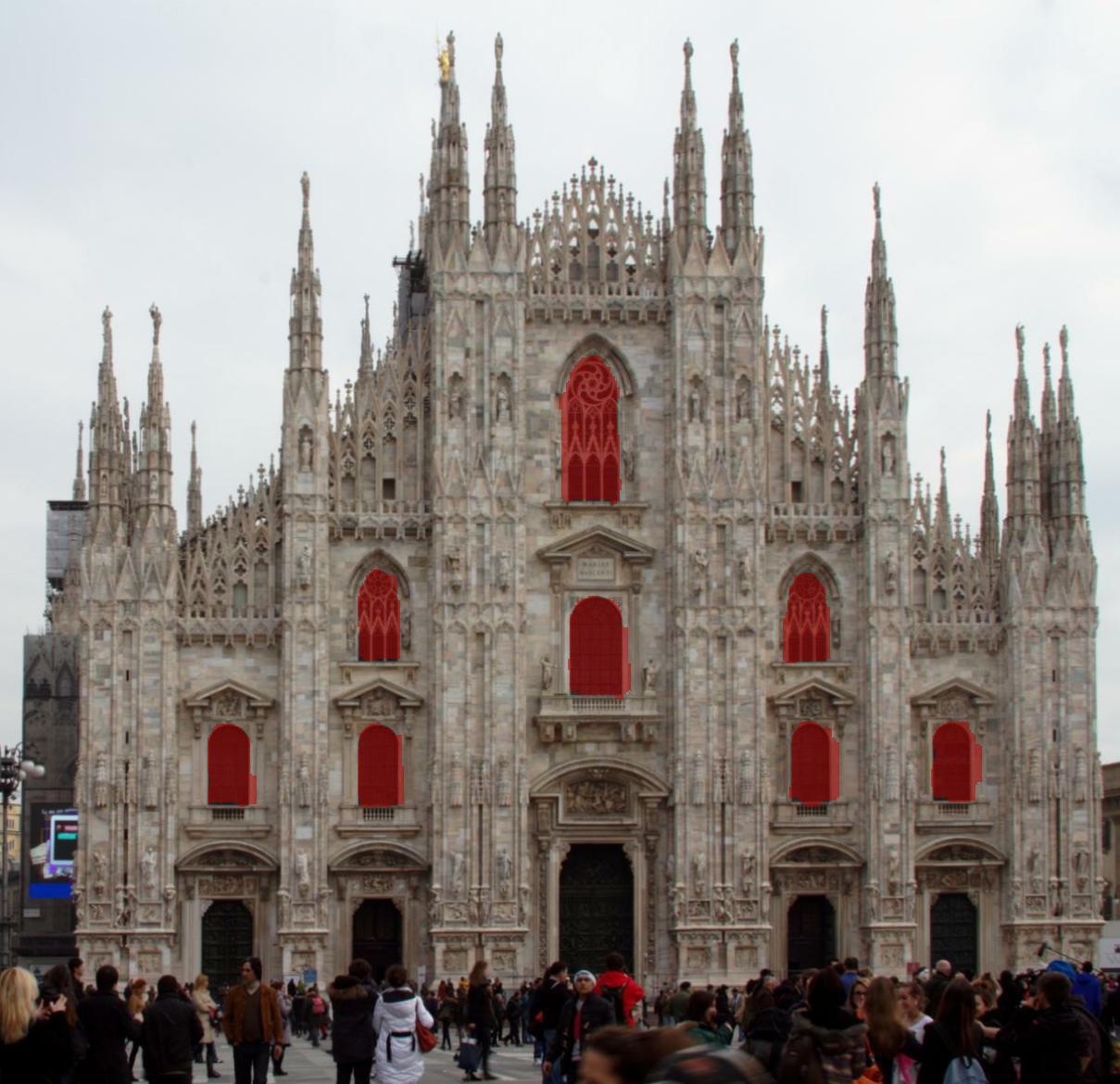}}{  }
    \hfill
    \jsubfig{\includegraphics[width=2.45cm]{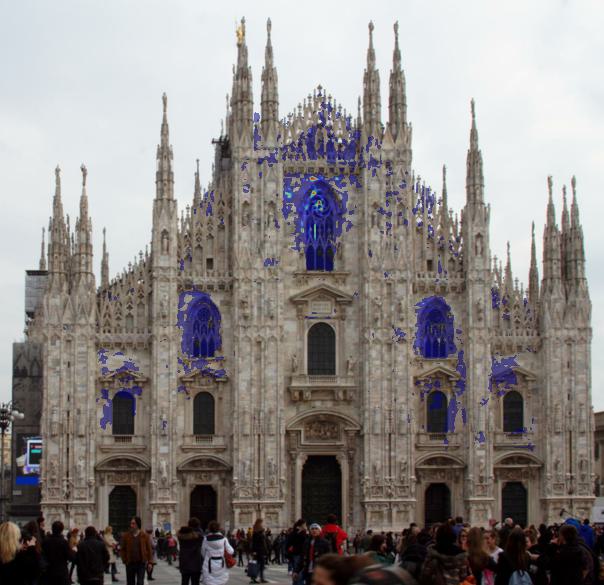}}{  }
    \hfill
    \jsubfig{\includegraphics[width=2.45cm]{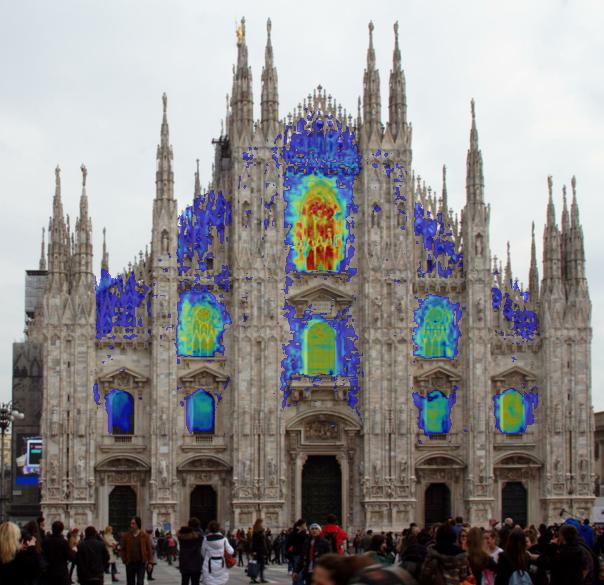}}{  }
    \hfill
    \jsubfig{\includegraphics[width=2.45cm]{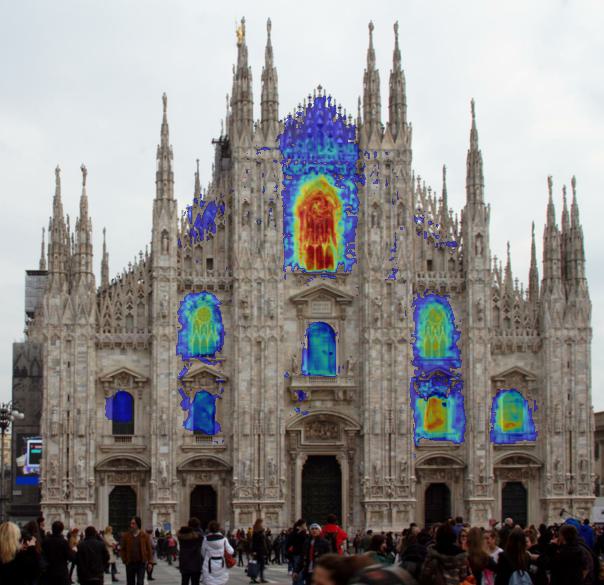}}{  }
    \hfill
    \jsubfig{\includegraphics[width=2.45cm]{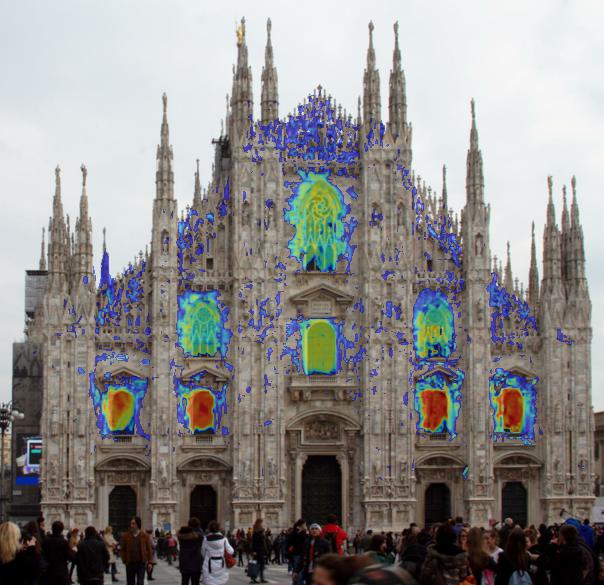}}{  }
    \hfill
    \jsubfig{\includegraphics[width=2.45cm]{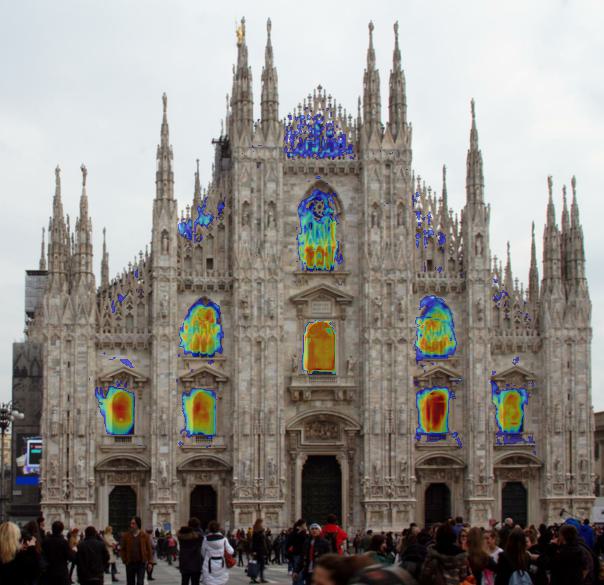}}{  }
    \hfill
    \jsubfig{\includegraphics[width=2.45cm]{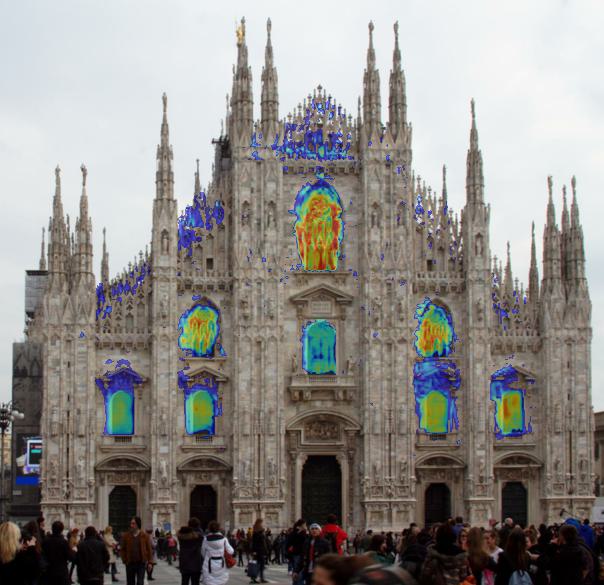}}{  }
    \\
    \rotatebox{90}{Blue Mosque}
    \jsubfig{\includegraphics[width=2.45cm]{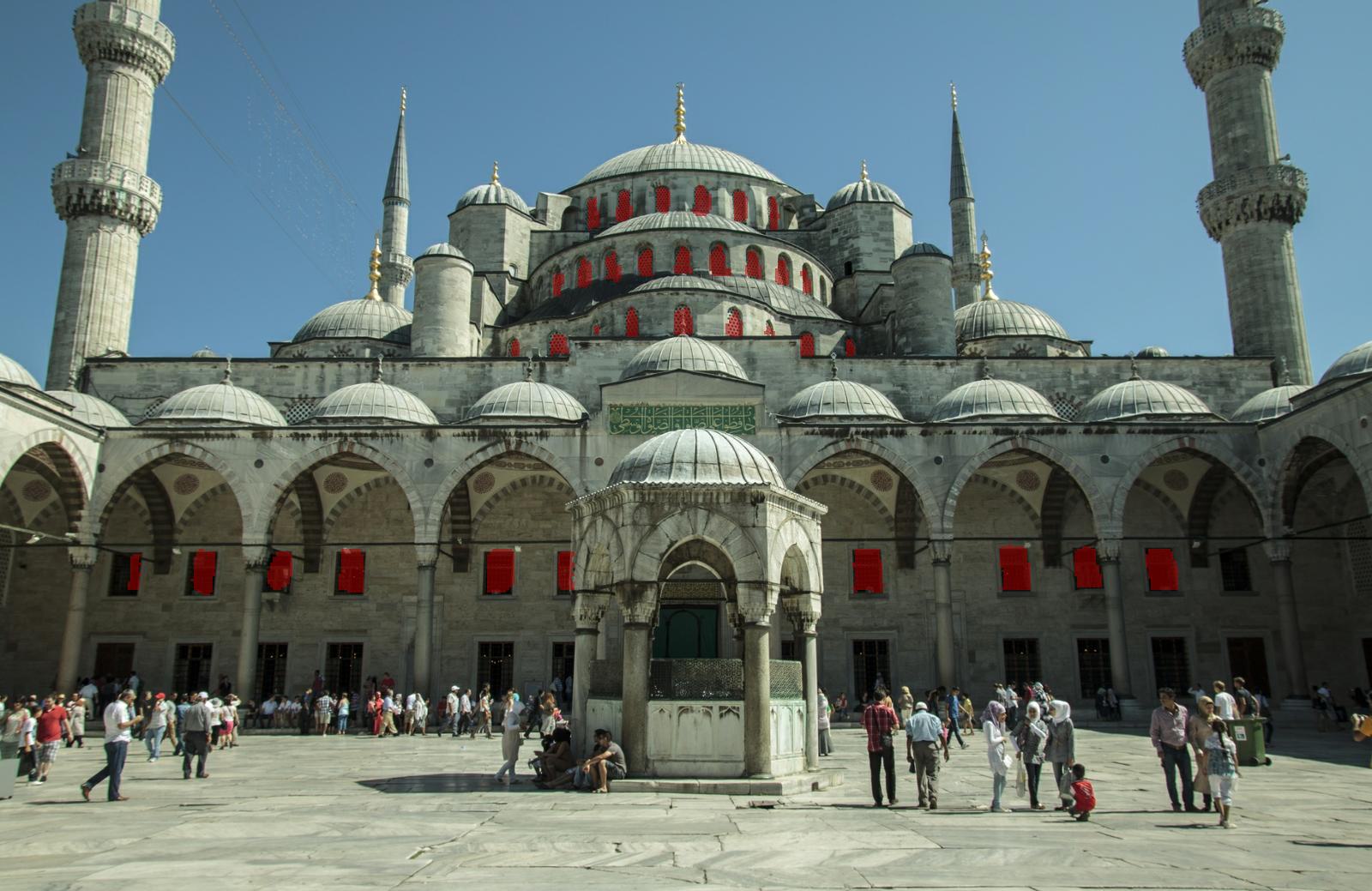}}{  }
    \hfill
    \jsubfig{\includegraphics[width=2.45cm]{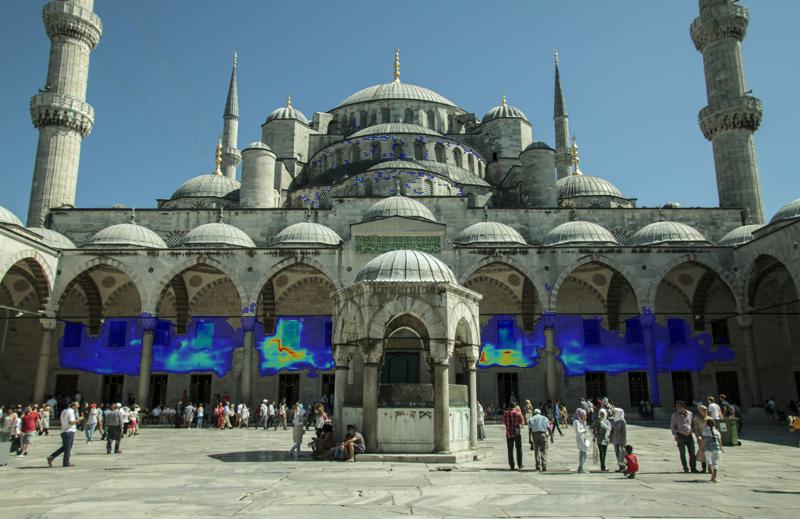}}{  }
    \hfill
    \jsubfig{\includegraphics[width=2.45cm]{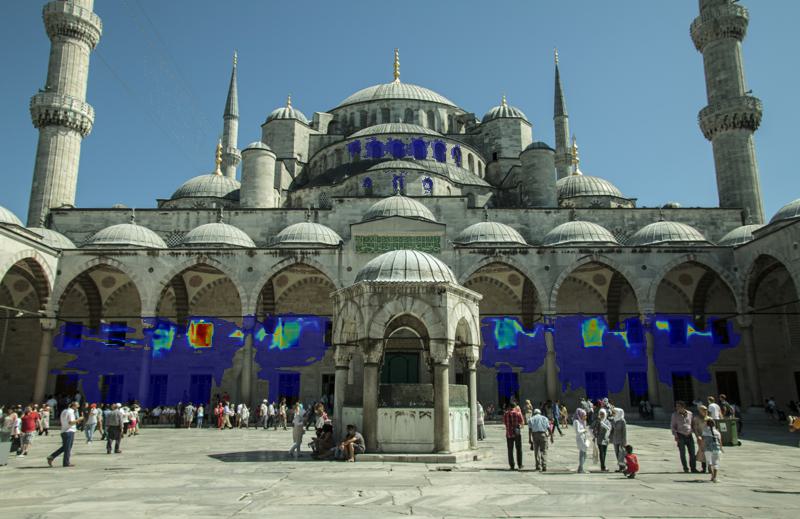}}{  }
       \hfill
    \jsubfig{\includegraphics[width=2.45cm]{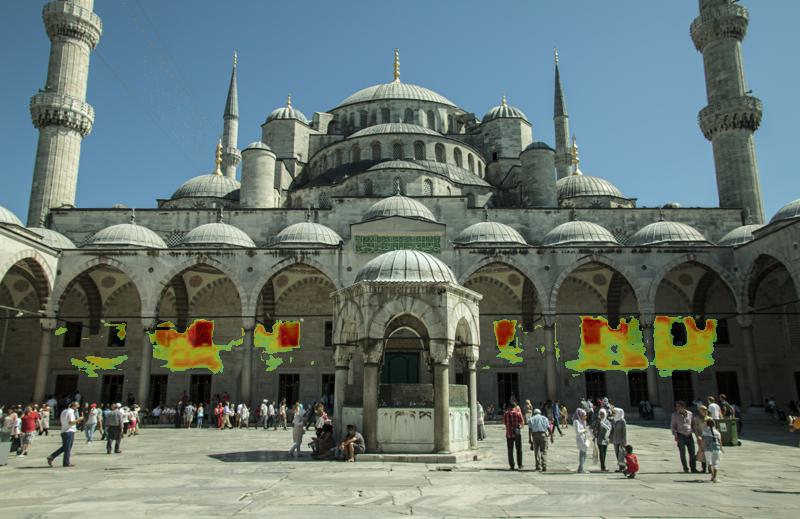}}{  }
    \hfill
    \jsubfig{\includegraphics[width=2.45cm]{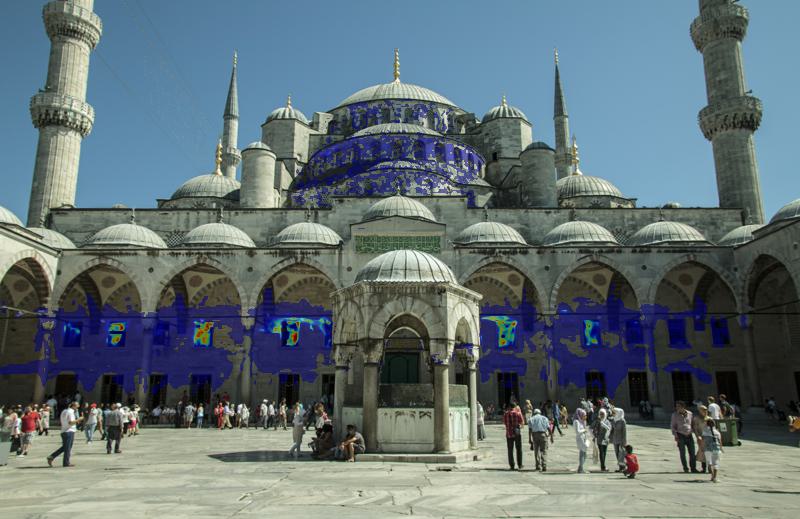}}{  }
    \hfill
    \jsubfig{\includegraphics[width=2.45cm]{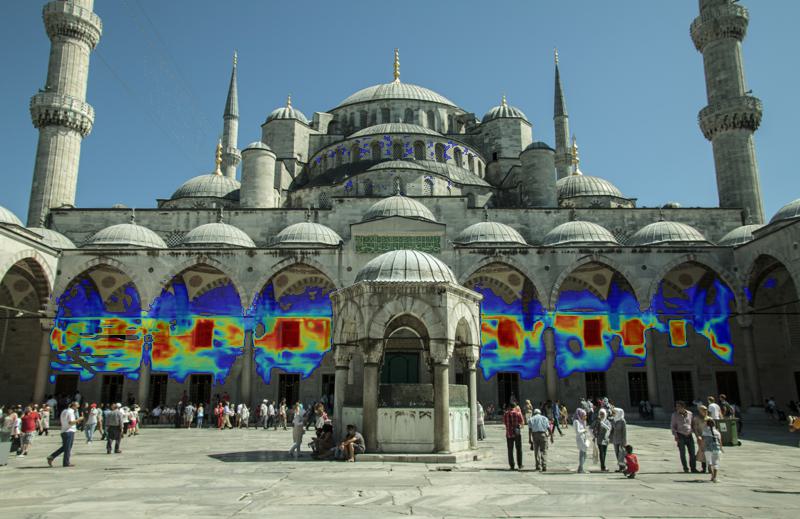}}{  }
    \hfill
    \jsubfig{\includegraphics[width=2.45cm]{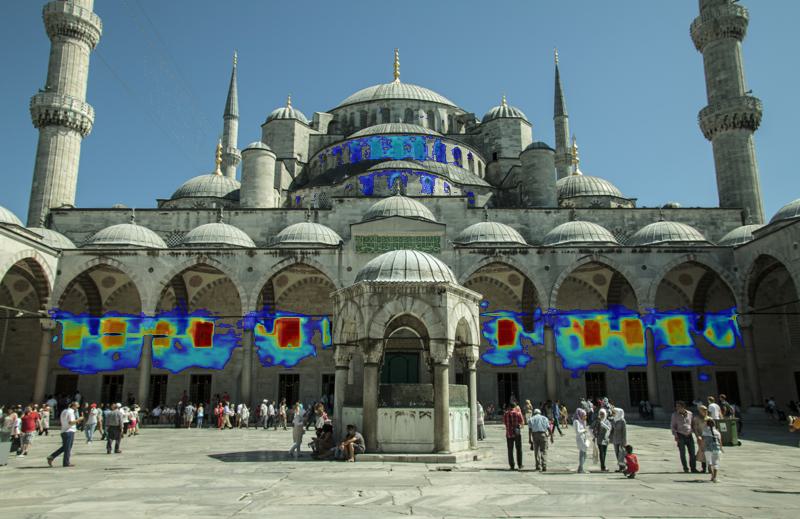}}{  }
    \\ %
    \rotatebox{90}{\whitetxt{xxp}St.Paul}
    \jsubfig{\includegraphics[width=2.45cm]{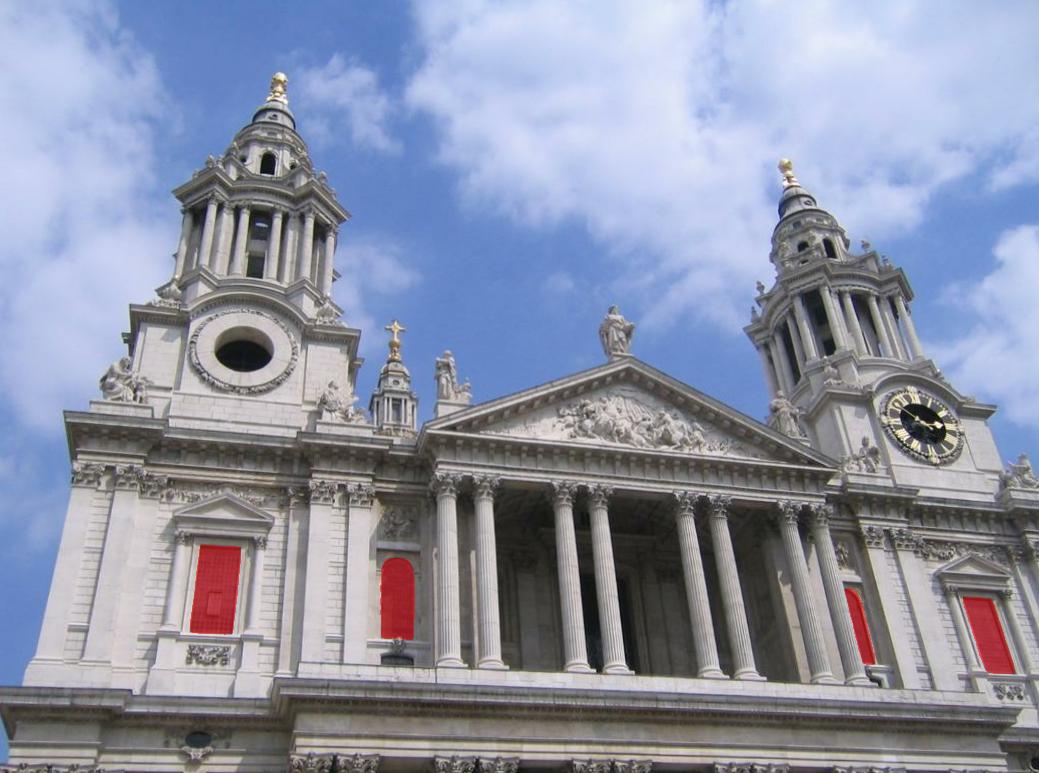}}{ \textit{GT} }
    \hfill
    \jsubfig{\includegraphics[width=2.45cm]{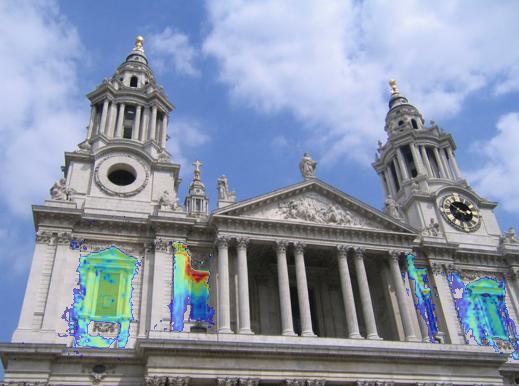}}{ \textit{w/o DINO} }
    \hfill
    \jsubfig{\includegraphics[width=2.45cm]{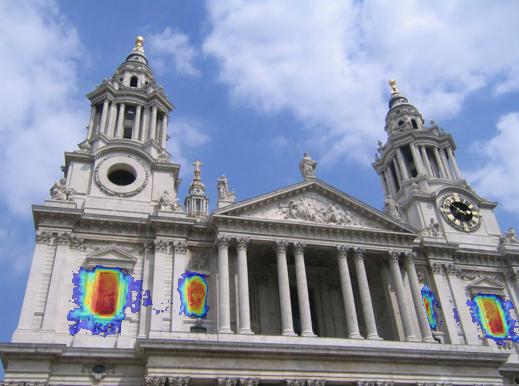}}{ \textit{w/o ADM (LR)} }
       \hfill
    \jsubfig{\includegraphics[width=2.45cm]{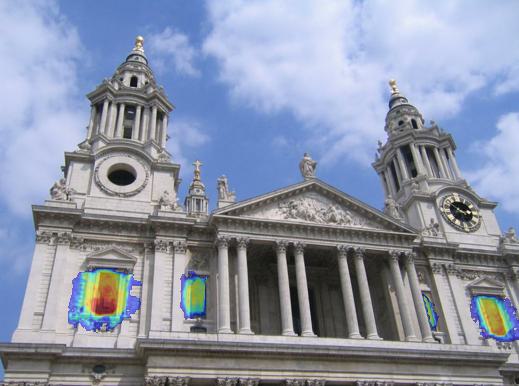}}{ \textit{w/o ADM (HR)} }
    \hfill
    \jsubfig{\includegraphics[width=2.45cm]{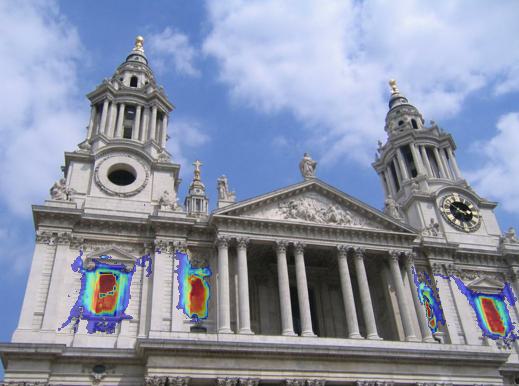}}{ \textit{w/o SAM reg.} }
    \hfill
    \jsubfig{\includegraphics[width=2.45cm]{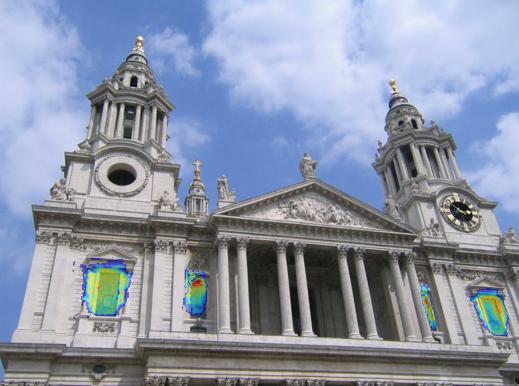}}{ \textit{w/o Phys. Scale} }
    \hfill
    \jsubfig{\includegraphics[width=2.45cm]{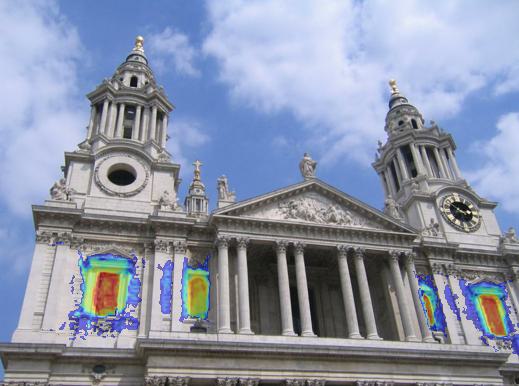}}{ \textit{Full Version}
    } \\
    \caption{\textbf{Semantic Distillation Ablations.} Above, we provide a qualitative ablation study, ablating our proposed architectural modifications and regularizations (as further detailed in Sec.~\ref{sec:ablations}). As illustrated above, the DINO regularization is essential for precise localization (\emph{e.g.}, see "w/o DINO", middle). The importance of the Attenuated Downsampler Module is noticeable in both ablated variants (\emph{e.g.}, the models struggle to detect the small windows in the Blue Mosque). Similarly, removing the SAM regularization and the physical scale selection mechanism negatively affects performance. 
      }
      \label{fig:ablations_attention_seg}
    
\end{figure*}

%% file: supp/supp_details.tex
\section{Additional Details}
\label{sec:additional_details}

In this section, we provide comprehensive implementation details for our method, including architectural specifications, training procedures, and dataset-specific adaptations.

\subsection{Gaussian Splatting In-The-Wild Adaptation}
\label{subsec:swag_details}

As mentioned in the main text, we integrate the SWAG method~\cite{dahmani2025swag} into our pipeline to handle the challenging in-the-wild nature of our scenes. Their code is not publicly available so we implement a version ourselves. SWAG enhances 3D Gaussian Splatting by modeling per-image appearance variations through learnable embedding vectors. Specifically, for each 3D Gaussian $g_j$ and input image $I_k$, a learnable per-image embedding vector $\mathbf{e}_k \in \mathbb{R}^{d_e}$ is processed by a neural network $f_{\text{app}}$ to predict appearance adaptations:
\begin{equation}
(\Delta\mathbf{c}_j, \Delta\alpha_j) = f_{\text{app}}(\mathbf{c}_j, \text{PE}(\boldsymbol{\mu}_j), \mathbf{e}_k)
\end{equation}
where $\Delta\mathbf{c}_j$ and $\Delta\alpha_j$ represent RGB color shifts and opacity shifts, respectively. These shifts dynamically adapt each Gaussian's base color $\mathbf{c}_j$ and opacity $\alpha_j$ for individual images, effectively mitigating artifacts from transient objects and varying illumination conditions. The positional encoding $\text{PE}(\boldsymbol{\mu}_j)$ provides spatial context for the appearance network.

\subsection{Query Prompt Enhancement}
\label{subsec:prompt_enhancement}

To improve robustness across architectural vocabularies, we employ LLM-based query prompt enhancement. For specialized terminology, we generate more common synonymous alternatives by prompting GPT‑4.5 with the following:

\textit{''Find synonyms that are more common or equally common synonyms for [Query Prompt]. Arrange the synonyms from the most common to the least common. Give your answer separated by commas. If there are no synonms that are more or equally common, return nothing.''}

For example, \textit{portal} becomes \textit{''doorways, gates, entrances, portals''}, providing multiple semantic anchors for CLIP~\cite{radford2021learning}.

\subsection{Language-SAM Building Mask}
\label{subsec:langsam_mask}

We employ LangSAM~\cite{ravi2024sam2segmentimages} to generate building masks for input images. These masks serve dual purposes: they are incorporated into training losses to focus learning on building regions while ignoring irrelevant background areas, and they are applied at inference time to constrain predictions exclusively to architectural structures (similar to the evaluation protocol conducted in HaLo-NeRF).

\subsection{Real-World Pixel Size Computation}
\label{subsec:pixel_size_computation}

Our physical scale pyramid approach requires accurate estimation of real-world pixel dimensions from COLMAP reconstruction data. We compute the physical width of a pixel using the following procedure:

\smallskip
\noindent \textbf{Step 1: Scene Distance Estimation.} We calculate the Euclidean distance from the camera center to the mean of visible 3D points:
\begin{equation}
d = \left\| \mathbf{C}_{\text{cam}} - \frac{1}{|\mathcal{P}|}\sum_{\mathbf{p} \in \mathcal{P}} \mathbf{p} \right\|_2
\end{equation}
where $\mathbf{C}_{\text{cam}}$ is the camera center and $\mathcal{P}$ represents the set of 3D points visible in the image. To reduce outlier influence, we exclude 3D points that are close to the image edges.

\smallskip
\noindent \textbf{Step 2: Angular Resolution Calculation.} We compute the instantaneous field of view (IFOV) per pixel:
\begin{equation}
\theta_{\text{per-pixel}} = \frac{\arctan\left(\frac{I_{\text{width}}}{f_x}\right)}{I_{\text{width}}}
\end{equation}
where $I_{\text{width}}$ is the image width in pixels and $f_x$ is the horizontal focal length.

\smallskip
\noindent \textbf{Step 3: Physical Size Conversion.} The real-world pixel width is computed by converting angular size to physical dimensions at distance $d$:
\begin{equation}
P_{\text{width}} = 2d \cdot \tan(\theta_{\text{per-pixel}})
\end{equation}

We compute the height similarly and use the average of width and height for determining CLIP pyramid crop window sizes. CLIP pyramids over several examples are illustrated in Fig.~\ref{fig:clip_pyramid_scales}.

\input{figures/ablation_results/clip_pyramids_scales/clip_pyramid_scales}

\subsection{Network Architecture Details}
\label{subsec:architecture_details}

\subsubsection{Multi-Resolution Hash Encoding}
Our semantic bottleneck employs a multi-level hash encoder inspired by Instant-NGP~\cite{M_ller_2022}. The hash grid is configured with an input dimension of 4, utilizing 16 levels with a logarithmic hashmap size of 8. The minimum and maximum resolutions are set to 16 and 2048, respectively. 

The input to the hash grid consists of our low-dimensional encoded features (dimension size of 3) concatenated with an index value. We iterate through the hash grid forwarding process, incrementing the index in each iteration until the complete feature vector with the required dimensionality is constructed.

\subsubsection{MLP Architecture}
Following the multi-layer hash encoding, features are processed through a compact MLP consisting of five layers. The network architecture is as follows:
\begin{enumerate}
    \item \texttt{Linear(clip\_dim, clip\_dim)}
    \item \texttt{Linear(clip\_dim, clip\_dim)}
    \item \texttt{Linear(clip\_dim, clip\_dim)} 
    \item \texttt{ReLU()} 
    \item \texttt{Linear(clip\_dim, clip\_dim + dino\_dim)} 
\end{enumerate}
We add weight decay of 1e-5 to this module.
The output features are with dimensionality corresponding to the concatenated CLIP and DINOv2~\cite{oquab2023dinov2} features, where \texttt{clip\_dim} is 512 and \texttt{dino\_dim} is 384.

\subsection{Training Details}
\label{subsec:training_dets}

We optimize RGB and semantic features jointly in the training loop. For our comparison results, we train all scenes for 70,000 iterations, with the exception of Badshahi Mosque, which requires 200,000 iterations due to its exceptional scale and complexity. To improve convergence in challenging scenes, we adjust the D-SSIM to L1 RGB loss ratio within the reconstruction objective $L_{\text{rec}}$. The loss function coefficients are set as follows: $\lambda_{\text{CLIP}} = 1.0$, \rev{$\lambda_{\text{DINO}} = 0.5$ and $\lambda_{\text{SAM}} = 0.005$}.

\subsubsection{CLIP Pyramid Scales}
Physical scales for CLIP pyramids are manually adjusted per scene based on the COLMAP-derived dimensions, which vary significantly across different landmarks. \rev{This process can be automated, for example, by using off-the-shelf metric depth estimators and the COLMAP scene dimensions.}

\subsubsection{Badshahi Mosque Special Parameters}
Given the extended training duration for the Badshahi Mosque scene, we modify several parameters to accommodate the large iteration count. The densify\_from\_iter is set to 200000, densification\_interval to 10000, and densify\_until\_iter to 5000. Position learning rates are reduced to 1.6e-6 initially and 1.6e-9 finally, with scaling learning rate set to 0.0005.

\subsubsection{Learning Rates and Hyper-parameters}
Our training employs a resolution factor of 2 with carefully tuned learning rates. The positional learning rate is initialized at 0.000016 and decays to 0.00000016. The feature learning rate for RGB is set to 0.00025, while the opacity learning rate is \rev{0.05}. Our semantic features utilize a learning rate of 0.001. \rev{DSSIM loss weight is set to 0.5} 
The percent dense parameter is set to 0.2 with densification occurring every 2000 iterations. Opacity reset intervals are extended to 100000 iterations 
, and densification continues until iteration 15000. For architectural components, both the features hash and MLP, as well as the Attenuated Downsampler, employ learning rates of 0.0005. The densification gradient threshold is set to 0.0002. The SWAG hash and MLP components also use 0.0005, while SWAG latent vectors are optimized with 0.001. Additionally, we incorporate MLP feature decay at 0.00001 and blur probabilities of 2.5.

During training, we resize rendered low-dimensional feature maps to random sizes between 80 and 120 in width and height before decoding through the Multi-Layer Hash Encoding. This approach allows different Gaussians to be more dominant at various scales while accommodating GPU memory constraints that prevent processing full-size maps during training. At inference, we process the full-resolution feature maps.

\rev{Overall, training a single scene is approximately 2 hours on a single NVIDIA RTX A5000 GPU and the number of
Gaussians in each scene ranges from 30K to 360K.}

\subsection{Comparisons Details}
\label{sec:comp_details}

Below we provide all the details needed to reproduce the comparisons shown in the paper.

\subsubsection{LangSAM}
We use the code provided by authors in \url{https://github.com/luca-medeiros/lang-segment-anything} with the default parameters.

\subsubsection{Feature3DGS}
We use the code provided by authors in \url{https://github.com/ShijieZhou-UCLA/feature-3dgs}. In order to run the method on large-scale in-the-wild scenes we added the SWAG model as the backbone and used our semantic bottleneck implementation for feature encoding.

\subsubsection{FMGS}
We were unable to run the FMGS code
due to excessive GPU memory requirements on large-scale scenes. Therefore, we adapted their code into our framework by incorporating their core algorithmic principles, specifically relocating the hash encoding and MLP embedding components to process Gaussian positions prior to rendering, rather than applying them to our
feature vectors, and adding their pixel alignment loss. We followed FMGS's default parameters, however in order to conduct a fair comparison, we increased the number of iterations in their warmup phase and increased the number of iterations in the feature optimization phase by the same ratio. Specifically, instead of using 30K iterations for warmup and 4.2K iterations for the feature optimization phase, we ran 70K iterations for warmup and $4.2K \cdot \frac{70}{30}=9.8K$ iterations for feature optimization, and for the Badshahi scene we ran 200K iterations for warmup and $4.2K \cdot \frac{200}{30}=28K$ iterations for feature optimization.

\subsubsection{LangSplat}
We used the code provided by authors in \url{https://github.com/minghanqin/LangSplat} with the default parameters.

\subsection{Semantic Bottleneck Ablation Details}
\label{sec:ablation_details}

We evaluate this ablation on the Milano Cathedral scene at 1/4 original image resolution (compared to 1/2 resolution used in standard experiments) to accommodate memory limitations of the alternative feature encoding methods. For each configuration, we maintain all proposed improvements including the Attenuated Downsampler Module and regularization losses while isolating only the feature encoding procedure for comparison. We train for 30,000 iterations with an additional 15,000 step iteration warm-up stage dedicated to RGB-only optimization.

\subsection{Evaluation Details}
\rev{Evaluation in the main paper is reported over the test sets extracted in \cite{dudai2024halo}. To also assess generalization to novel viewpoints, we constructed a validation subset from the full HolyScenes dataset. The original test set contains different ground-truth images over each semantic classes (e.g., views present in one class may be absent in another). To create the new validation set, we select up to five ground-truth images per class, prioritizing images shared across multiple classes within the same scene. This approach yields a balanced test set of 52 images with maximally overlapping viewpoints across different semantic categories. The subset contains 11 images for the \emph{Blue Mosque} scene, 9 for the \emph{Badshahi Mosque} scene, 8 for \emph{Milano} scene, 7 for \emph{St.~Paul} scene, 6 for \emph{Notre Dame} scene, and 11 for \emph{Hurba} scene. %
}

%% file: figures/ablation_results/clip_pyramids_scales/clip_pyramid_scales.tex
\begin{figure}[htbp]
    \centering
    \hbox{
        \hbox{
            \includegraphics[height=0.168\textwidth]{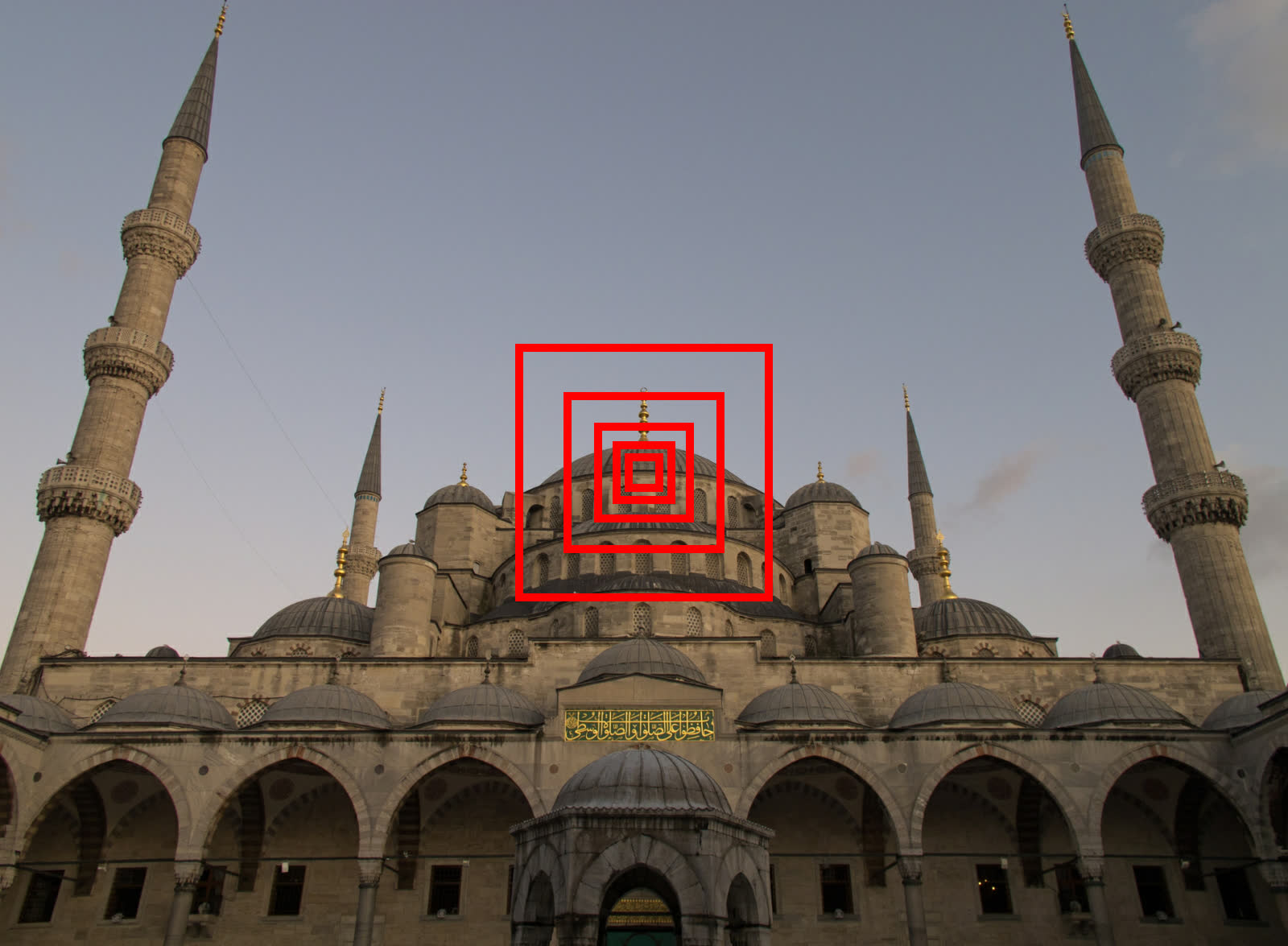}
            \includegraphics[height=0.168\textwidth]{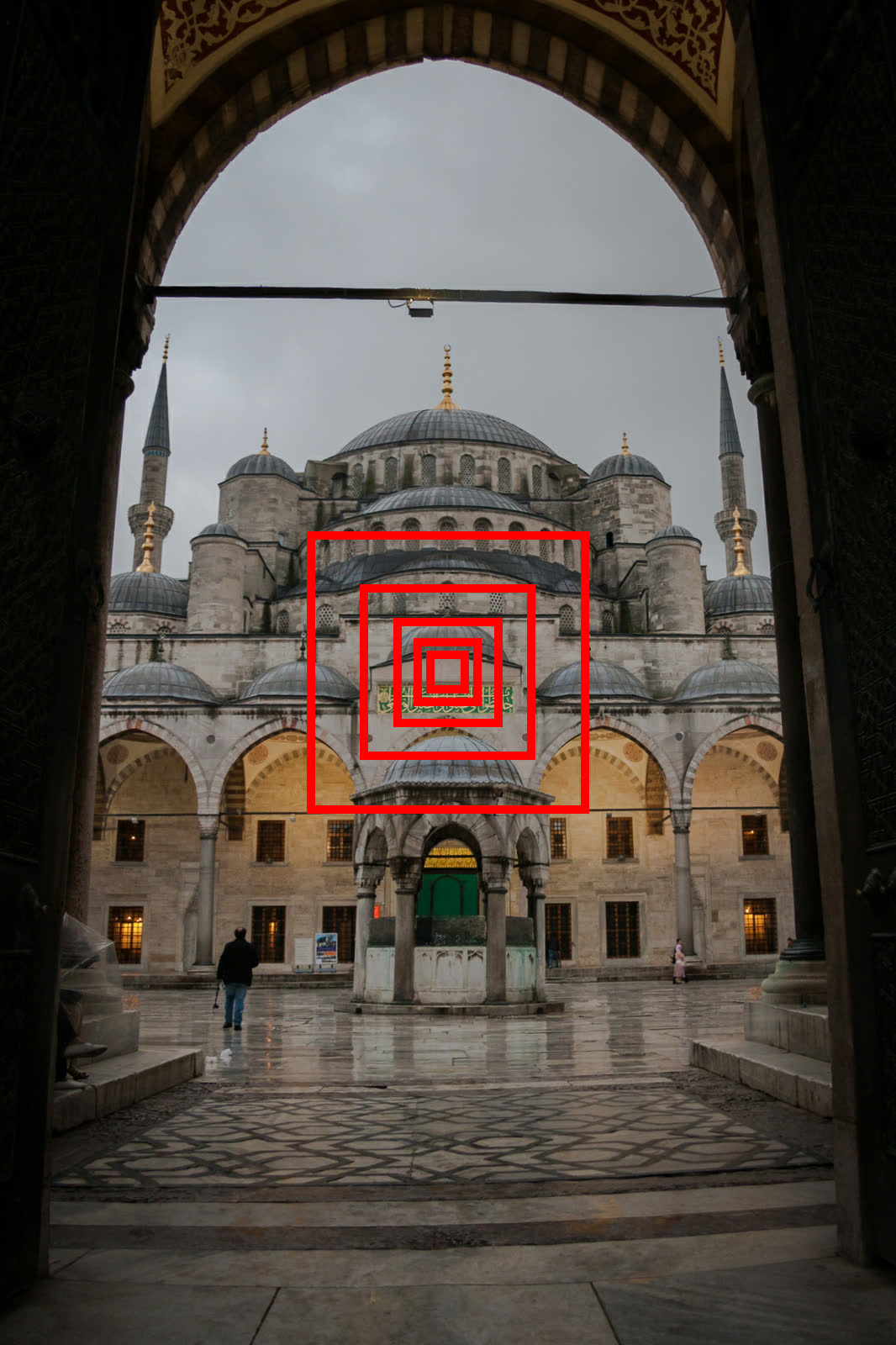}
            \includegraphics[height=0.168\textwidth]{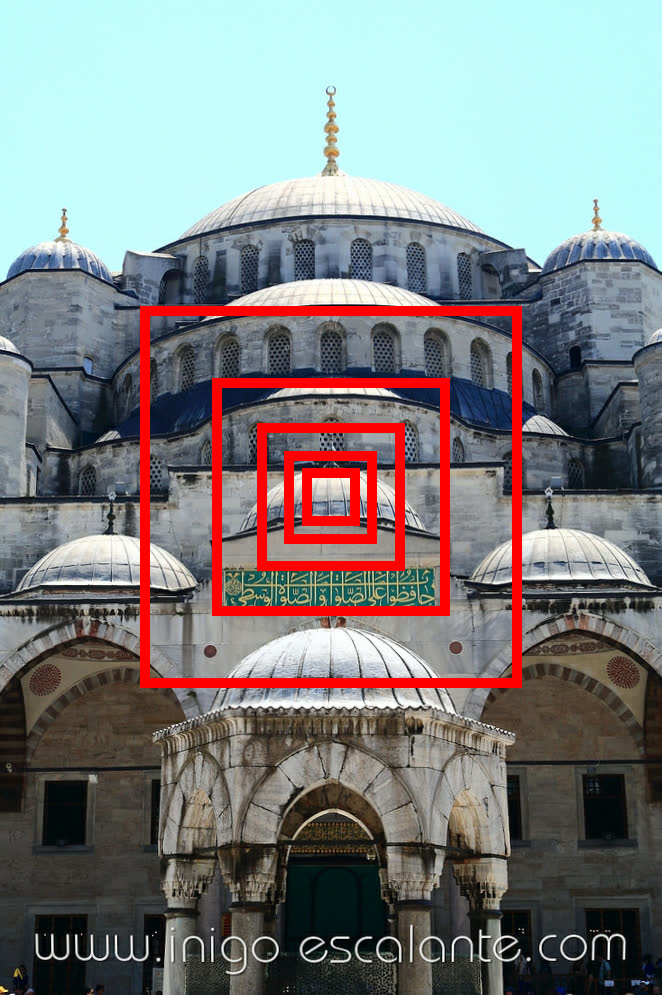}
        }
    }
    \hbox{
        \hbox{
            \includegraphics[height=0.16\textwidth]{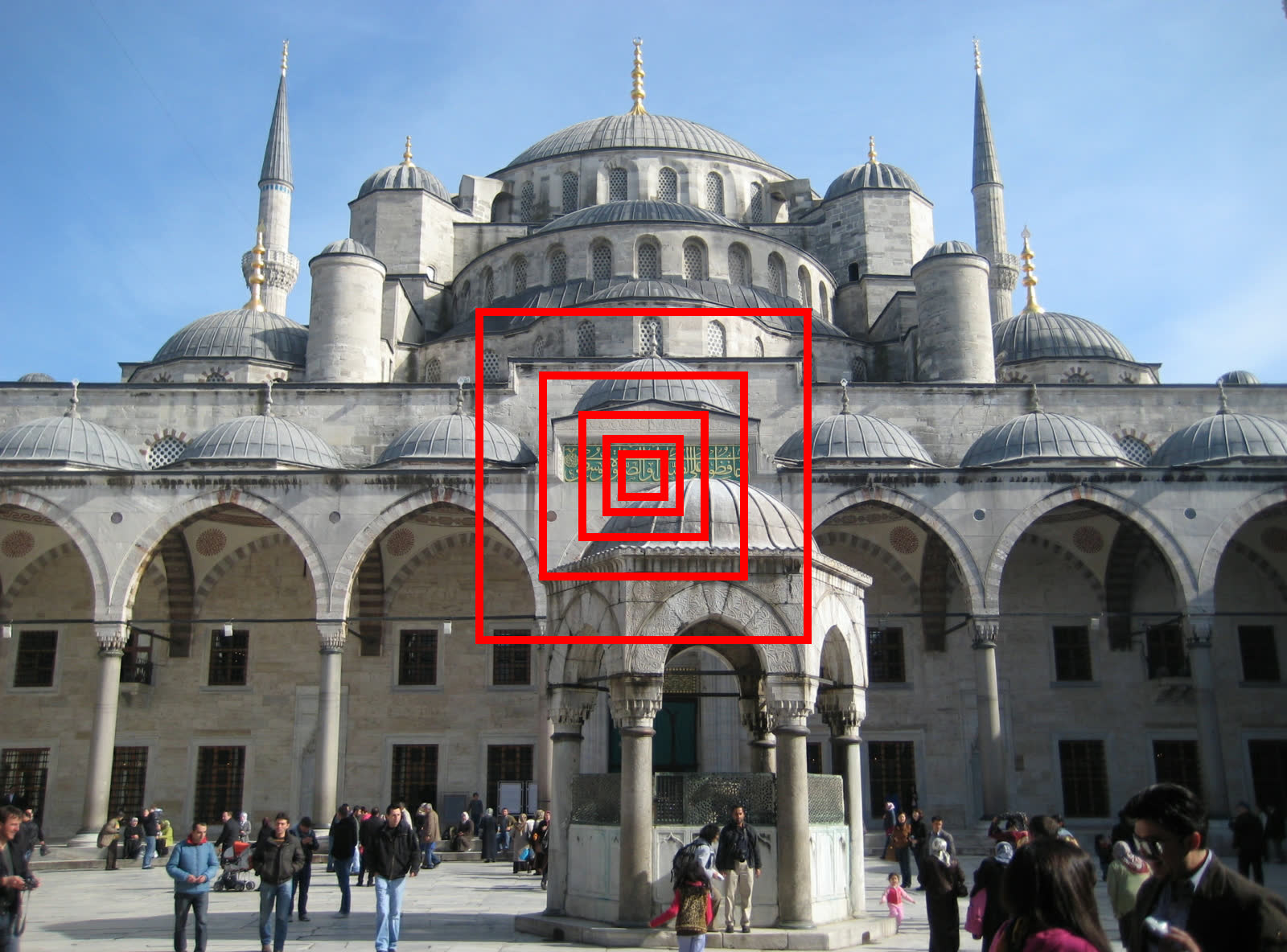}
            \includegraphics[height=0.16\textwidth]{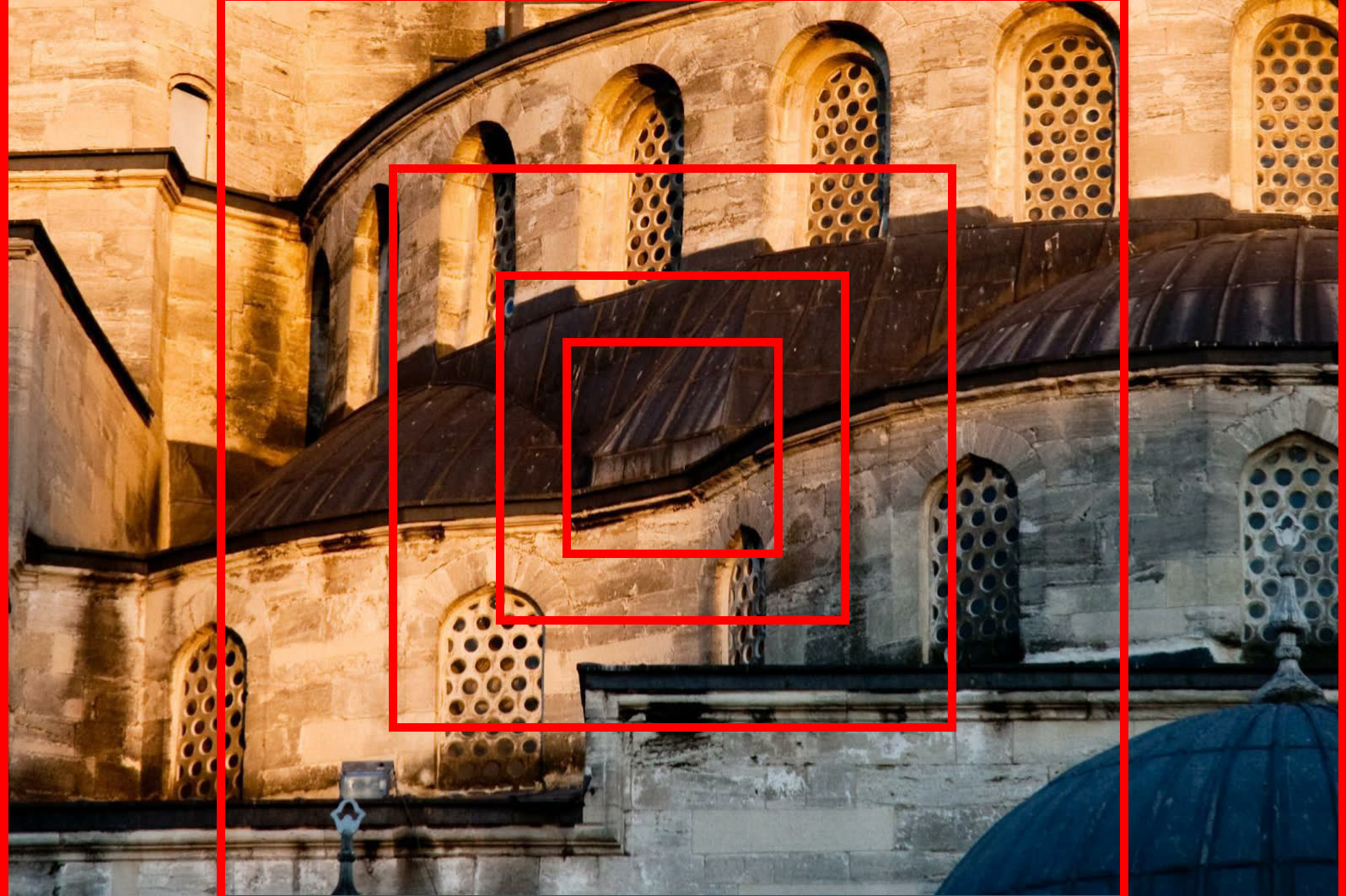}
        }
    }
    \caption{Visualization of our CLIP Pyramids Physical Scales in images with different FOV. Notice that by using physical scales, the smallest crop is the size of a window even when looking at images with different FOV.}
    \label{fig:clip_pyramid_scales}
\end{figure}

%% file: supp/supp_additional_results.tex
\input{figures/final_results/more_results}
\input{figures/final_results/DL3DV-10K_results}
\input{figures/final_results/novel_views_results}
\input{figures/final_results/lerf_table}

\input{figures/final_results/lerf_results}

\section{Additional Visualizations and Results}
\label{sec:additional_vis_and_results}

\input{figures/clip_visualization/clip_visualization_bilinear}

\subsection{CLIP Pyramid Feature Misalignment}
\label{subsec:clip_pyramid_analysis}

Fig.~\ref{fig:ablations_temporal} illustrates the semantic misalignment associated with CLIP pyramid features. We visualize 2D CLIP features at multiple scales for St. Paul's Cathedral, showing PCA projections (top row) followed by probability heatmaps extracted using the standard relevancy score procedure from LERF~\cite{kerr2023lerf}. This feature extraction procedure suffers from semantic misalignments where architectural features leak to spatially adjacent areas. 

\subsection{Performance Breakdown}
\label{sec:ab}

We report performance per scene and category in Tab.~\ref{tab:per_category_results} for our method, as well as for HaLo-NeRF~\cite{dudai2024halo}. Additional qualitative results are presented in Fig.~\ref{fig:more_results}. %

\subsection{\rev{Evaluation on Additional Datasets}}
\label{sec:additional_datasets}
\rev{To further demonstrate the adaptability and robustness of our method, we evaluate on two additional datasets. The DL3DV-1K \cite{ling2023dl3dv10klargescalescenedataset} dataset contains large-scale real-world indoor environments with complex geometric structures and varying lighting conditions. We selected a representative scene and compared against Feature3DGS. Since this dataset lacks semantic annotations, we conducted a user study with 25 participants, who preferred our method in 77$\%$ of cases.  Figure~\ref{fig:dl3dv_results_figure} demonstrates our method's ability to precisely localize different prompts in these challenging indoor environments.
The LERF-OVS \cite{kerr2023lerf} dataset contains general indoor scenes. A quantitative evaluation on this dataset is provided in Table~\ref{tab:dataset_accuracy}. Our method achieves competitive performance with an average accuracy of 88$\%$, compared to 83$\%$ for LERF \cite{kerr2023lerf} and 93$\%$ for FMGS \cite{zuo2024fmgsfoundationmodelembedded}. In particular, we substantially outperform both baselines on the \textit{ramen} scene (95$\%$ vs. 63$\%$ and 90$\%$, respectively) and match FMGS on the \textit{bouquet} scene (achieving perfect accuracy). Fig.~\ref{fig:lerf_results} showcases our results on this dataset, demonstrating that our approach produces accurate semantic localizations.}

\subsection{\rev{Novel View Generalization}}
\label{sec:novel_view_rendering}

\rev{Figure~\ref{fig:novel_views_results_figure} illustrates our method's ability to render novel views along a dynamic camera trajectory, demonstrating the stability and consistency of synthesized features across varying perspectives.}

\input{tables/status}

%% file: figures/final_results/more_results.tex
\begin{figure}[t]
    \centering
    \begin{tabular}{ccc}
        \jsubfig{\includegraphics[height=1.76cm]{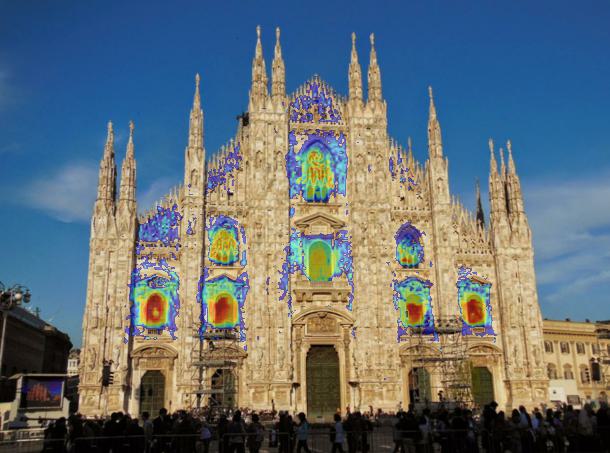}}{ {\small \textit{Windows }} } &
        \jsubfig{\includegraphics[height=1.76cm]{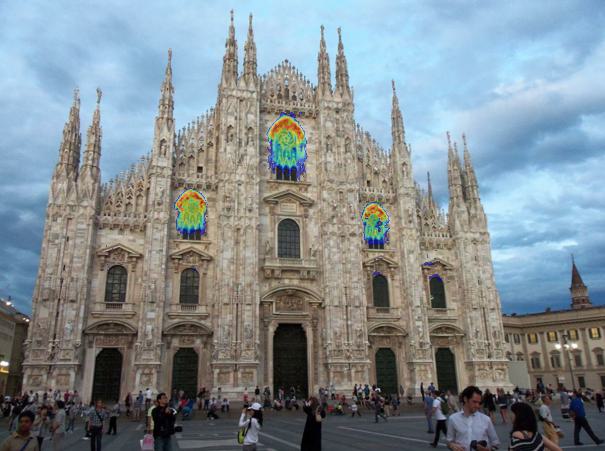}}{ {\small \textit{Rose Windows }} } &
        \jsubfig{\includegraphics[height=1.76cm]{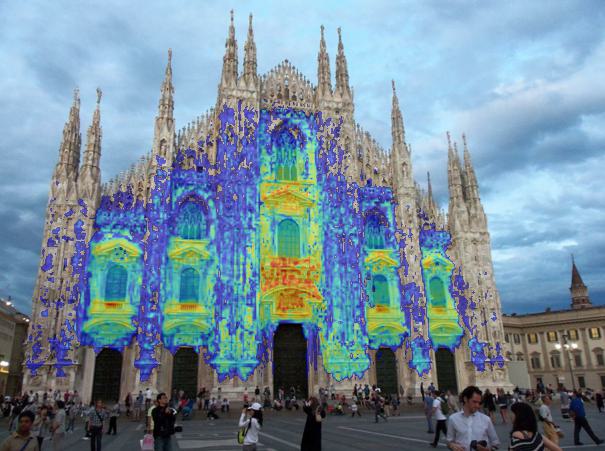}}{ {\small \textit{Lintel }} } \\[4pt]

        \jsubfig{\includegraphics[height=1.76cm]{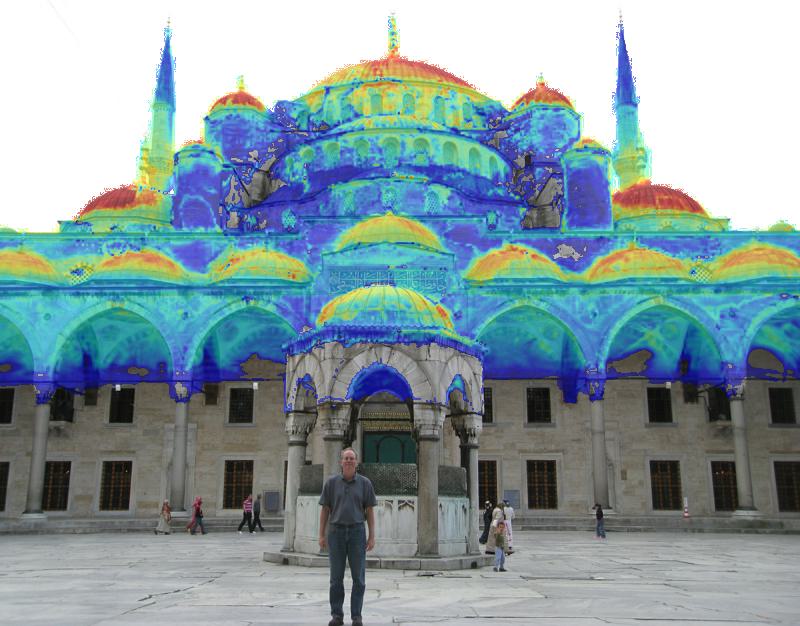}}{ {\small \textit{Domes }} } &
        \jsubfig{\includegraphics[height=1.76cm]{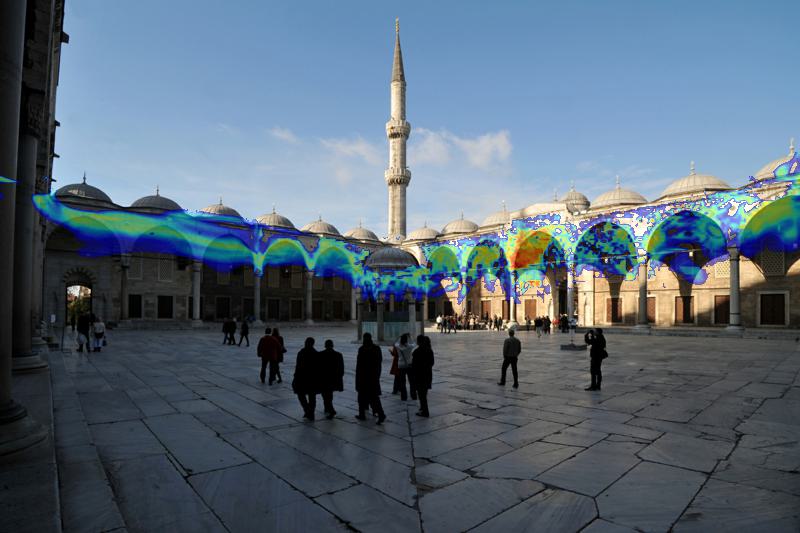}}{ {\small \textit{Ornamental Arches }} } &
        \jsubfig{\includegraphics[height=1.76cm]{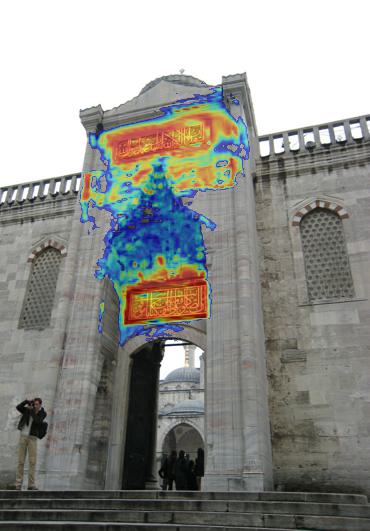}}{ {\small \textit{Calligraphy Panels }} } \\[4pt]

        \jsubfig{\includegraphics[height=1.76cm]{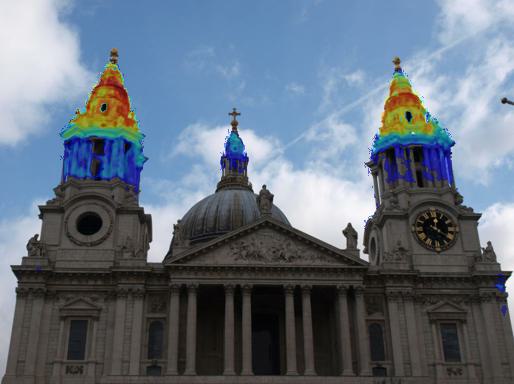}}{ {\small \textit{Towers }} } &
        \jsubfig{\includegraphics[height=1.76cm]{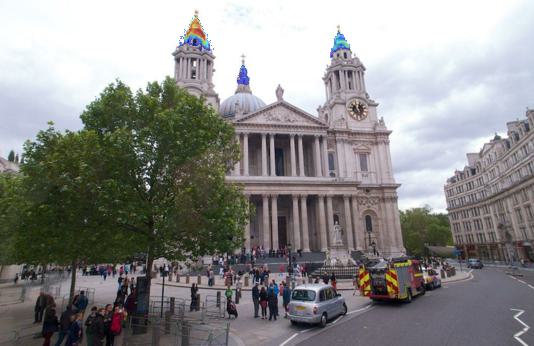}}{ {\small \textit{Pinnacles }} } &
        \jsubfig{\includegraphics[height=1.76cm]{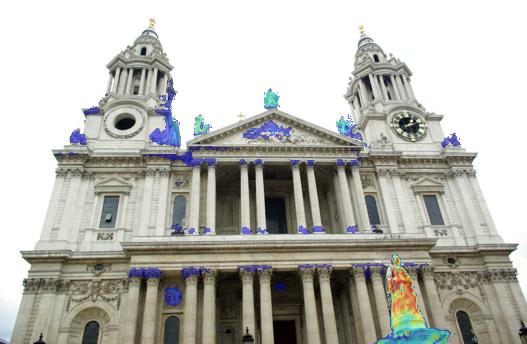}}{ {\small \textit{Statues }} }
    \end{tabular}

    \caption{\textbf{Additional Qualitative Results}. Above we illustrate additional results over diverse text prompts.%
    }
    \label{fig:more_results}
\end{figure}

%% file: figures/final_results/DL3DV-10K_results.tex
\begin{figure}
    \rotatebox{90}{\whitetxt{xxp}\footnotesize{\textit{Table}}}
    \jsubfig{\includegraphics[height=1.5cm]{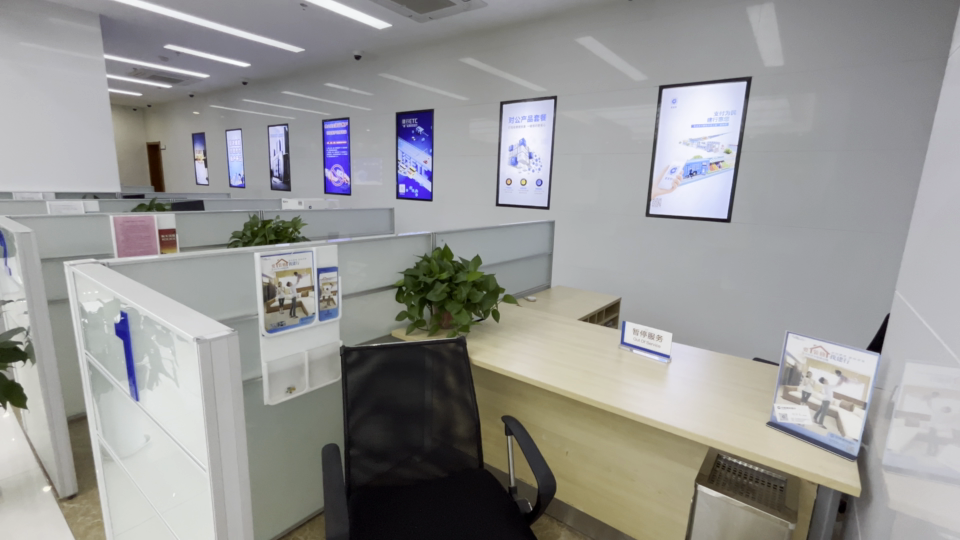}}{}
    \hfill
    \jsubfig{\includegraphics[height=1.5cm]{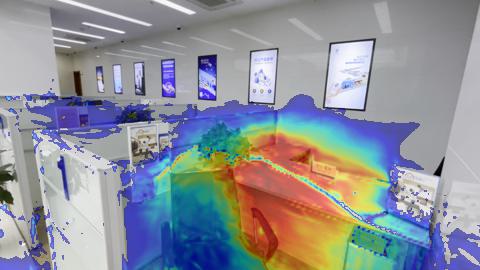}}{}
    \hfill
    \jsubfig{\includegraphics[height=1.5cm]{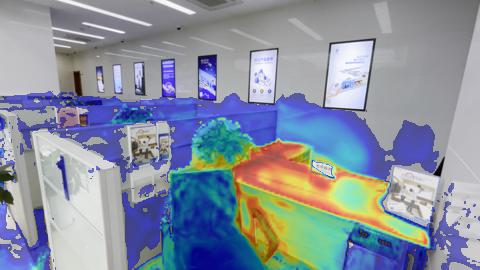}}{}
    \\
    \rotatebox{90}{\whitetxt{xxp}\footnotesize{\textit{Chair}}}
    \jsubfig{\includegraphics[height=1.5cm]{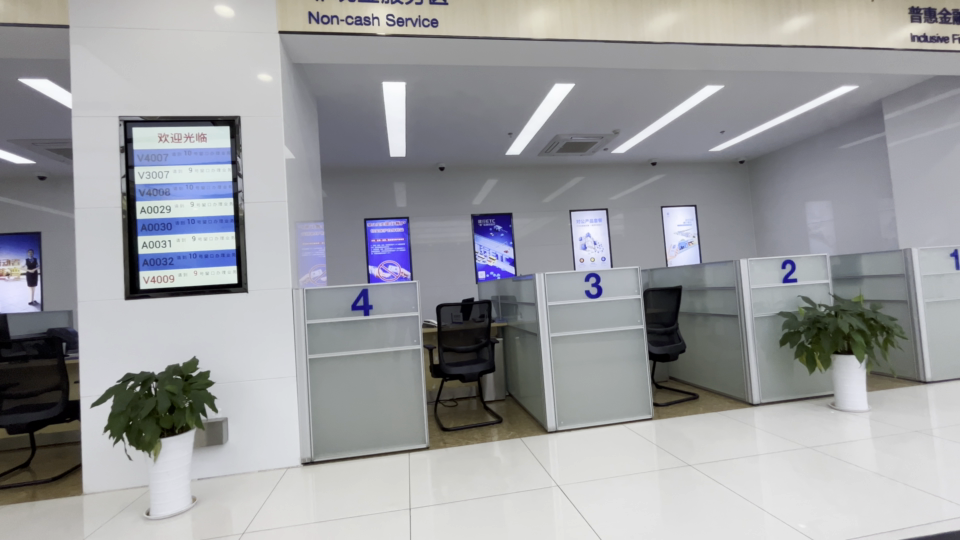}}{}
    \hfill
    \jsubfig{\includegraphics[height=1.5cm]{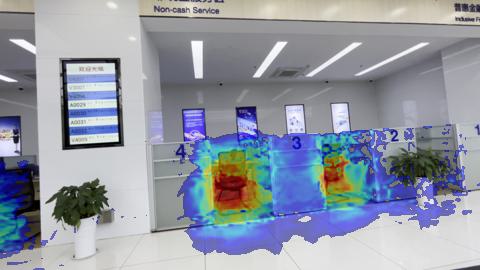}}{}
    \hfill
    \jsubfig{\includegraphics[height=1.5cm]{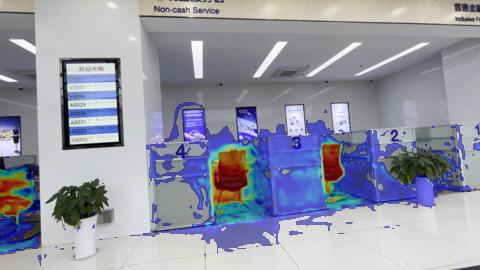}}{}
    \\
    \rotatebox{90}{\whitetxt{xp}\footnotesize{\textit{FL-lamp}}}
    \jsubfig{\includegraphics[height=1.5cm]{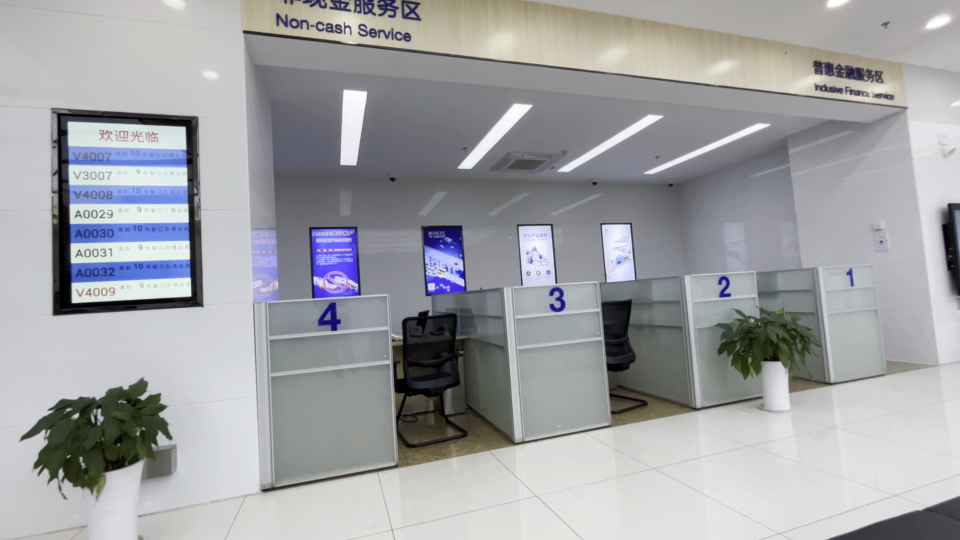}}{}
    \hfill
    \jsubfig{\includegraphics[height=1.5cm]{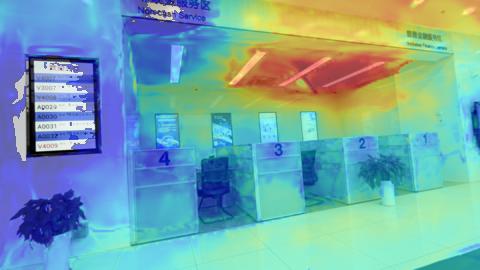}}{}
    \hfill
    \jsubfig{\includegraphics[height=1.5cm]{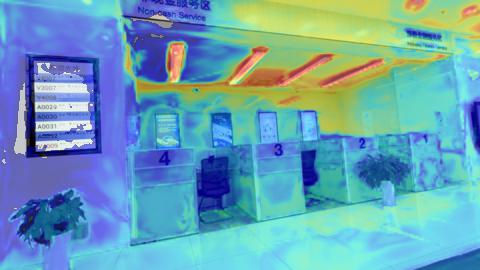}}{}
    \\
    \rotatebox{90}{\whitetxt{xp}\footnotesize{\textit{Trash can}}}
    \jsubfig{\includegraphics[height=1.5cm]{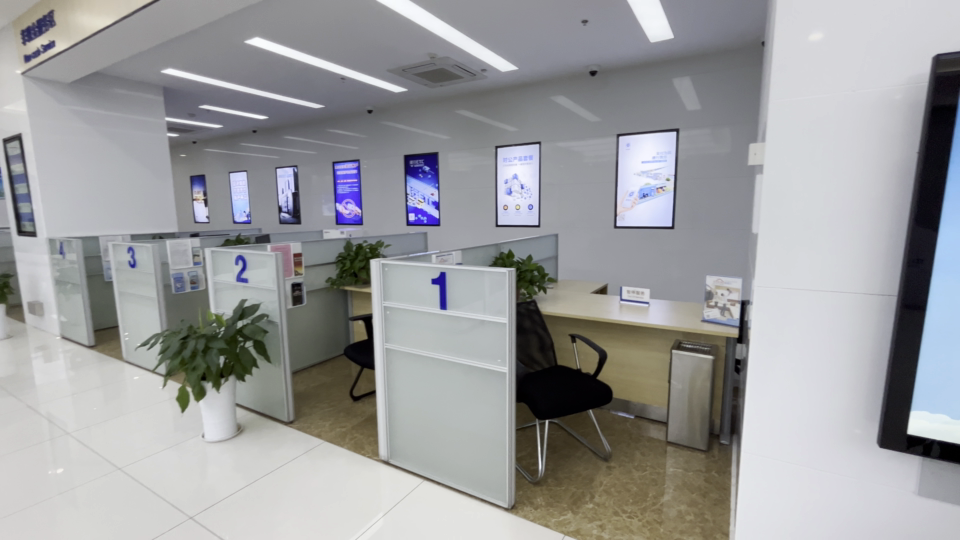}}{}
    \hfill
    \jsubfig{\includegraphics[height=1.5cm]{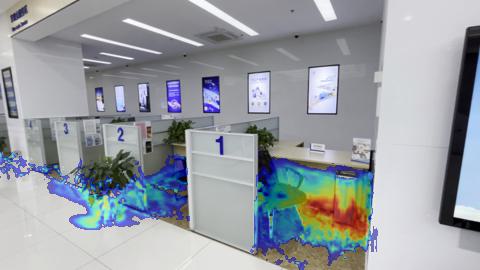}}{}
    \hfill
    \jsubfig{\includegraphics[height=1.5cm]{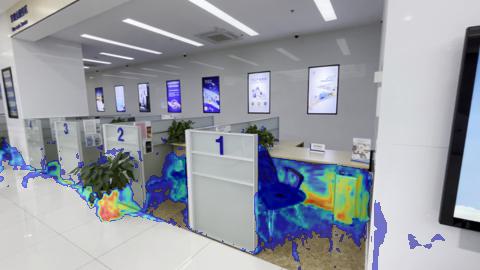}}{}
    \\
    \rotatebox{90}{\whitetxt{xp}\footnotesize{\textit{Numbers}}}
    \jsubfig{\includegraphics[height=1.5cm]{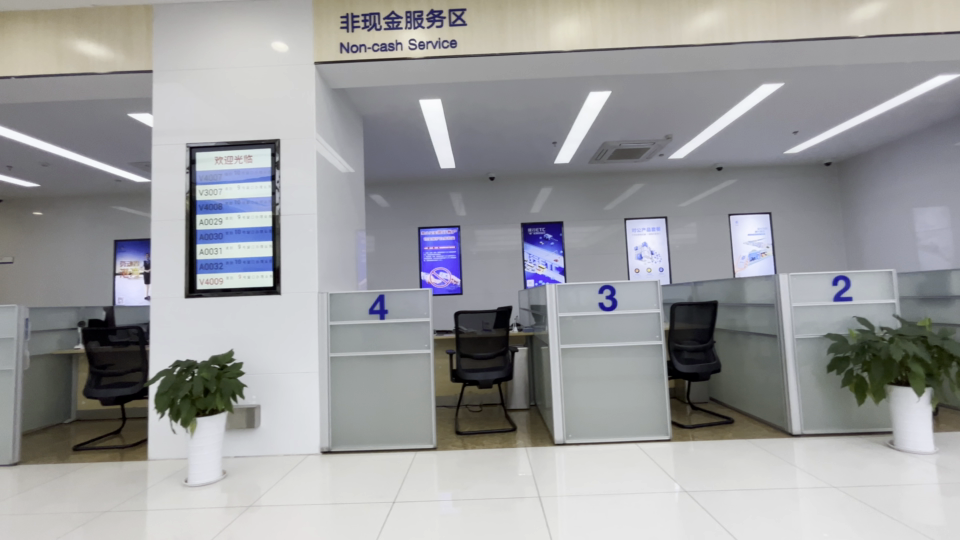}}{}
    \hfill
    \jsubfig{\includegraphics[height=1.5cm]{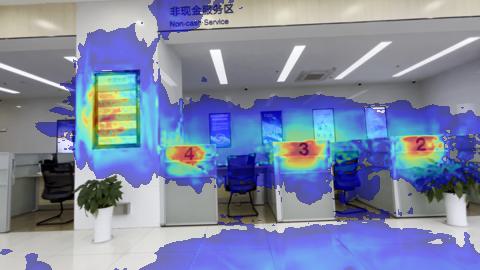}}{}
    \hfill
    \jsubfig{\includegraphics[height=1.5cm]{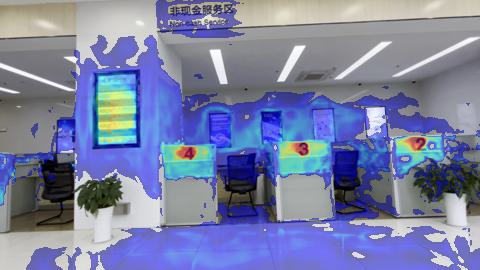}}{}
    \\
    \rotatebox{90}{\whitetxt{xxp}\footnotesize{\textit{Vent}}}
    \jsubfig{\includegraphics[height=1.5cm]{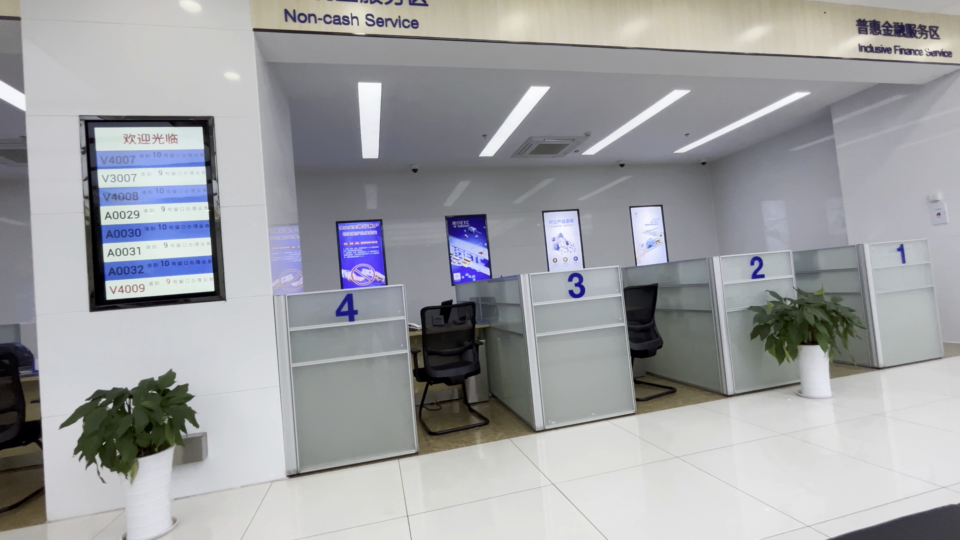}}{}
    \hfill
    \jsubfig{\includegraphics[height=1.5cm]{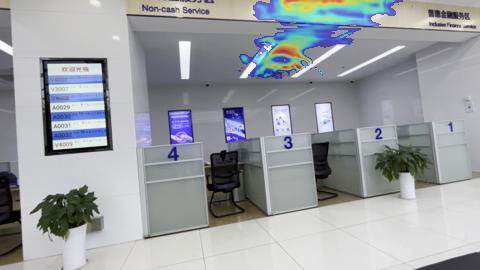}}{}
    \hfill
    \jsubfig{\includegraphics[height=1.5cm]{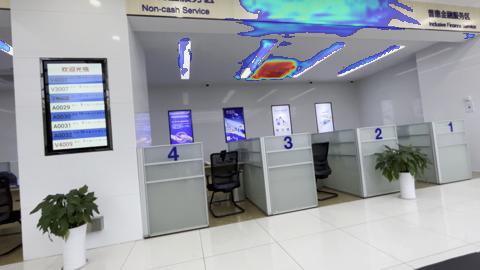}}{}
    \\
    \rotatebox{90}{\whitetxt{xxp}\footnotesize{\textit{Plant}}}
    \jsubfig{\includegraphics[height=1.5cm]{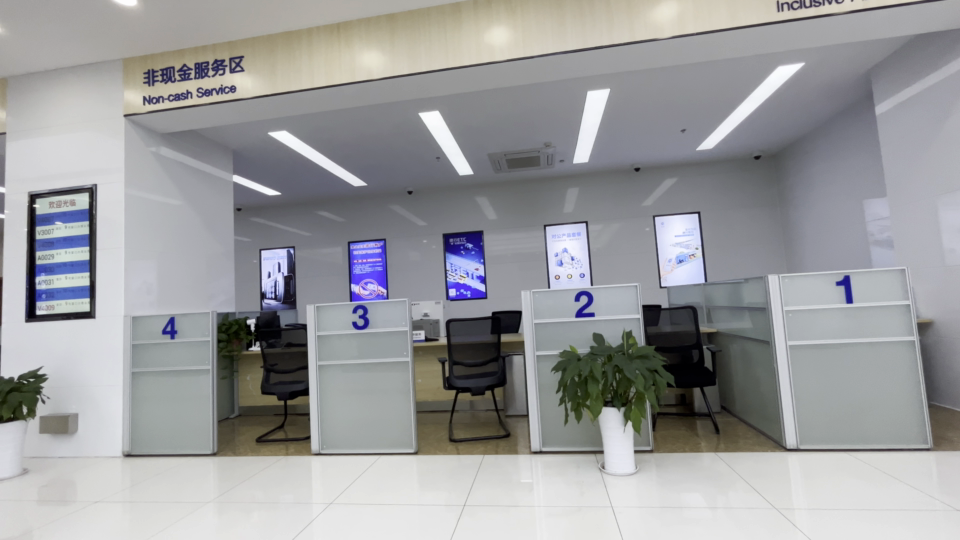}}{\textit{Image}}
    \hfill
    \jsubfig{\includegraphics[height=1.5cm]{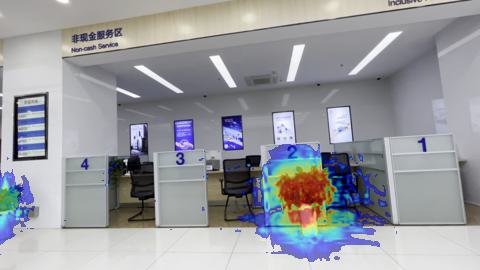}}{\textit{Feature3DGS}}
    \hfill
    \jsubfig{\includegraphics[height=1.5cm]{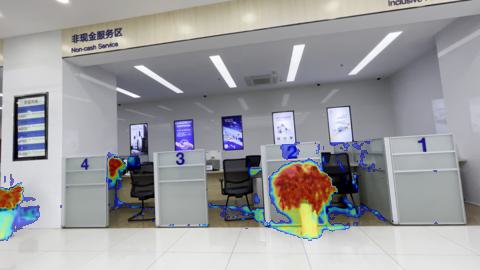}}{\textit{\methodname}}
    \caption{
        \rev{\textbf{Qualitative Comparison to Feature3DGS on the DL3DV dataset}. We compare our method against Feature3DGS across different semantic prompts in an indoor scene from the DL3DV dataset. Feature3DGS struggles to precisely localize specific objects within the indoor environment. In contrast, our methods achiecves precise and accurate localizations.
        }
    }
    \label{fig:dl3dv_results_figure}
\end{figure}

%% file: figures/final_results/novel_views_results.tex
\begin{figure*}
    \rotatebox{90}{\whitetxt{xxxp}\footnotesize{\textit{PCA}}}
    \jsubfig{\includegraphics[height=1.8cm]{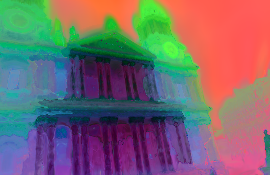}}{}
    \hfill
    \jsubfig{\includegraphics[height=1.8cm]{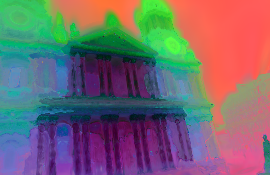}}{}
    \hfill
    \jsubfig{\includegraphics[height=1.8cm]{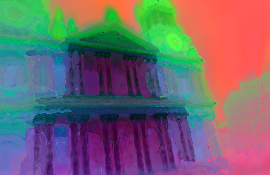}}{}
    \hfill
    \jsubfig{\includegraphics[height=1.8cm]{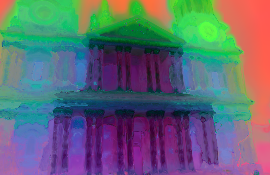}}{}
    \hfill
    \jsubfig{\includegraphics[height=1.8cm]{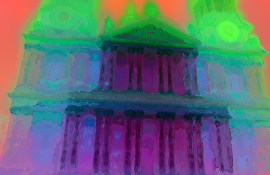}}{}
    \hfill
    \jsubfig{\includegraphics[height=1.8cm]{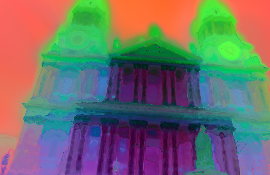}}{}
    \\
    \rotatebox{90}{\whitetxt{xxp}\footnotesize{\textit{Windows}}}
    \jsubfig{\includegraphics[height=1.8cm]{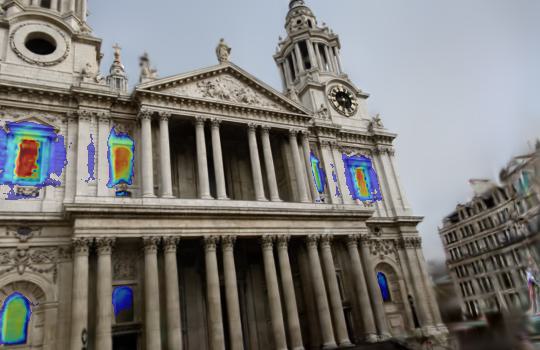}}{}
    \hfill
    \jsubfig{\includegraphics[height=1.8cm]{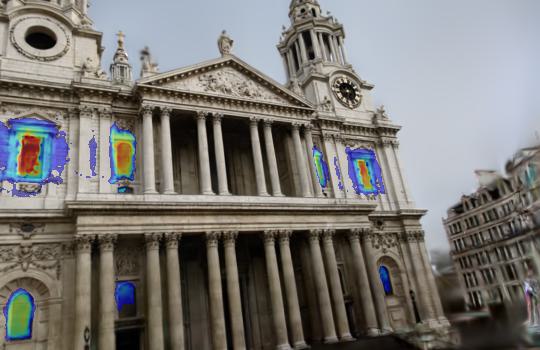}}{}
    \hfill
    \jsubfig{\includegraphics[height=1.8cm]{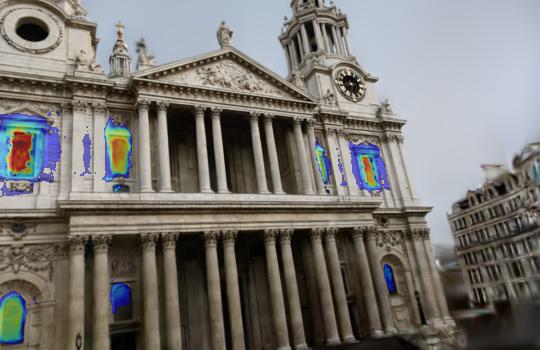}}{}
    \hfill
    \jsubfig{\includegraphics[height=1.8cm]{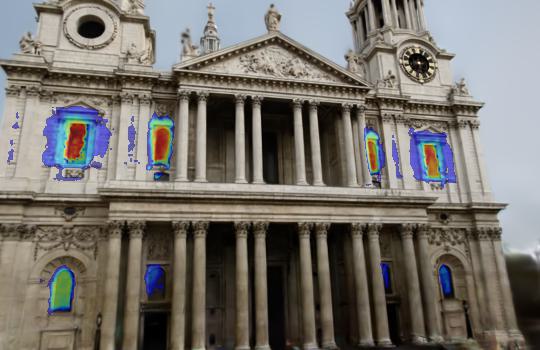}}{}
    \hfill
    \jsubfig{\includegraphics[height=1.8cm]{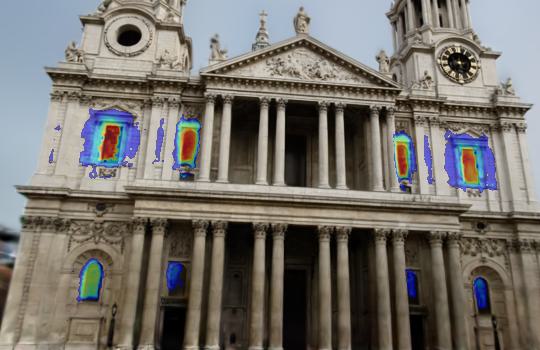}}{}
    \hfill
    \jsubfig{\includegraphics[height=1.8cm]{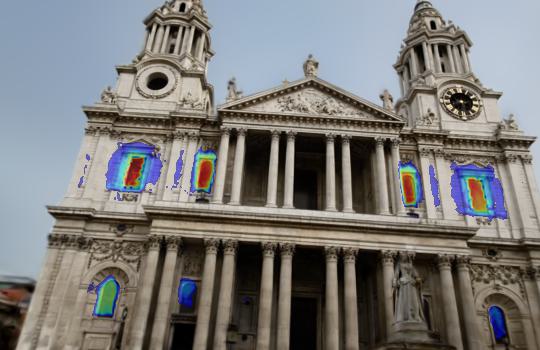}}{}
    \\
    \rotatebox{90}{\whitetxt{xxxp}\footnotesize{\textit{Towers}}}
    \jsubfig{\includegraphics[height=1.8cm]{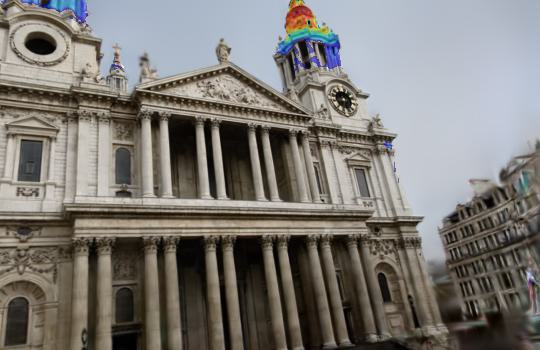}}{}
    \hfill
    \jsubfig{\includegraphics[height=1.8cm]{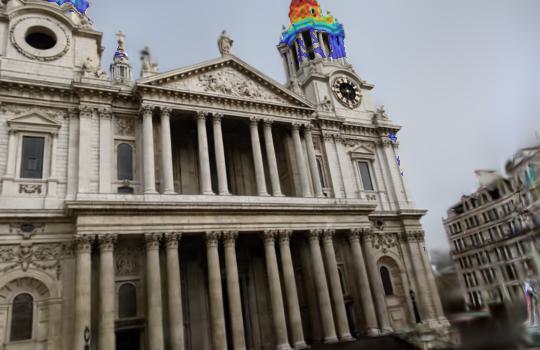}}{}
    \hfill
    \jsubfig{\includegraphics[height=1.8cm]{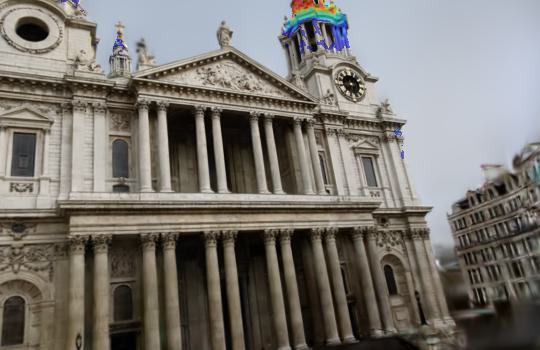}}{}
    \hfill
    \jsubfig{\includegraphics[height=1.8cm]{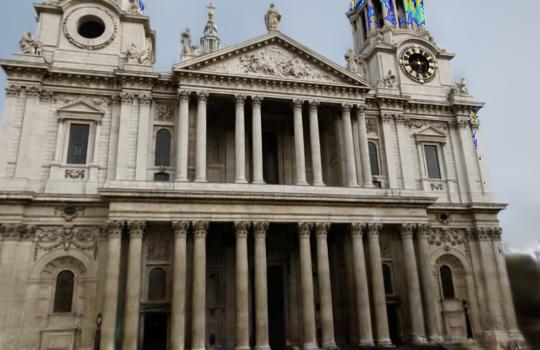}}{}
    \hfill
    \jsubfig{\includegraphics[height=1.8cm]{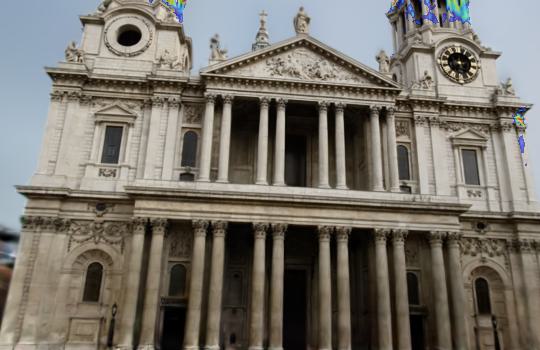}}{}
    \hfill
    \jsubfig{\includegraphics[height=1.8cm]{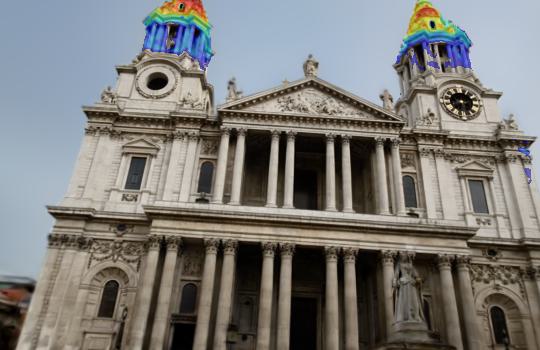}}{}
    \\
    \rotatebox{90}{\whitetxt{xxxp}\footnotesize{\textit{Relief}}}
    \jsubfig{\includegraphics[height=1.8cm]{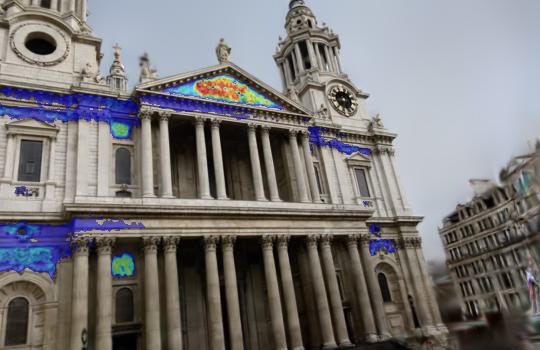}}{}
    \hfill
    \jsubfig{\includegraphics[height=1.8cm]{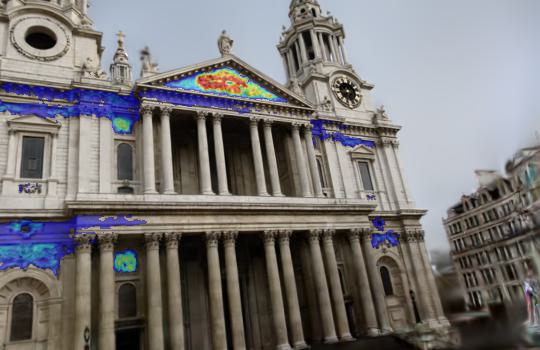}}{}
    \hfill
    \jsubfig{\includegraphics[height=1.8cm]{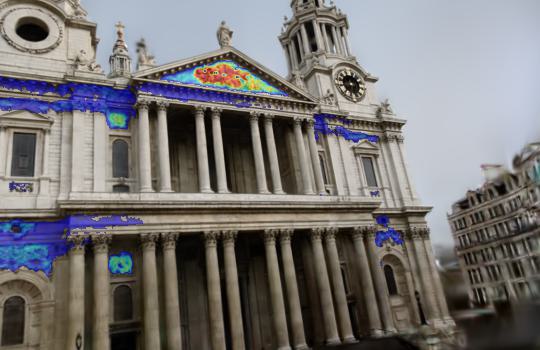}}{}
    \hfill
    \jsubfig{\includegraphics[height=1.8cm]{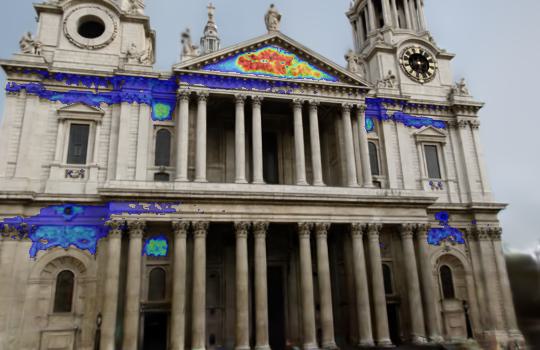}}{}
    \hfill
    \jsubfig{\includegraphics[height=1.8cm]{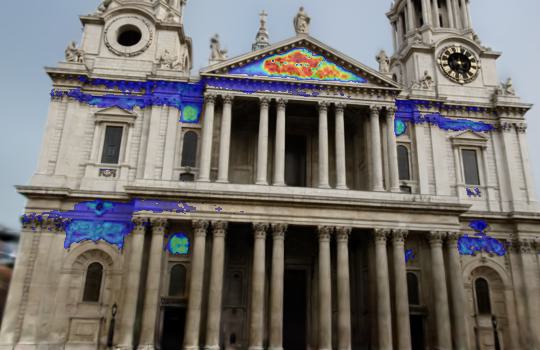}}{}
    \hfill
    \jsubfig{\includegraphics[height=1.8cm]{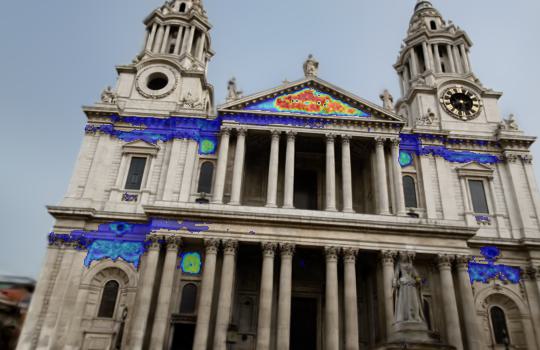}}{}
    \\
    \rotatebox{90}{\whitetxt{xxxp}\footnotesize{\textit{PCA}}}
    \jsubfig{\includegraphics[height=1.725cm]{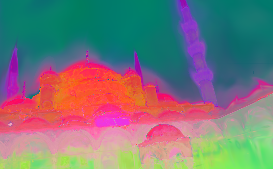}}{}
    \hfill
    \jsubfig{\includegraphics[height=1.725cm]{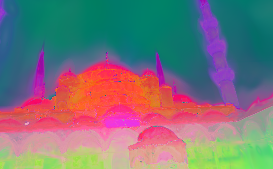}}{}
    \hfill
    \jsubfig{\includegraphics[height=1.725cm]{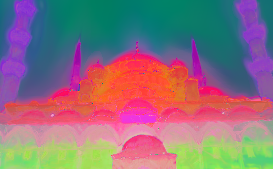}}{}
    \hfill
    \jsubfig{\includegraphics[height=1.725cm]{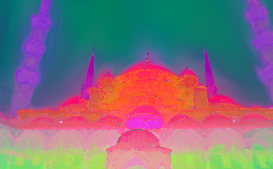}}{}
    \hfill
    \jsubfig{\includegraphics[height=1.725cm]{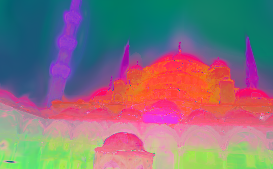}}{}
    \hfill
    \jsubfig{\includegraphics[height=1.725cm]{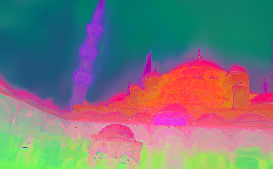}}{}
    \\
    \rotatebox{90}{\whitetxt{xxxp}\footnotesize{\textit{Domes}}}
    \jsubfig{\includegraphics[height=1.725cm]{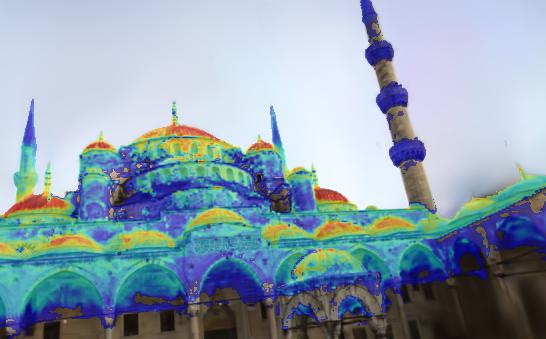}}{}
    \hfill
    \jsubfig{\includegraphics[height=1.725cm]{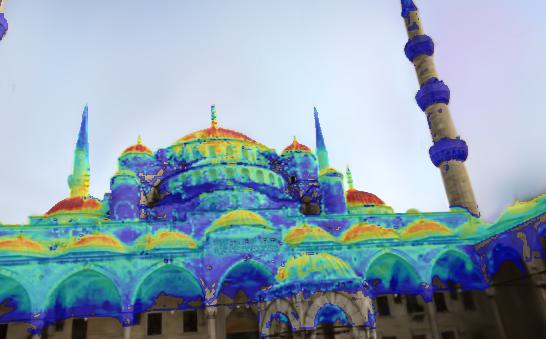}}{}
    \hfill
    \jsubfig{\includegraphics[height=1.725cm]{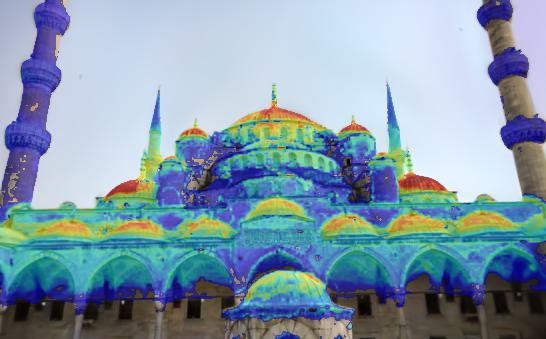}}{}
    \hfill
    \jsubfig{\includegraphics[height=1.725cm]{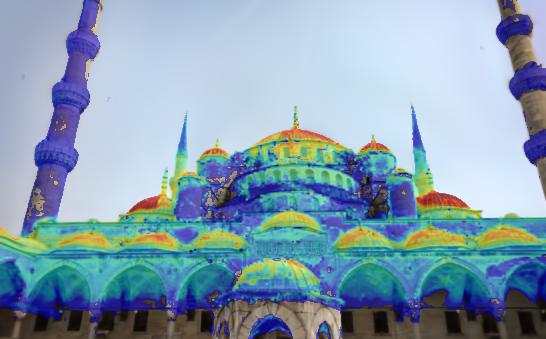}}{}
    \hfill
    \jsubfig{\includegraphics[height=1.725cm]{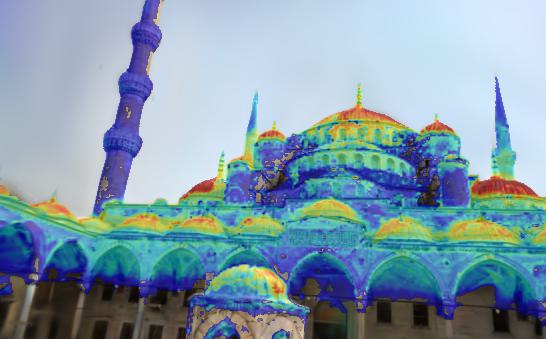}}{}
    \hfill
    \jsubfig{\includegraphics[height=1.725cm]{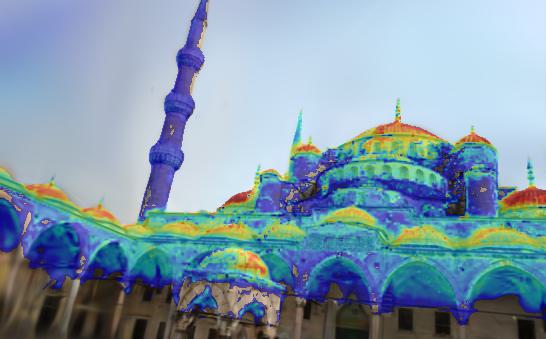}}{}
    \\
    \rotatebox{90}{\whitetxt{xxp}\footnotesize{\textit{Minarets}}}
    \jsubfig{\includegraphics[height=1.725cm]{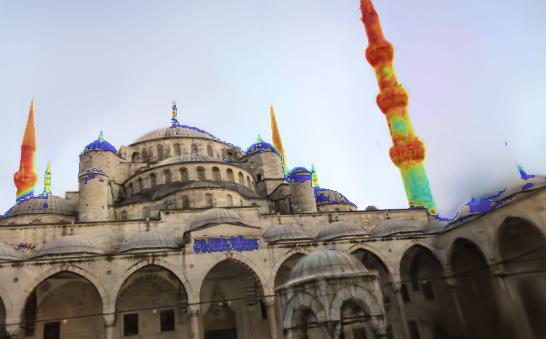}}{}
    \hfill
    \jsubfig{\includegraphics[height=1.725cm]{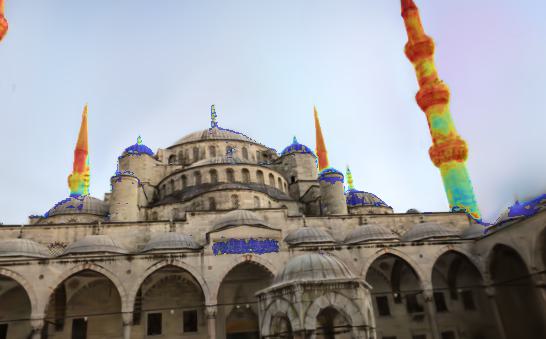}}{}
    \hfill
    \jsubfig{\includegraphics[height=1.725cm]{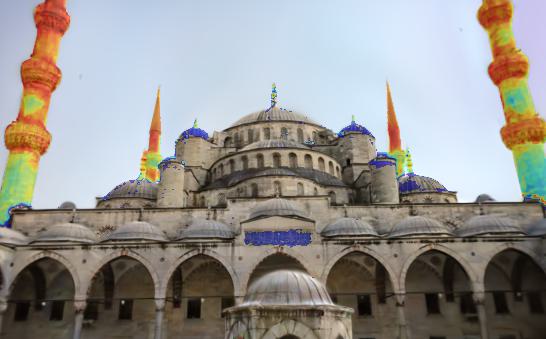}}{}
    \hfill
    \jsubfig{\includegraphics[height=1.725cm]{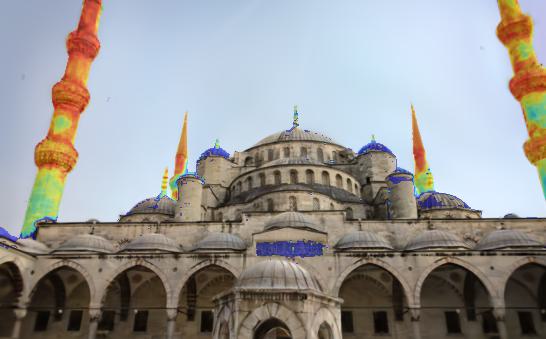}}{}
    \hfill
    \jsubfig{\includegraphics[height=1.725cm]{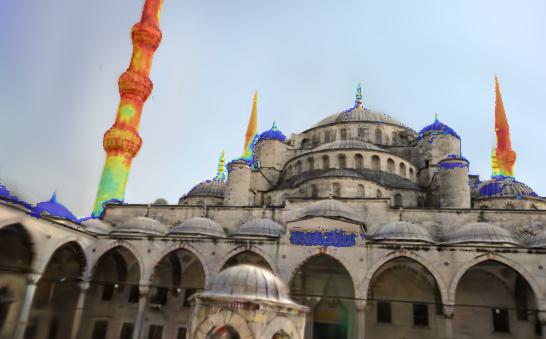}}{}
    \hfill
    \jsubfig{\includegraphics[height=1.725cm]{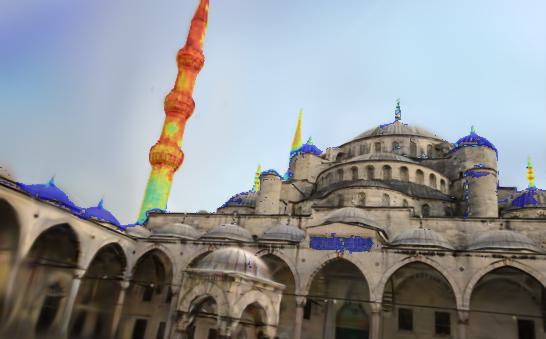}}{}
    \\
    \rotatebox{90}{\whitetxt{xp}\footnotesize{\textit{Caligraphy}}}
    \jsubfig{\includegraphics[height=1.725cm]{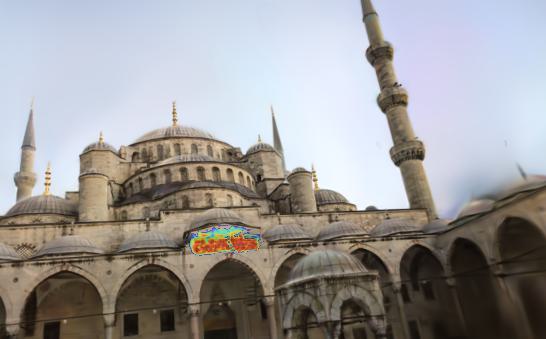}}{}
    \hfill
    \jsubfig{\includegraphics[height=1.725cm]{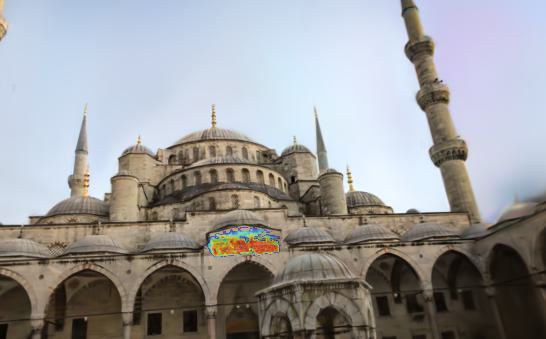}}{}
    \hfill
    \jsubfig{\includegraphics[height=1.725cm]{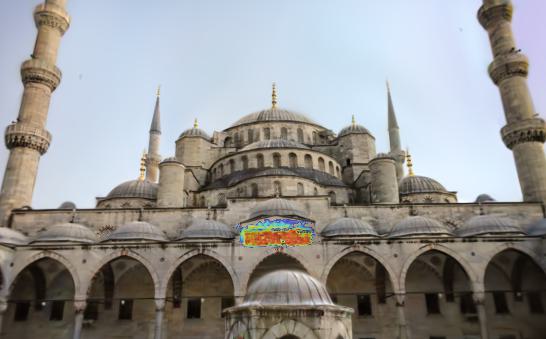}}{}
    \hfill
    \jsubfig{\includegraphics[height=1.725cm]{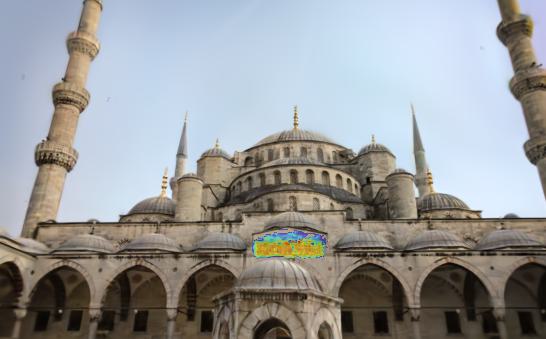}}{}
    \hfill
    \jsubfig{\includegraphics[height=1.725cm]{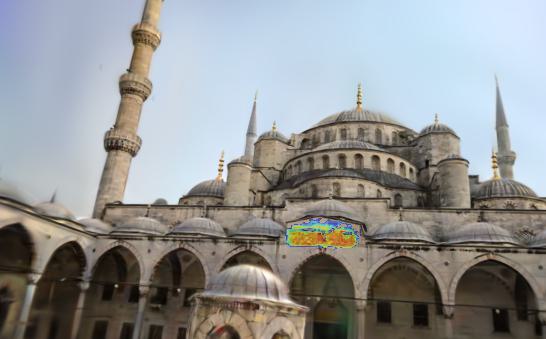}}{}
    \hfill
    \jsubfig{\includegraphics[height=1.725cm]{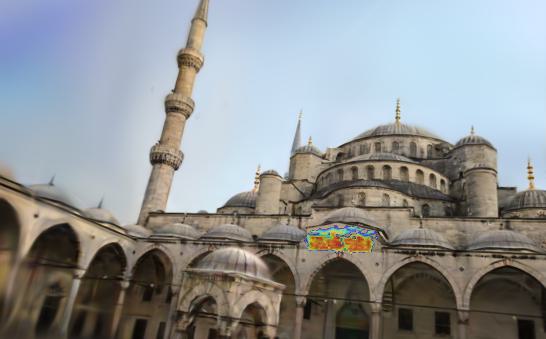}}{}
    \\
    \rotatebox{90}{\whitetxt{xxxp}\footnotesize{\textit{PCA}}}
    \jsubfig{\includegraphics[height=1.75cm]{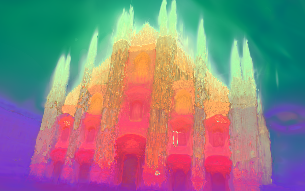}}{}
    \hfill
    \jsubfig{\includegraphics[height=1.75cm]{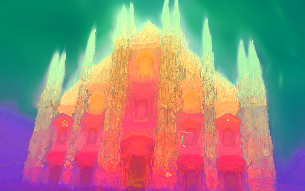}}{}
    \hfill
    \jsubfig{\includegraphics[height=1.75cm]{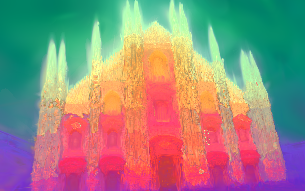}}{}
    \hfill
    \jsubfig{\includegraphics[height=1.75cm]{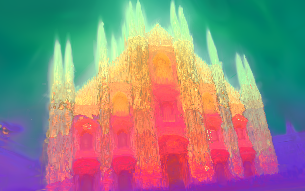}}{}
    \hfill
    \jsubfig{\includegraphics[height=1.75cm]{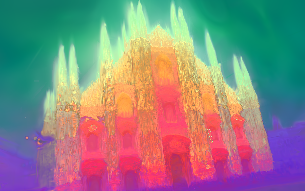}}{}
    \hfill
    \jsubfig{\includegraphics[height=1.75cm]{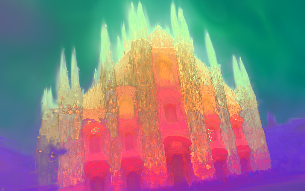}}{}
    \\
    \rotatebox{90}{\whitetxt{xxp}\footnotesize{\textit{Windows}}}
    \jsubfig{\includegraphics[height=1.75cm]{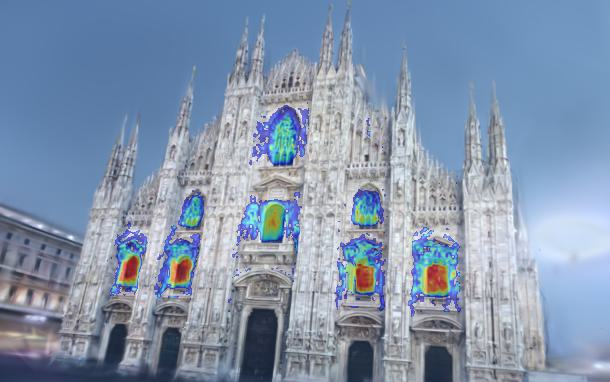}}{}
    \hfill
    \jsubfig{\includegraphics[height=1.75cm]{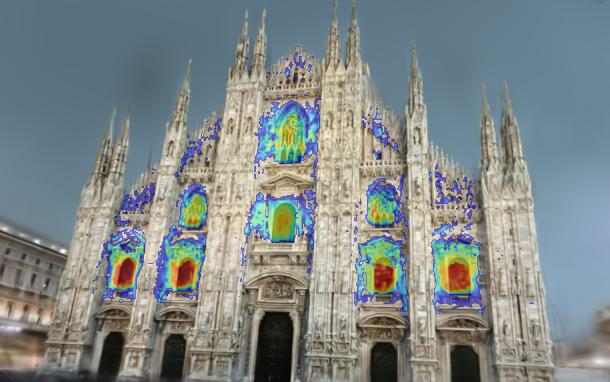}}{}
    \hfill
    \jsubfig{\includegraphics[height=1.75cm]{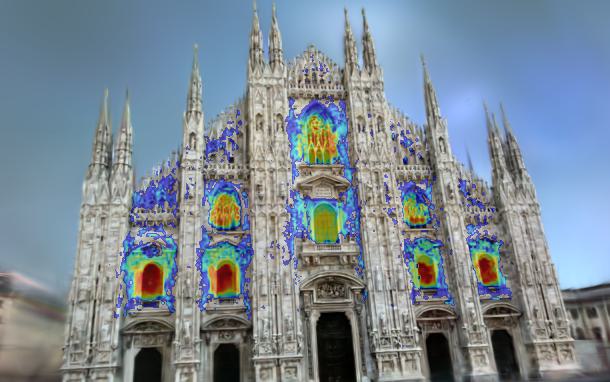}}{}
    \hfill
    \jsubfig{\includegraphics[height=1.75cm]{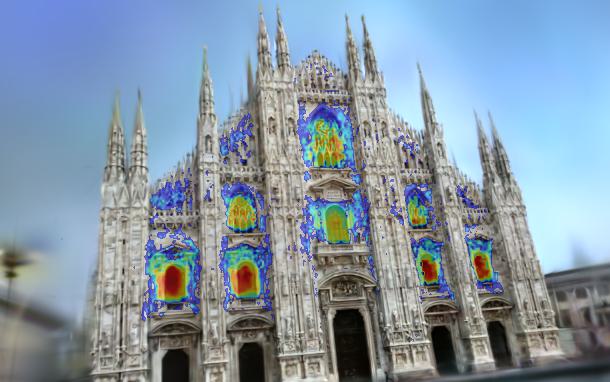}}{}
    \hfill
    \jsubfig{\includegraphics[height=1.75cm]{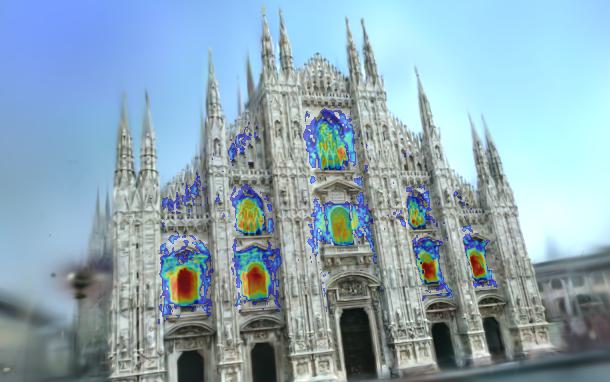}}{}
    \hfill
    \jsubfig{\includegraphics[height=1.75cm]{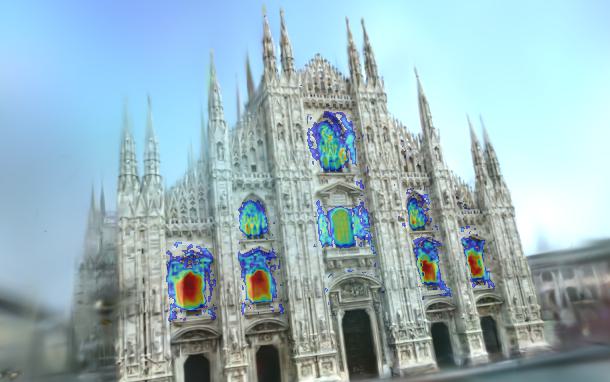}}{}
    \\
    \rotatebox{90}{\whitetxt{xxp}\footnotesize{\textit{Spires}}}
    \jsubfig{\includegraphics[height=1.75cm]{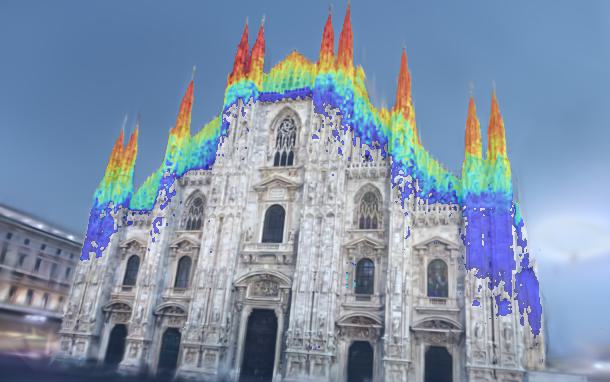}}{}
    \hfill
    \jsubfig{\includegraphics[height=1.75cm]{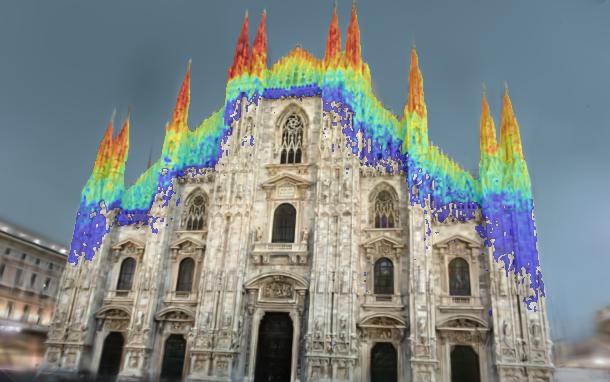}}{}
    \hfill
    \jsubfig{\includegraphics[height=1.75cm]{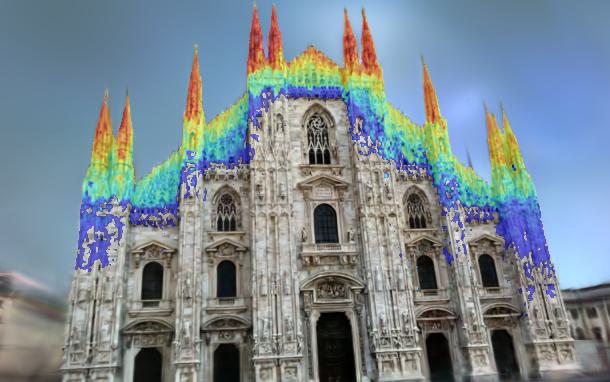}}{}
    \hfill
    \jsubfig{\includegraphics[height=1.75cm]{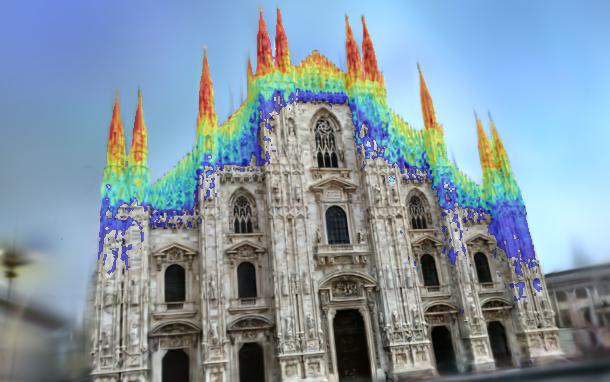}}{}
    \hfill
    \jsubfig{\includegraphics[height=1.75cm]{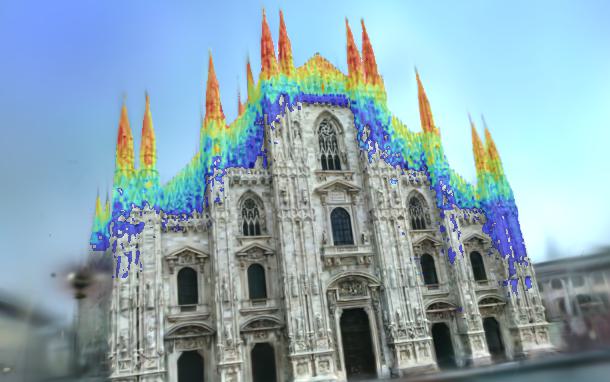}}{}
    \hfill
    \jsubfig{\includegraphics[height=1.75cm]{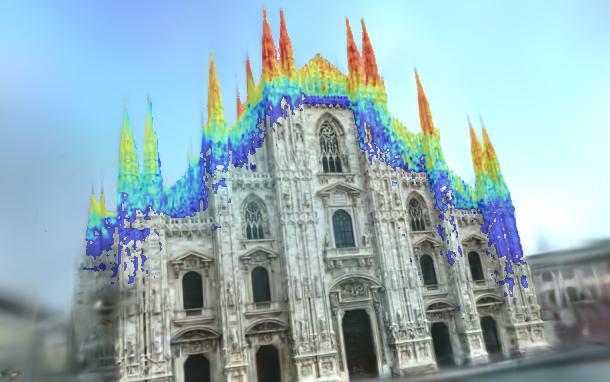}}{}
    
    \caption{
        \rev{\textbf{Novel View Generalization Results.} We demonstrate our method's ability to maintain feature consistency across viewpoints by rendering novel views along a dynamic camera trajectory. The figure shows semantic localization results for the text queries listed on the left, along with PCA visualizations of the learned features. Both the segmentation maps and the PCA features exhibit strong coherence across different viewpoints, confirming the stability of our 3D feature representation under camera motion and our method's ability to generalize under novel views.}
    }
    \label{fig:novel_views_results_figure}
\end{figure*}

%% file: figures/final_results/lerf_table.tex
\begin{table}[t]
\centering

\caption{\rev{\textbf{Quantitative Evaluation on the LERF-OVS dataset.} We report object localization accuracy across individual scenes, comparing our method to LERF and FMGS. Bold indicates the best performance per scene.}}

\setlength{\tabcolsep}{6pt}
\begin{tabular}{@{} lccccc @{}}
\hline
{\small Scene} & {\small \textbf{FFD-LSeg}} 
               & {\small \textbf{OWL-ViT}} 
               & {\small \textbf{LERF}} 
               & {\small \textbf{FMGS}} 
               & {\small \textbf{Ours}} \\
\hline
{\small Bouquet}   & 0.50 & 0.67 & 0.83 & \textbf{1.00} & \textbf{1.00} \\
{\small Figurines} & 0.09 & 0.39 & 0.87 & \textbf{0.90} & 0.79 \\
{\small Ramen}     & 0.15 & 0.93 & 0.63 & 0.90 & \textbf{0.95} \\
{\small Teatime}   & 0.28 & 0.75 & \textbf{0.97} & 0.94 & 0.94 \\
{\small Kitchen}   & 0.13 & 0.43 & 0.85 & \textbf{0.93} & 0.70 \\
\hline
\textbf{Avg} & 0.18 & 0.55 & 0.83 & \textbf{0.93} & 0.88 \\
\hline
\end{tabular}

\label{tab:dataset_accuracy}
\end{table}

%% file: figures/final_results/lerf_results.tex
\begin{figure*}[t]
    \centering
    \begin{tabular}{ccccc}
        \rotatebox{90}{\whitetxt{xxp}\footnotesize{\textit{Ramen}}}
        \jsubfig{\includegraphics[height=1.76cm]{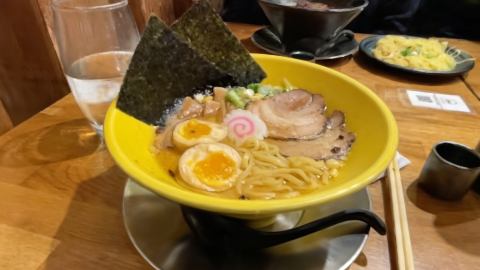}}{ {\small \textit{RGB}} } &
        \jsubfig{\includegraphics[height=1.76cm]{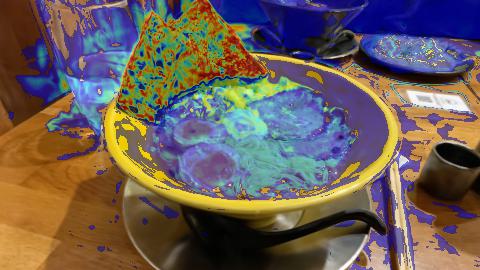}}{ {\small \textit{Nori}} } &
        \jsubfig{\includegraphics[height=1.76cm]{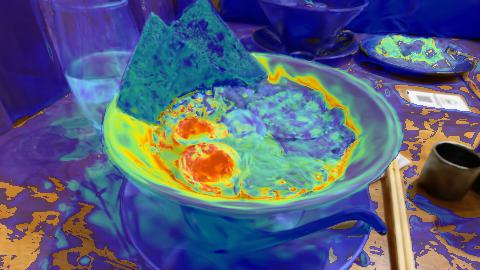}}{ {\small \textit{egg}} } &
        \jsubfig{\includegraphics[height=1.76cm]{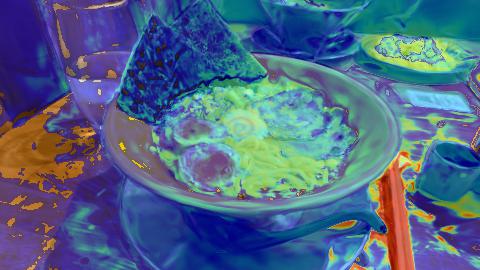}}{ {\small \textit{chopsticks}} } \\[4pt]

        \rotatebox{90}{\whitetxt{xp}\footnotesize{\textit{Figurines}}}
        \jsubfig{\includegraphics[height=1.76cm]{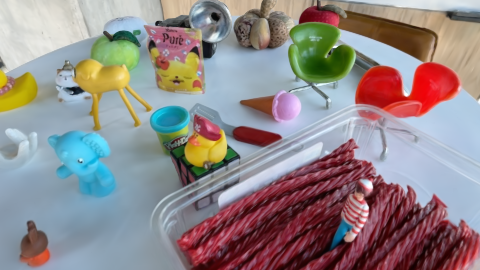}}{ {\small \textit{RGB}} } &
        \jsubfig{\includegraphics[height=1.76cm]{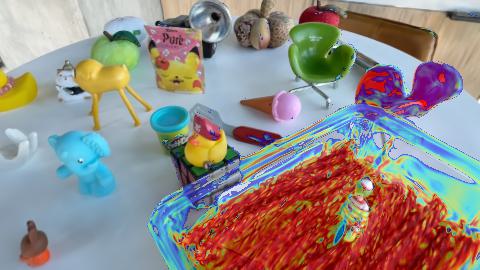}}{ {\small \textit{twizzlers}} } &
        \jsubfig{\includegraphics[height=1.76cm]{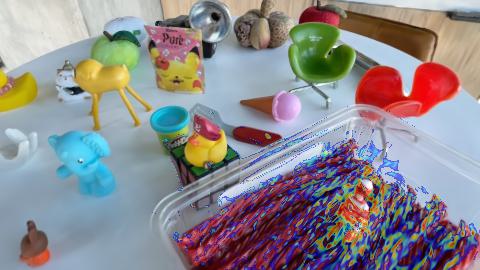}}{ {\small \textit{waldo}} } &
        \jsubfig{\includegraphics[height=1.76cm]{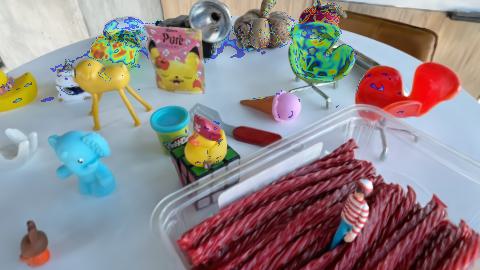}}{ {\small \textit{green apple}} } \\[4pt]

        \rotatebox{90}{\whitetxt{xp}\footnotesize{\textit{Teatime}}}
        \jsubfig{\includegraphics[height=1.76cm]{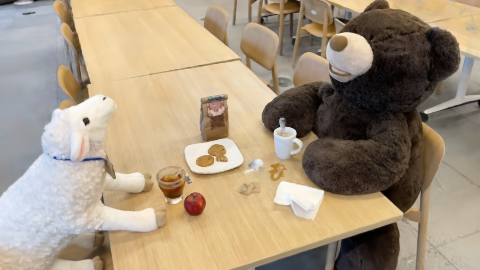}}{ {\small \textit{RGB}} } &
        \jsubfig{\includegraphics[height=1.76cm]{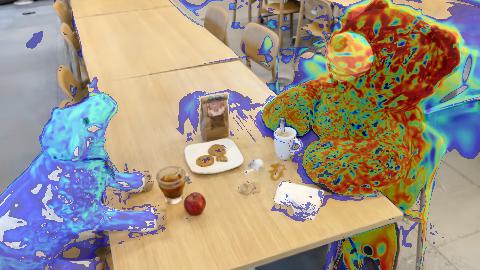}}{ {\small \textit{stuffed bear}} } &
        \jsubfig{\includegraphics[height=1.76cm]{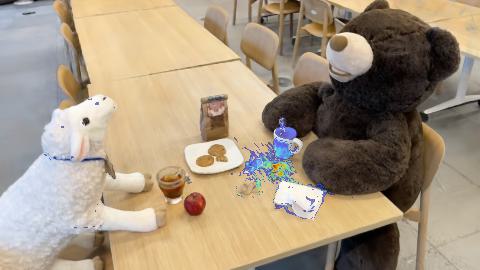}}{ {\small \textit{spill}} } &
        \jsubfig{\includegraphics[height=1.76cm]{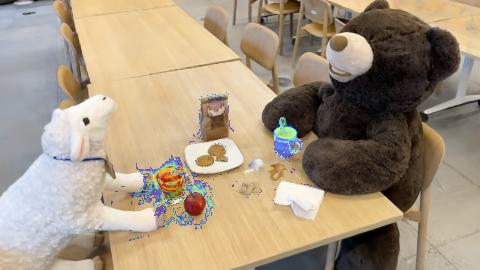}}{ {\small \textit{tea in a glass}} } \\[4pt]
    \end{tabular}

    \caption{\rev{\textbf{Object Localization Results on LERF-OVS}. We illustrate object localization results of our method across diverse indoor scenes and prompts from the LERF-OVS dataset. Our approach effectively localizes the various semantic concepts.
    }}
    \label{fig:lerf_results}
\end{figure*}

%% file: figures/clip_visualization/clip_visualization_bilinear.tex
\begin{figure*} %
\centering
\rotatebox{90}{\whitetxt{ssssssss}\texttt{PCA}}
\jsubfig{\includegraphics[height=3cm]{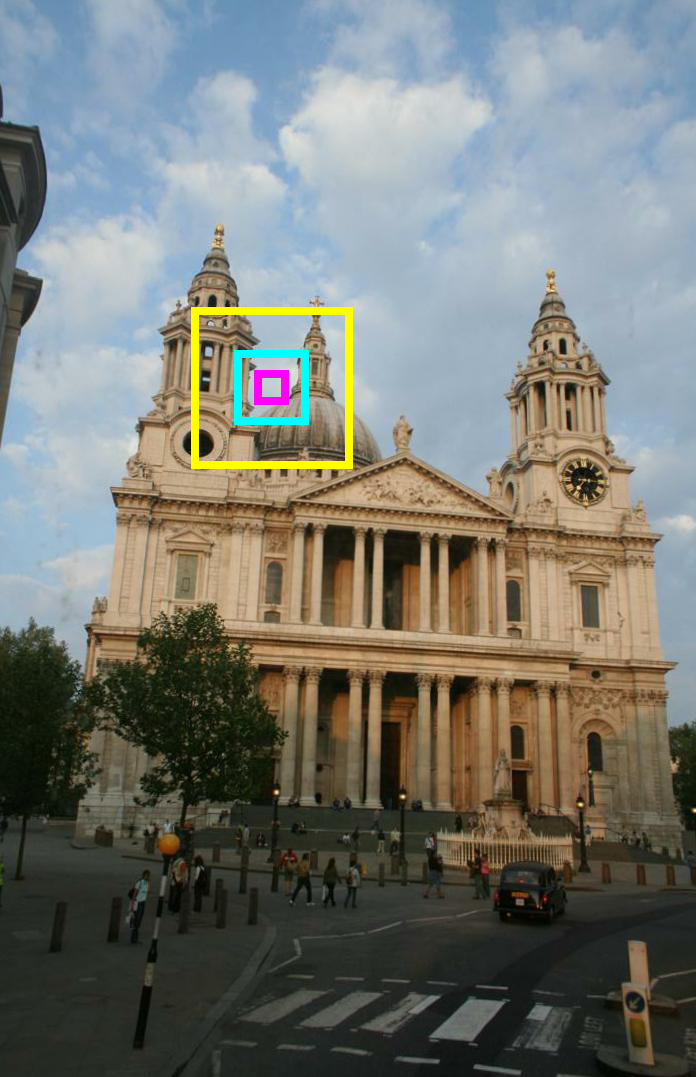}
\fcolorbox{magenta}{magenta}{\includegraphics[height=3cm]{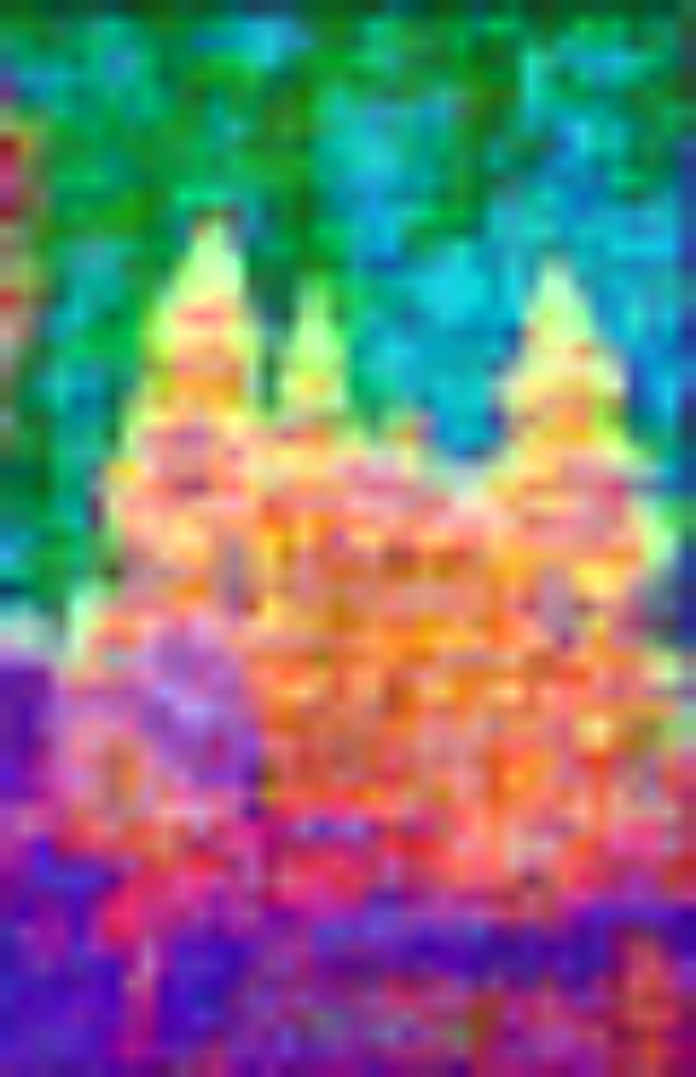}}
\fcolorbox{cyan}{cyan}{
\includegraphics[height=3cm]{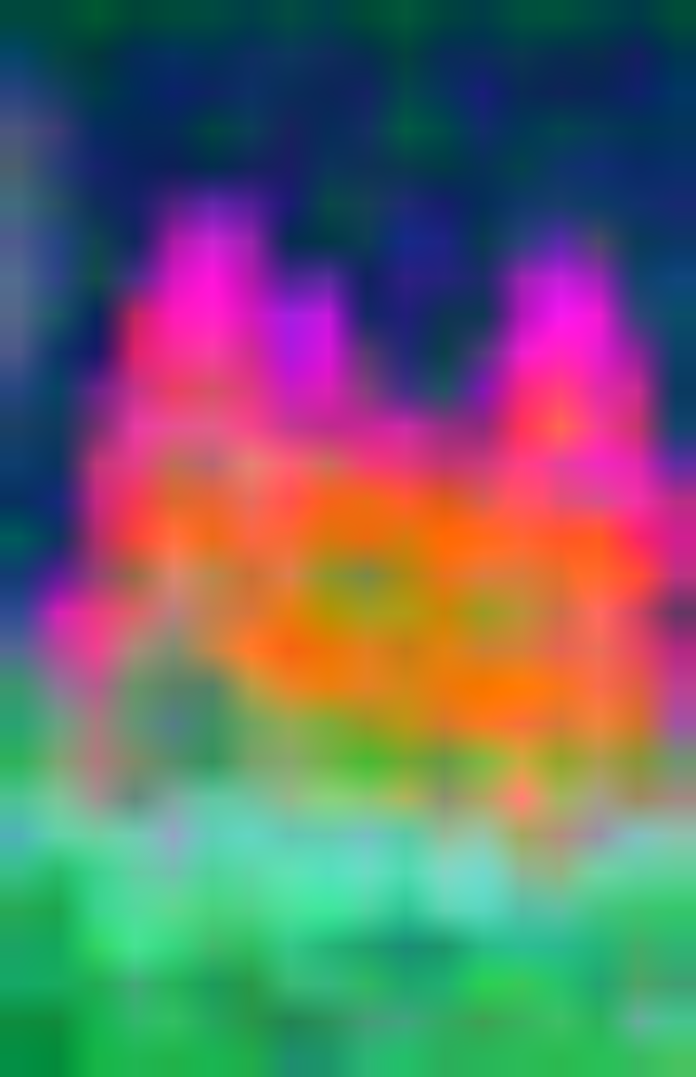}}
\fcolorbox{yellow}{yellow}{
\includegraphics[height=3cm]{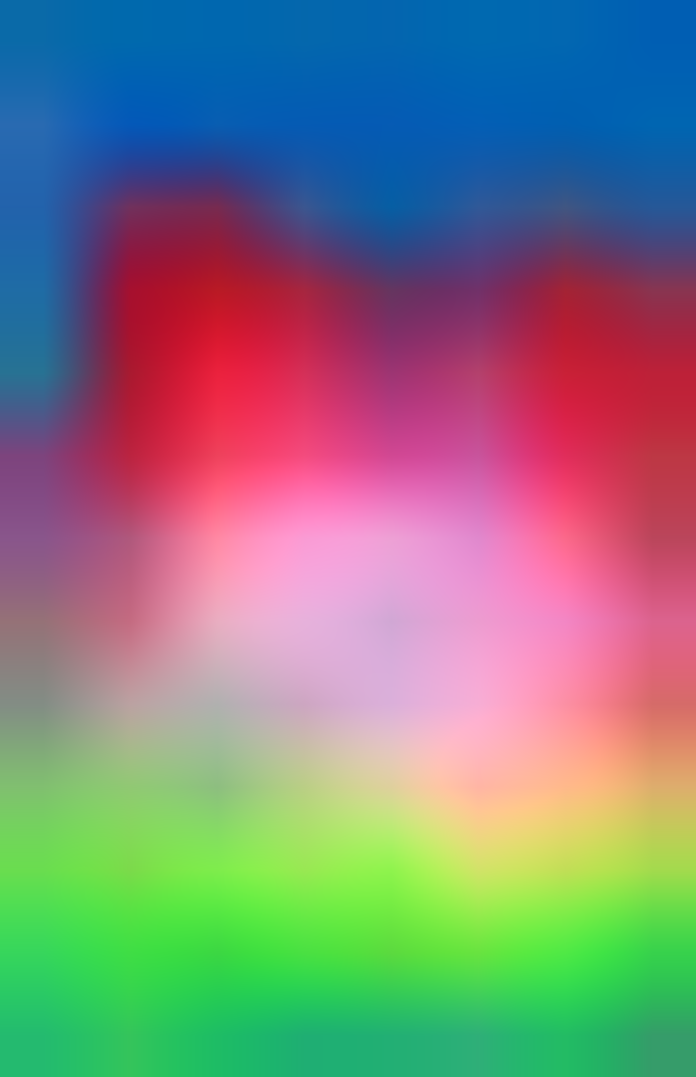}}
\fcolorbox{orange}{orange}{
\includegraphics[height=3cm]{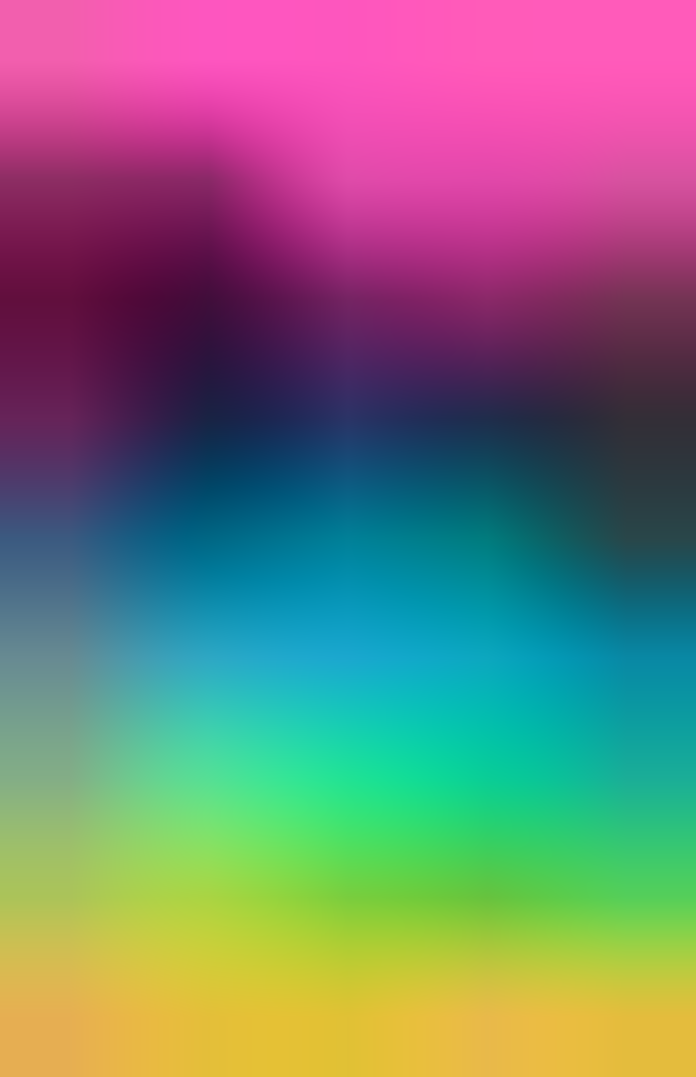}}
\fcolorbox{red}{red}{
\includegraphics[height=3cm]{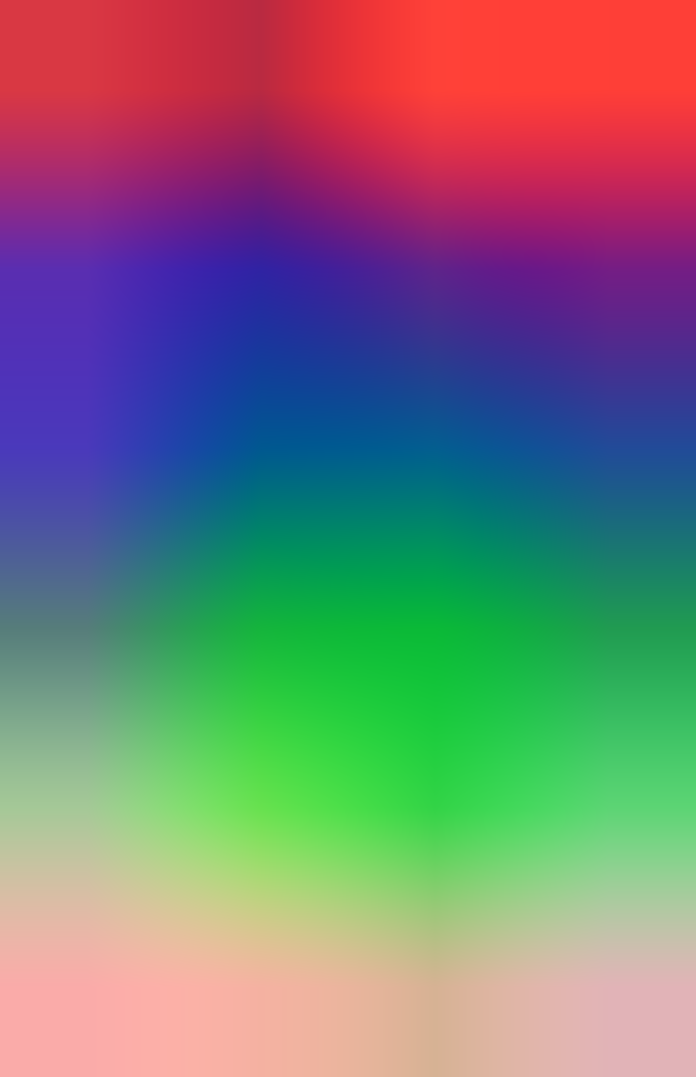}}
\includegraphics[height=3cm]{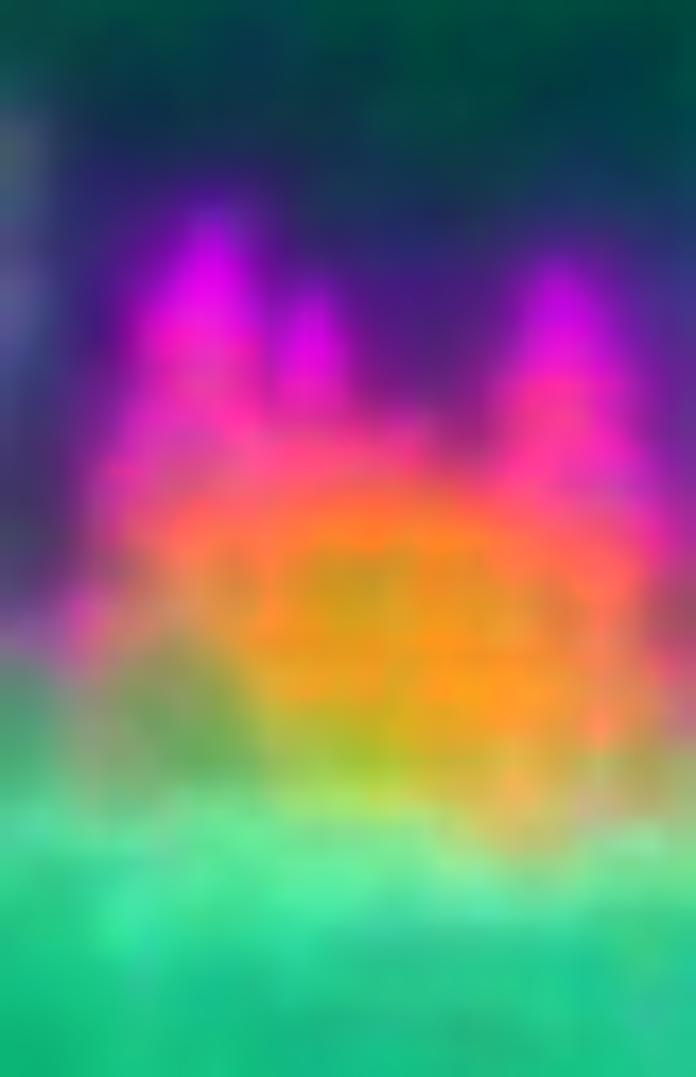}}{}%
\\ %
\rotatebox{90}{\whitetxt{ssssssss}\texttt{Dome}}
\jsubfig{\includegraphics[height=3cm]{figures/clip_visualization/vis_images_bilinear/orig.png}
\fcolorbox{magenta}{magenta}{\includegraphics[height=3cm]{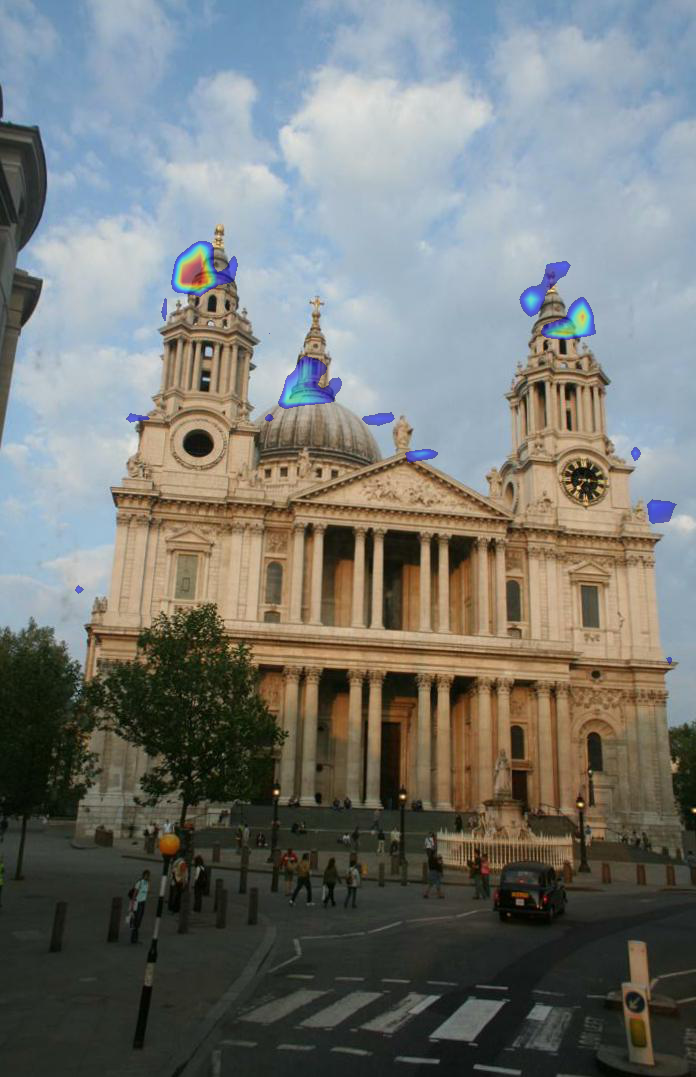}}
\fcolorbox{cyan}{cyan}{
\includegraphics[height=3cm]{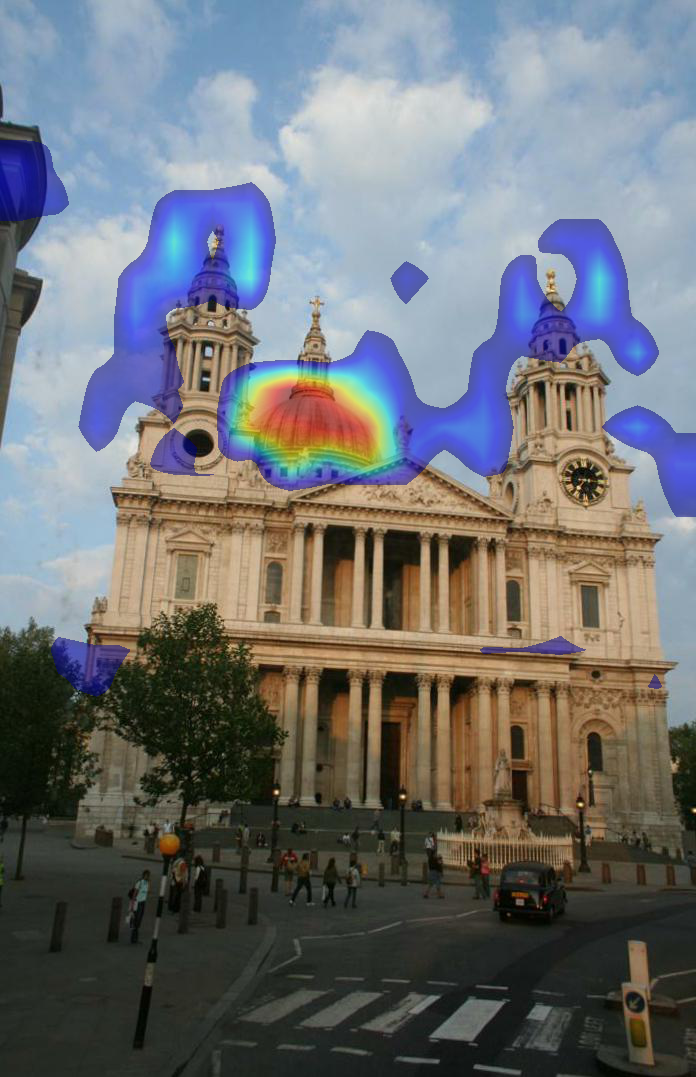}}
\fcolorbox{yellow}{yellow}{
\includegraphics[height=3cm]{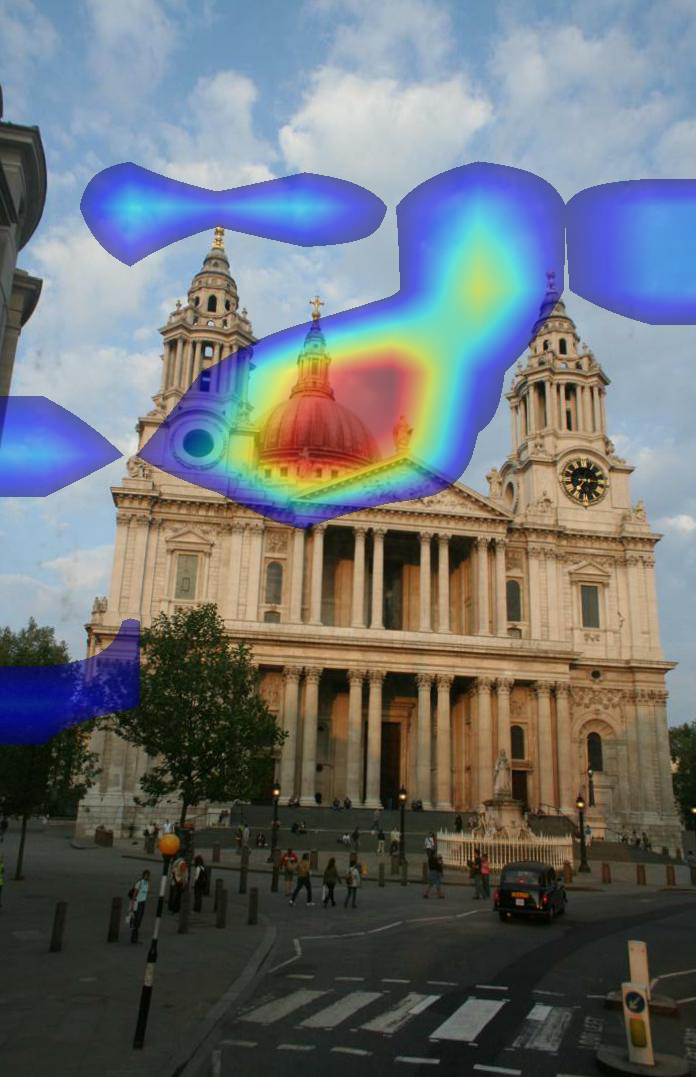}}
\fcolorbox{orange}{orange}{
\includegraphics[height=3cm]{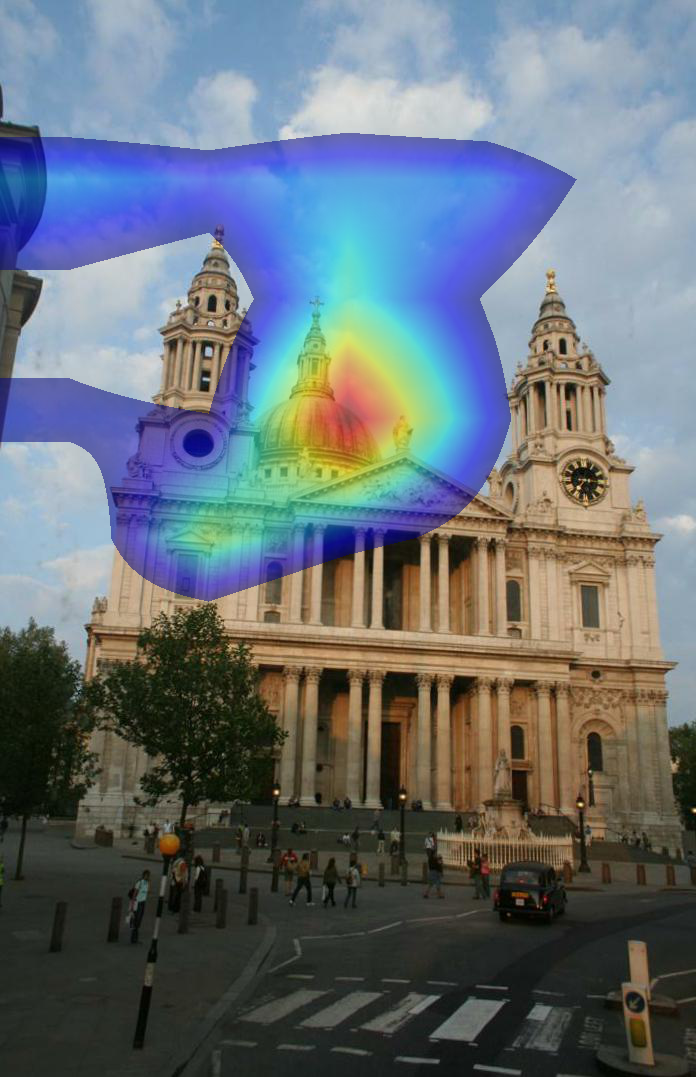}}
\fcolorbox{red}{red}{
\includegraphics[height=3cm]{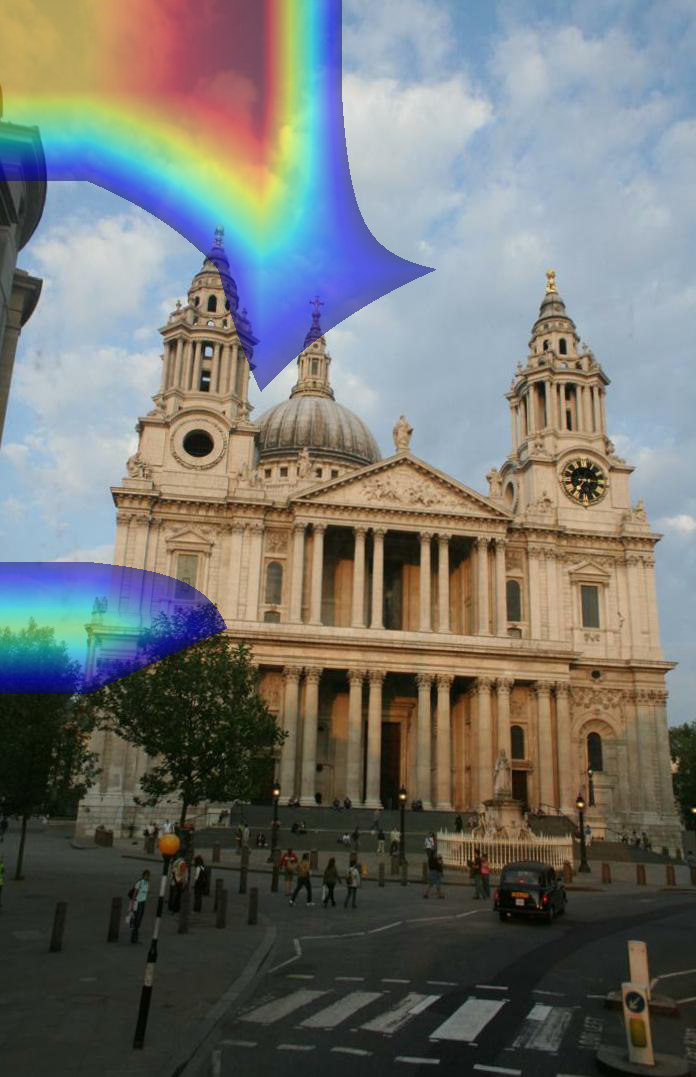}}
\includegraphics[height=3cm]{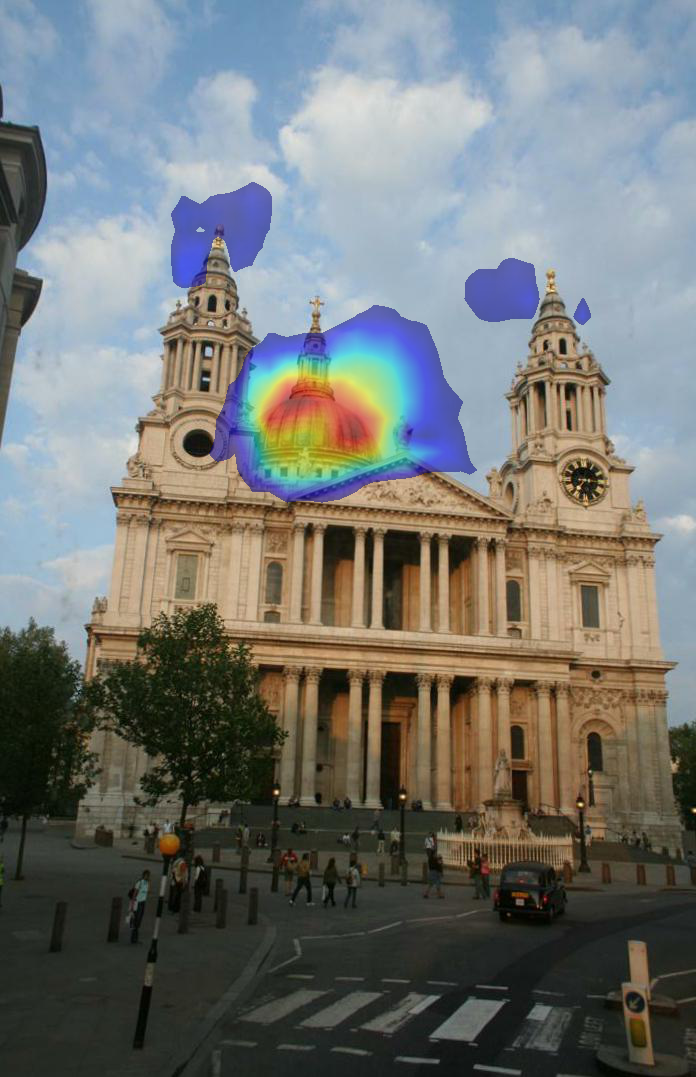}}{}%
\hspace{0.001pt}
\\ %
\rotatebox{90}{\whitetxt{ssssssss}\texttt{Tower}}
\jsubfig{\includegraphics[height=3cm]{figures/clip_visualization/vis_images_bilinear/orig.png}
\fcolorbox{magenta}{magenta}{\includegraphics[height=3cm]{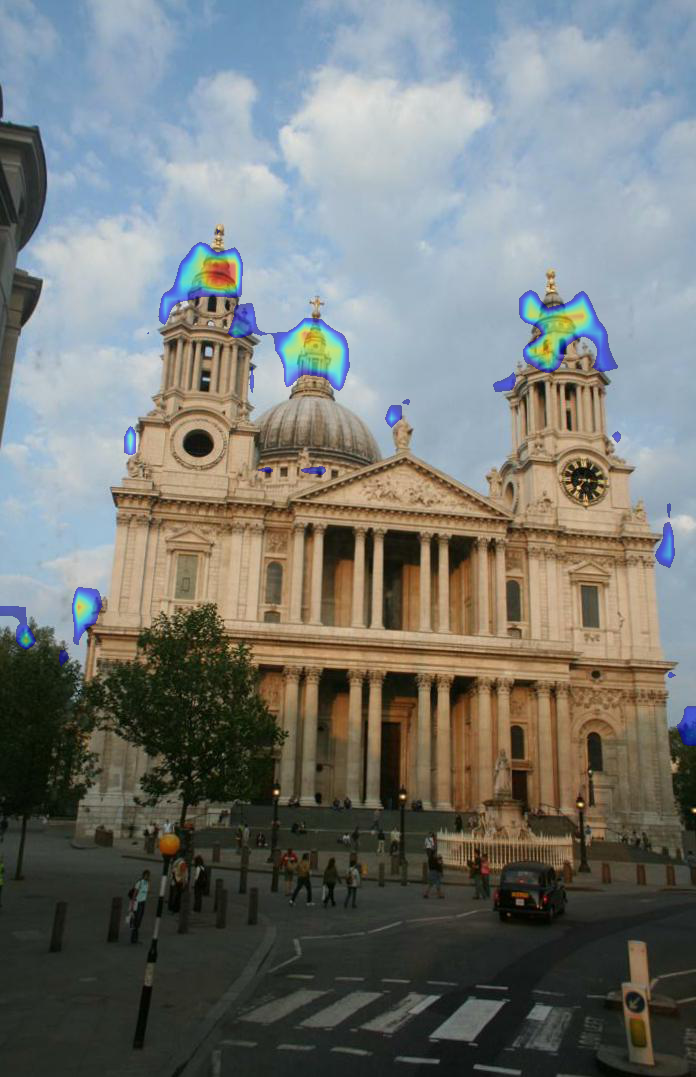}}
\fcolorbox{cyan}{cyan}{
\includegraphics[height=3cm]{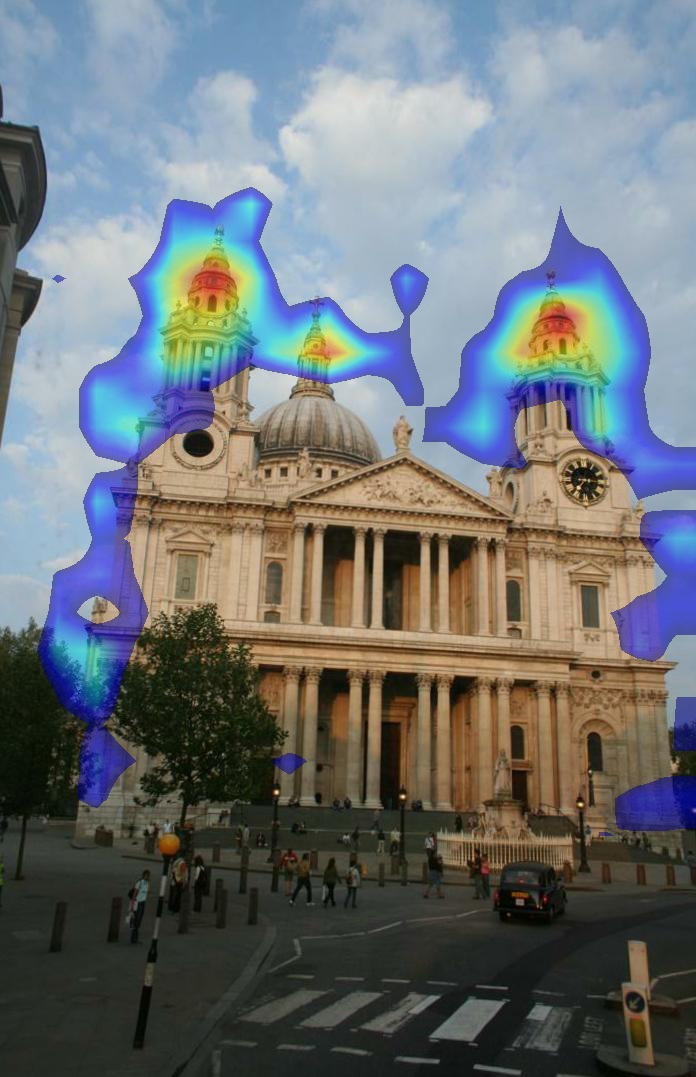}}
\fcolorbox{yellow}{yellow}{
\includegraphics[height=3cm]{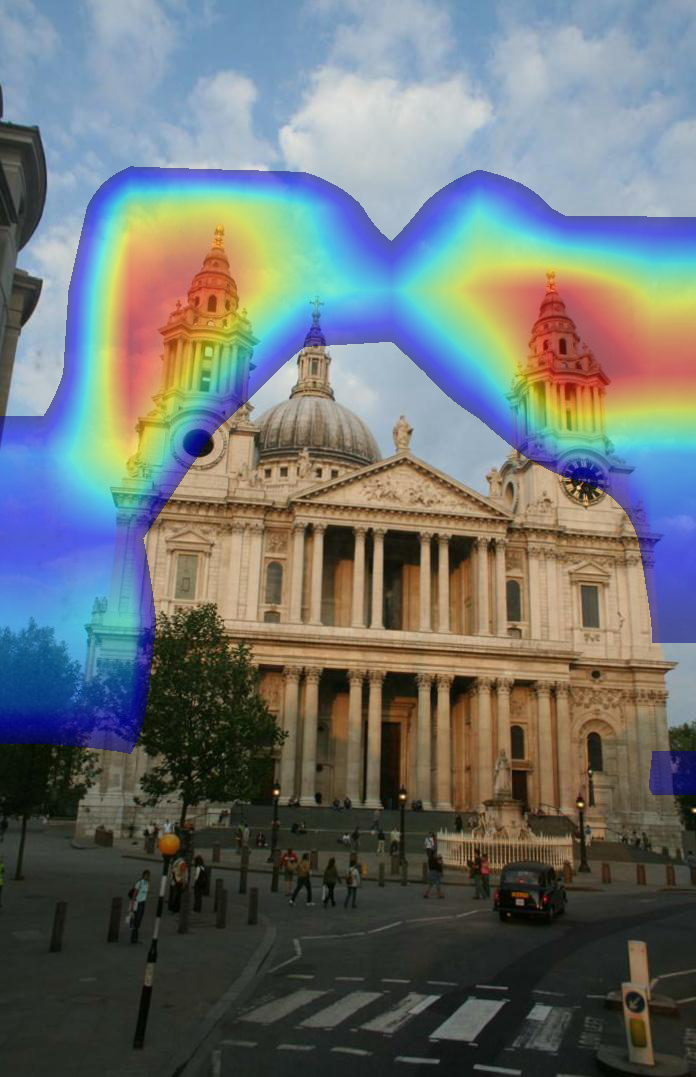}}
\fcolorbox{orange}{orange}{
\includegraphics[height=3cm]{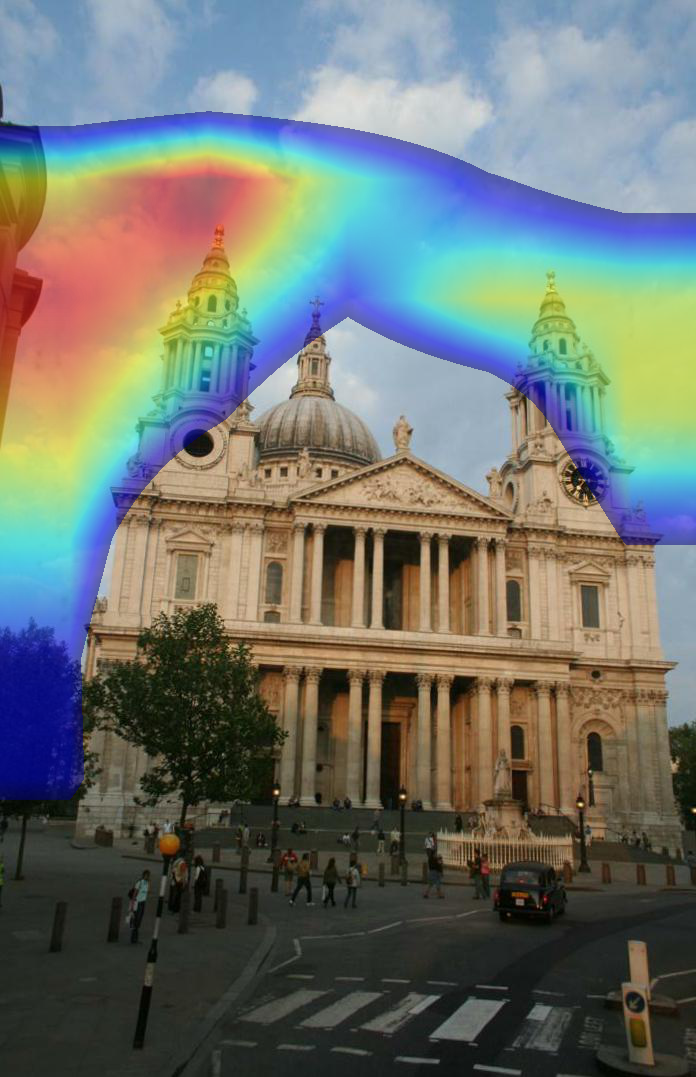}}
\fcolorbox{red}{red}{
\includegraphics[height=3cm]{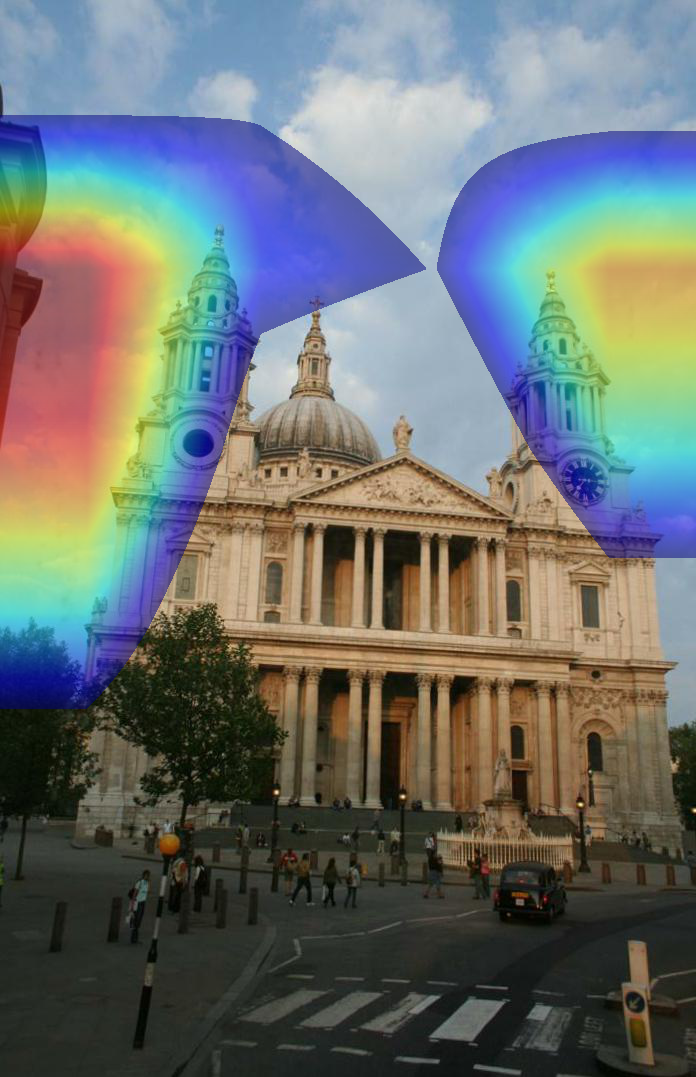}}
\includegraphics[height=3cm]{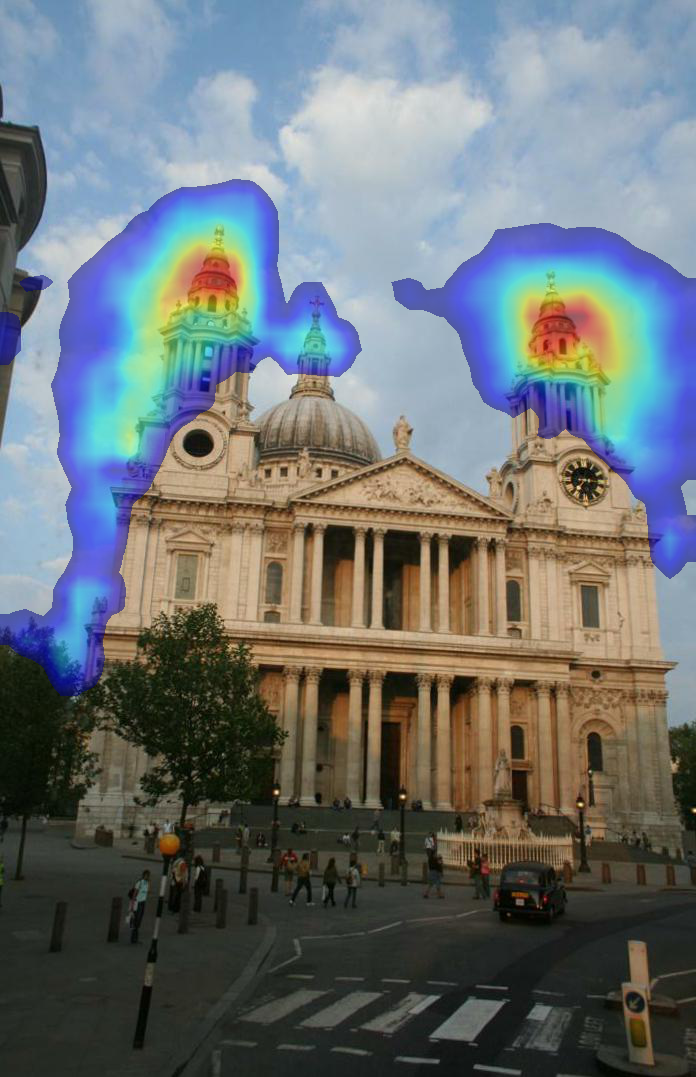}}{}%
\hspace{0.001pt}
\\ %
\rotatebox{90}{\whitetxt{ssssssss}\texttt{Sky}}
\jsubfig{\includegraphics[height=3cm]{figures/clip_visualization/vis_images_bilinear/orig.png}
\fcolorbox{magenta}{magenta}{\includegraphics[height=3cm]{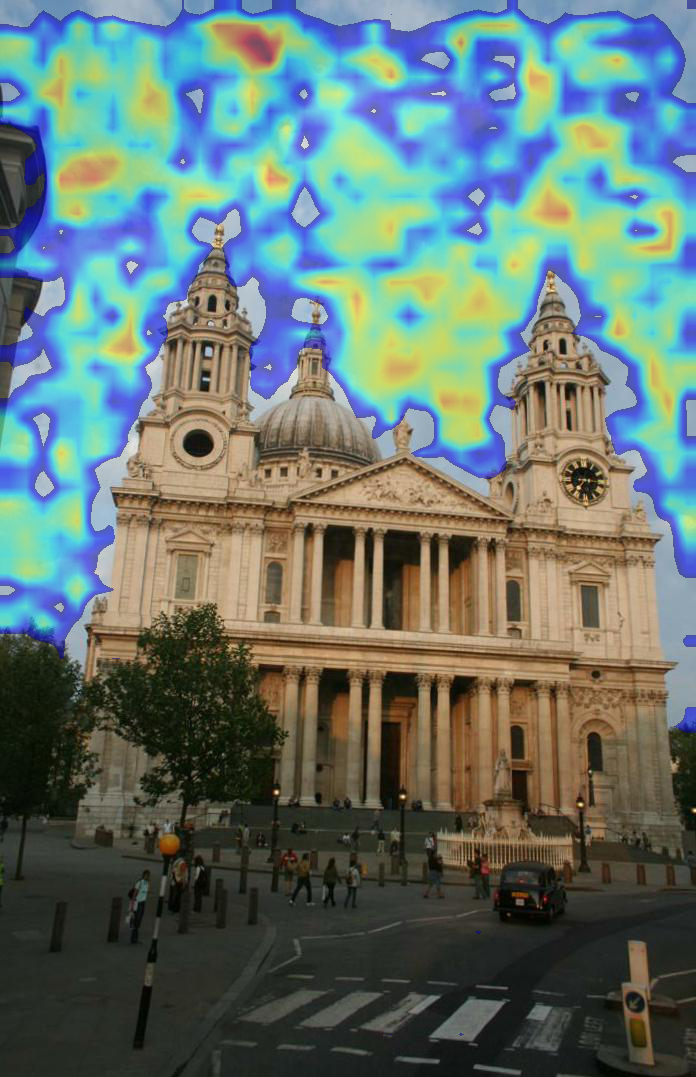}}
\fcolorbox{cyan}{cyan}{
\includegraphics[height=3cm]{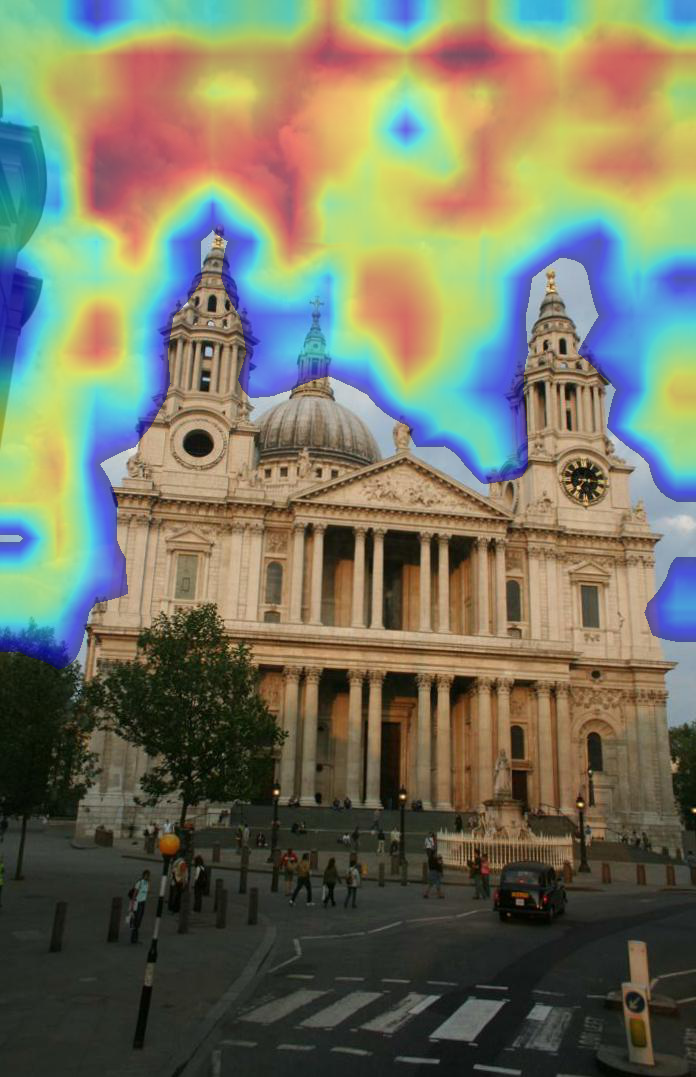}}
\fcolorbox{yellow}{yellow}{
\includegraphics[height=3cm]{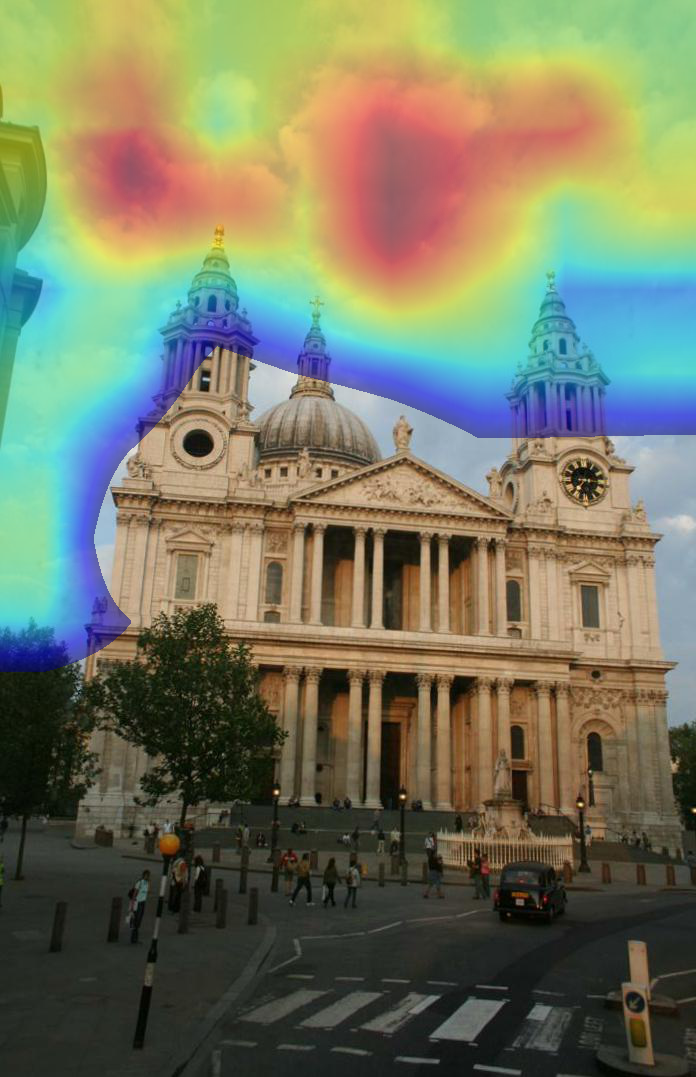}}
\fcolorbox{orange}{orange}{
\includegraphics[height=3cm]{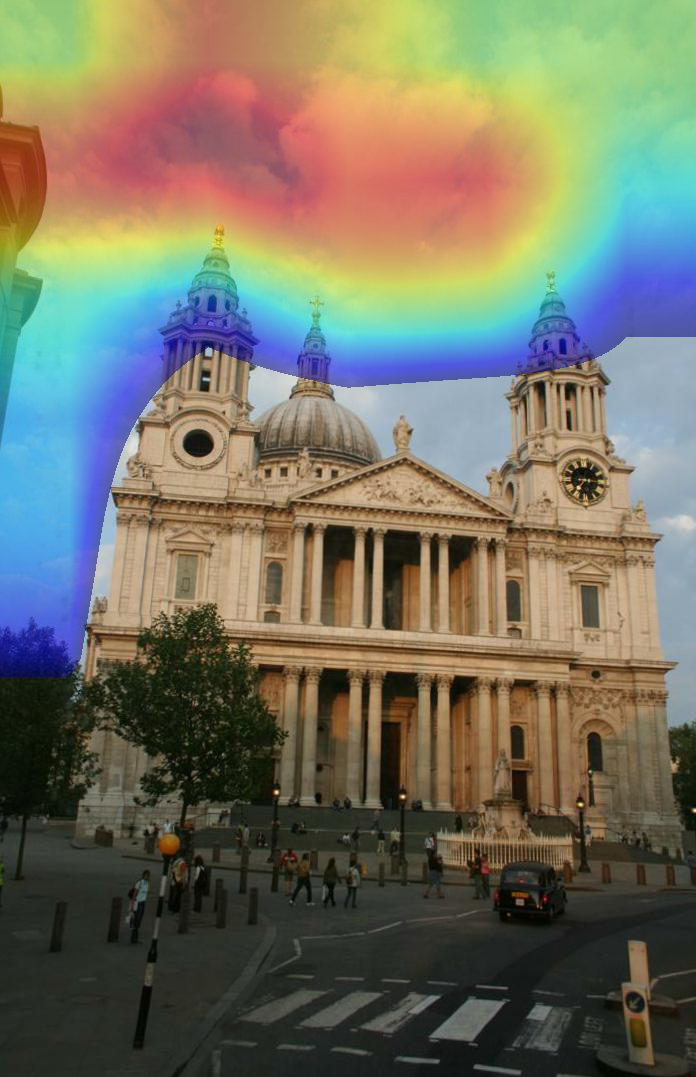}}
\fcolorbox{red}{red}{
\includegraphics[height=3cm]{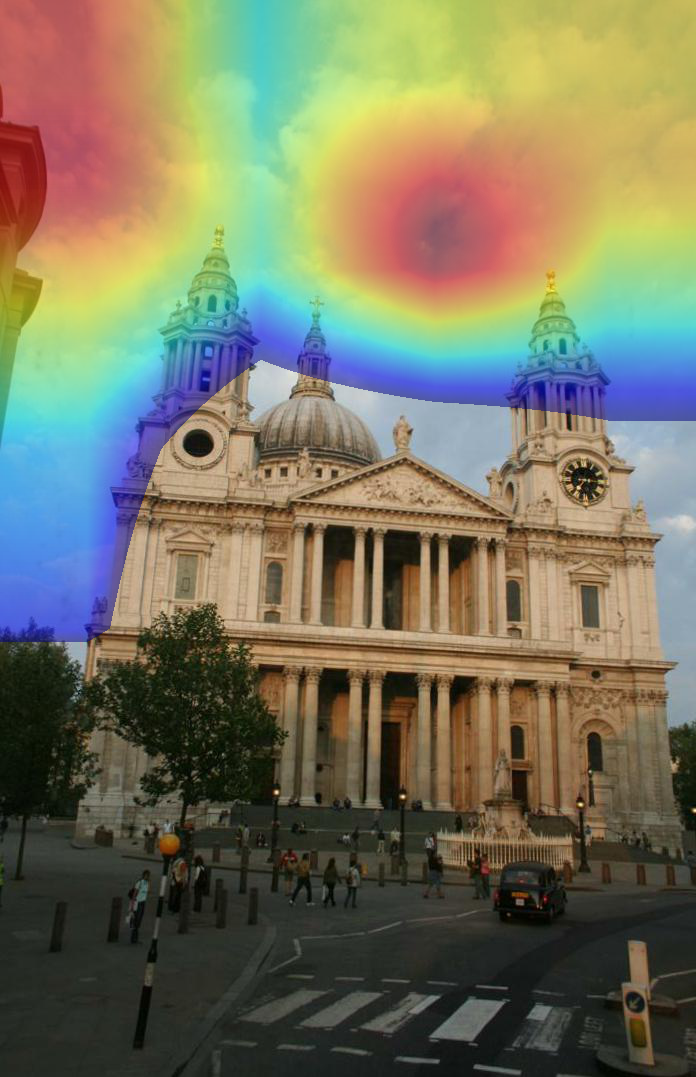}}
\includegraphics[height=3cm]{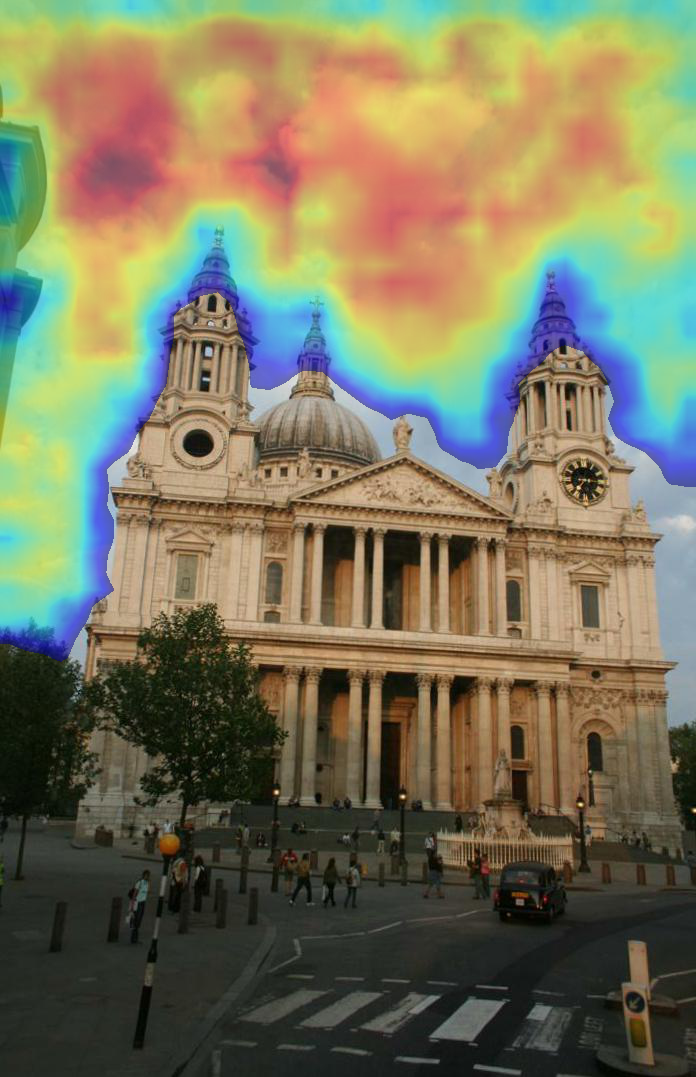}}{}%
\caption{\textbf{Visualizing 2D CLIP Features at Multiple Scales $s$}, with the scales shown on top of the input image on the left. We illustrate PCA visualizations (top row), followed by probability heatmaps extracted following the standard relevancy score procedure proposed in~\cite{kerr2023lerf} over several different textual queries. From left to right, we show: the original image, clip responses a scale of $s=0.05, 0.1, 0.23, 0.34, 0.5$, average of all scales, and the distilled features. As illustrated above, this procedure yields \emph{semantic misalignments}, where the semantics of certain regions leaks to spatially-close regions.
} 
\label{fig:ablations_temporal}
\end{figure*}

%% file: tables/status.tex
\begin{table*}[t]
\centering

\caption{\textbf{Performance breakdown over the HolyScenes dataset.} We report performance for each scene and category in the HolyScenes dataset.}

\setlength{\tabcolsep}{4.5pt}
\begin{tabular}{@{} lcccccccccc @{} }
\hline
& \shortstack{ \small Windows\\(Milano)} & \shortstack{ \small Spires\\(Milano)} & \shortstack{ \small Portals\\(Milano)} & \shortstack{ \small Domes\\(St.Paul)} & \shortstack{ \small Towers\\(St.Paul)} & \shortstack{ \small Windows\\(St.Paul)} & \shortstack{ \small Portals\\(St.Paul)} & \shortstack{ \small Domes\\(Hurba)} & \shortstack{ \small Minarets\\(Badshahi)} & \shortstack{ \small Domes\\(Badshahi)} \\
\hline
Halo-Nerf & \textbf{0.85} & 0.61 & \textbf{0.60} & 0.61 & \textbf{0.90} & 0.68 & \textbf{0.48} & \textbf{0.72} & 0.78 & \textbf{0.95} \\
\rev{Ours} & 0.74 & \textbf{0.67} & 0.40 & \textbf{0.80} & 0.52 & \textbf{0.72} & 0.02 & 0.65 & \textbf{0.82} & 0.71 \\
\hline
\end{tabular}

\setlength{\tabcolsep}{3pt}
\begin{tabular}{@{} lcccccccc @{} }
\hline
& \shortstack{ \small Windows\\(Hurba)} & \shortstack{ \small Portals\\(Hurba)} & \shortstack{ \small Windows\\(Notre-Dame)} & \shortstack{ \small Towers\\(Notre-Dame)} & \shortstack{ \small Portals\\(Notre-Dame)} & \shortstack{ \small Minarets\\(Blue-Mosque)} & \shortstack{ \small Domes\\(Blue-Mosque)} & \shortstack{ \small Windows\\(Blue-Mosque)} \\
\hline
Halo-Nerf & \textbf{0.53} & 0.00 & \textbf{0.63} & \textbf{0.82} & \textbf{0.88} & 0.81 & \textbf{0.68} & \textbf{0.40} \\
\rev{Ours} & 0.45 & \textbf{0.67} & 0.20 & 0.40 & 0.77 & \textbf{0.87} & 0.64 & 0.39 \\
\hline
\end{tabular}

\label{tab:per_category_results}
\end{table*}